**The Upper-Rhine Artificial Intelligence Symposium**
**UR-AI 2019**

# Artificial Intelligence

## From Research to Application

**Andreas Christ, Franz Quint (eds.)**

# The Upper-Rhine Artificial Intelligence Symposium
# UR-AI 2019

**Conference Chairs:**

Crispino Bergamaschi, Univ. of Appl. Sciences and Arts Northwestern Switzerland
Winfried Lieber, Offenburg University of Applied Sciences

**Program Committee:**

Andreas Christ, Offenburg University of Applied Sciences
Martin Christen, University of Applied Sciences and Arts Northwestern Switzerland
Ulrich Mescheder, Furtwangen University of Applied Sciences
Pierre Parrend, ECAM Strasbourg-Europe
Franz Quint, Karlsruhe University of Applied Sciences
Karl-Herbert Schäfer, Kaiserslautern University of Applied Sciences
Marie Wolkers, Alsace Tech Strasbourg

**Organising Committee:**

Thorsten Fitzon, Furtwangen University of Applied Sciences
Katja Fortenbacher-Nagel, Offenburg University of Applied Sciences
Jean Pacevicius, TriRhenaTech Alliance
Ira Pawlowski, Offenburg University of Applied Sciences

**Publication Chair:**

Elena Stamm, Karlsruhe University of Applied Sciences

# Table of Contents

# Foreword

Artificial intelligence (AI) has, with its subdomains deep learning and natural language processing (NLP), progressed in leaps and bounds in recent years. This is particularly so as computing power has received massive boosts thanks to more powerful graphics processes (GPUs) and cloud solutions, cheaper data processing and the generation of untold amounts of usable data.

AI is now used in virtually every area of life and the economy. It is even leveraged, in part at least, by our smartphones (speech recognition), laptops (spam filters) and cars (assistance systems). AI is also increasingly used for bespoke recommendations (Netflix, Amazon, etc.), recognising cancer cells, automated document processing (insurers' claim receipts, customs declarations, etc.), recognising creditworthiness and credit defaults, identifying fraud, and forecasts of all kinds.

The next step will involve developing artificial intelligence capable of carrying out multiple tasks simultaneously. Initial moves in this direction are already underway. An example of this is Zero-Shot translation, which facilitates translation between different languages. It has been programmed to translate both ways between English and Japanese and English and Korean, among others. This then makes it able to perform fairly reasonable translations between Korean and Japanese – without the system having explicitly learned this pairing. The example shows that so-called transfer learning enables the use of general translation know-how. That said, transfer learning is currently restricted to relatively similar scenarios: we are still some way away from real AI.

Notwithstanding the above, developments in the areas of deep learning and NLP are proceeding apace. Not yet a decade old, the methodology underpinning deep learning has already spawned far-reaching innovation – the possibilities and applications have mushroomed.

No one knows what course these developments will follow going forward. Will we still be working for a living in 2040 – or increasingly pursuing our personal interests? One thing is for sure: as AI, with its technical, social and psychological issues, develops, the universities of the TriRhenaTech Alliance will be making interdisciplinary and correspondingly solution-oriented contributions.

*Prof. Dr. Crispino Bergamaschi,*
*Chair of the TriRhenaTech Alliance,*
*President of FHNW University of Applied Sciences and Arts Northwestern Switzerland*

# The *Mobile MEASURE* Metrology Knowledge Schema: an Artificial Intelligence Reasoning over Metrology

Florent Bourgeois[1], Pierre Arlaud[2], Philippe Studer[1], and Jean-Marc Perronne[1]

[1] Université de Haute Alsace, IRIMAS - département d'informatique, Mulhouse
{florent.bourgeois, philippe.studer, jean-marc.perronne}@uha.fr
[2] Actimage GmbH, Kehl
pierre.arlaud@actimage.com

**Abstract.** Mobile devices offer measuring capabilities using embedded or connected sensors. Development teams find many uses of those in order to retrieve data that enables to capture the execution context of applications. Considering that performed measurements might be used in rigorous contexts ; measurement procedures are critical and must produce reliable results.
The definition of correct measurement procedures requires to handle the rules specified by metrology. Such expertise is rarely found in development teams.
We present here how to implement the *sub_SI*, a basic Knowledge Representation encoding specific rules and knowledge of the metrology. This works as an expert system able to reason and answer questions about simple systems of units. The capabilities of the *Metrology Knowledge Schema*, an extension of the presented work are presented. It is dedicated to the validation of measurement procedures accordingly to users defined system of units.

**Keywords:** Knowledge Representation, First Order Logic, Metrology, Reasoning

## 1 Introduction

In the last decade happened the emergence of mobile platforms. They are now powerful devices able to perform a lot of computing tasks and connected to either local or distant networks with different communication technologies. In addition, they embed plenty of measuring devices [1]. Those capabilities make them the perfect support for measurement assistants ; applications dedicated to the assistance of specific end users measuring processes. Those guide the user in the execution of a specific measurement procedure, indicating how to perform the different measurements with either embedded or communicating sensors, performing automated computation, dedicated to answer the user needs.

In this context, Actimage GmbH[3], the partner of those researches, wants to develop a set of applications dedicated to assist operators in mobility contexts. Examples of applications would be to estimate a solar panel installation cost or to diagnose an industrial facility. Those applications require : to produce the specific quantities the user needs, to handle the user-specific instruments and methods and to retrieve not traditional "*measurements*" (client data, colors, design type, . . . )

The development of any of those applications requires expertise in software development and in metrology [2–4]. Each of those applications present similarities ; they are built on a measurement procedure, highly customisable and must conform to the metrology rules. In order to maximise the confidence in the measurement and reduce the development time, a Model Driven Engineering platform has been proposed in previous work [5]. This platform, named mobile MEASURE, describes concepts and tools to model, validate and execute measurement assistants.

[3] http://www.actimage.de

This paper briefly describes what Knowledge Representations [6] are. It then presents how to use First Order Logic to model a simple coherent system of units. At last, presents the reasoning capabilities of the existing *Metrology Knowledge Schema* (MKS) ; the metrology semantics validation Knowledge Representation implemented in the context of the mobile MEASURE platform.

## 2 Knowledge Representations

A Knowledge Representation (KR) is a numerical encoding of a specific domain knowledge. This encoding enables to share the knowledge and to reason over the domain as a human being would. The most used approach is the mathematical logic.

KR hold elements and relations between the elements. From this point of view, they are identical to basics of relational databases. Nevertheless KR also holds complex relations which bind elements and relations. Those describe the rules of the domain to encode. KR description languages are associated with strong semantics derived from mathematical logic. The association of this semantics and an inference engine allows for reasoning over the whole set of entities and rules described. The reasoning deduces relations that are not explicitly expressed in the KR [7]. The association of a KR and an inference engine is usually called an expert system. This approach was the premise of the first Artificial Intelligence (AI) using the LISP and Prolog Languages and it is still employed widely.

## 3 Encoding a Simple System of Units

The International System of Units (SI) paper [3] describes the system of units rules. In this paper we will focus on the creation of a really simple system. Hence we will use a subset of the SI entities and rules to describe our system, the *sub_SI*. The *sub_SI* holds several units. A unit is a reference enabling to encode a real-world event into a quantity. The different kinds of events that can be encoded (length, mass, ...) are named dimensions. We consider that units are either declared as a dimension reference or directly related to an existing unit. Relations between dimensions, conversions of units, prefixes, aliases, unit compositions, dimensional analysis and representational theory of measurement are out of the scope of this paper.

Table 1 lists First Order Logic terms and predicates to model the described *sub_SI* set of rules. Then, using the defined set of predicates, table 2 defines a small system of units with length and time dimensions and some units in each of them.

From the predicates *is_ref_dim(d,u)* and *is_ref_unit(u,v)* it is possible to describe a tree that has a unit reference as root and all other units of the dimension as nodes. It is then possible to add a complex recursive rule *unit_dim(u,d)*from which the dimension of every unit can be deduced :

$$\forall u, d \; is_ref_dim(d, u) \Rightarrow unit_dim(u, d) \tag{1}$$

$$\forall u, v, d \; unit_dim(v, d) \wedge is_ref_unit(v, u) \Rightarrow unit_dim(u, d) \tag{2}$$

**Table 1.** Termes and predicates to model the *sub_SI*

| Variables | Description |
|---|---|
| u, v, d, ... | are any variable |
| **Predicates** | **Description** |
| is_unit(u) | u is a unit |
| is_dim(d) | d is a dimension |
| is_ref_dim(d,u) | defines the unit u as a reference for dimension d |
| is_ref_unit(u,v) | defines the unit v as a reference for unit u |

**Table 2.** A *sub_SI* instance description

| | |
|---|---|
| is_base_dim(length) | is_base_dim(time) |
| is_unit(metre) | is_unit(mile) |
| is_unit(inch) | is_unit(second) |
| is_ref_dim(length, metre) | |
| is_ref_dim(time, second) | |
| is_ref_unit(metre, mile) | |
| is_ref_dim(mile,inch)) | |

The *sub_SI* KR described here can be implemented using Prolog's Horn clauses. Using the inference engine it is then possible to request the produced expert system to get answers to the following questions : does that dimension or unit exists ; what is the dimension of that unit ; what are all the units for that dimension ; what unit defines a given one or what units are defined by a given one.

Generating a system with more units or dimensions is straightforward. The current KR is then able to model any system of units compliant to our *sub_SI* set of rules.

## 4 The Metrology Knowledge Schema

The *Metrology Knowledge Schema* (MKS) is the evolution of the previous section KR. It has to be able to define if a measurement procedure is valid or not accordingly to the semantics of metrology. Our previous work present how an expert system ; based upon a FOL KR can be employed to validate specific domain processes [8].

To answer to those requests ; the MKS models the dimensional analysis brought by the metrology standards and the measurement scales analysis brought by the representational theory of measurement. Both analysis are brought as complex rules.

In order to handle any kind of units and correct dimensional analysis ; base and composed dimensions have been defined. Also, a set of rule is specified to enable units composition with : prefixes, aliases, unit powers, units product, units quotient. Such units can be parsed to deduce the dimension of any unit ; which combined with the constraints brought by dimensional analysis enables to validate any kind of unit operation.

Using the measurement scales analysis enables to constrain the set of operations that can be performed on a specific quantity. Thanks to those constraints it is possible to define new dimensions that are out of the SI. Indeed, it becomes possible to handle ordered (e.g. IQ tests or school grades) or categorised (e.g. colors, human gender) quantities.

## 5 Conclusions

This paper presents an overview of the Artificial Intelligence researches performed in the context of the Mobile MEASURE project ; a project financed by the German BMWI as an AIF Projekt in the context of the *Zentrales Innovationsprogramm Mittelstand.* The objective of the project is the creation of a development platform based on Model Driven Engineering concepts able to model, validate and execute mobile measurement assistants. The presented research is focussed on the validation of the measurement procedures implied in those measurement assistants.

First, the paper presents the Mobile Measure project context. Second, it describes briefly what are Knowledge Representations and how to use them. These notions are employed in the following section to model a KR able to hold a system of units constrained by a reduced set of rules called the *sub_SI.* A set of entities is proposed to describe an instance of a system and its reasoning capabilities are described. Last, the paper summarizes the current state of the Metrology Knowledge Schema ; which is the current version of the Mobile MEASURE measurement procedure validation Knowledge Representation.

Further works imply the development of a web api that enables any user to request an instance of a system of units in the MKS. With such capabilities ; the MKS will be an autonomous micro-service that can answer metrological queries. Then, an expected evolution will be to enable the system to be dynamic ; an administrator will have access to the API in order to modify the instance of the system in order to either populate it or specialize it ; depending on the field of application of the measurement assistant. Also, we recognize that it is currently difficult to follow explicitly describe the current MKS instance and its behavior when asserting that a measurement procedure is valid or when a quantity conversion is performed. Hence, we are looking for integrating the API with a web interface that will be used as a pedagogical platform to explain the MKS inference process.

# HALFBACK Project: The Use of Machine Learning to Achieve High-Availability in Production

Christoph Reich[1] and Ahmed Samet[2]

[1] Furtwangen University of Applied Sciences
christoph.reich@hs-furtwangen.de
[2] INSA, Strasbourg, France
ahmed.samet@insa-strasbourg.fr

**Abstract.** The HALFBACK project's goal is to achieve a high-available production, mainly by predicting failure of a manufacturing machine and machine tool to avoid downtime, associated costs, and reputation loss.
In this paper two different approaches of machine learning prediction are described. First a pattern mining approach analysing condition events and second a neural network based approach. Further, the paper discusses the need for data pre-processing and ideas how to achieve high-availabilty in production.

## 1 Introduction

The unexpected failure of machines or tools has a direct impact on production availability (+6% increase in costs in 2015 according to the Association of German Machine Tool Builders (VDW)). This gives rise to risks in terms of product quality, profitability and competitiveness. The exact planning of a preventive maintenance is therefore an essential prerequisite for positively influencing quality and production.

The HALFBACK project extracts expert knowledge from sensor data of the production line in order to be able to detect defects and to detect them, and to implement optimized maintenance planning. In order to improve the availability of your own production without having to rely on cost-intensive reserve machines or other means of minimizing downtime, it is necessary to reduce production in part and in each case by to be able to outsource the production as required. For this purpose, an intelligent machine broker will be implemented, which will coordinate the and machinery. The HALFBACK project thus aims to improve the cross-border production process between SMEs. This makes it possible to realize the enormous potential for value creation, strengthens the the digitization competencies of companies, intensifies industrial cooperation and promotes the formation of Company networks in the Upper Rhine.

The aim of this paper is to focus on machine learning concepts, which have been used in the HALFBACK project in order to predict machine maintenance and quality maintenance to re-plan the production line to achieve high available smart production.

The paper is organized as follows: Section 1 outlines the HALFBACK project and its need for machine learning. After the Section 2 two machine learning based approaches are described for predictive maintenance (Section 3) and quality prediction (Section 4). Section 5 points out the importance of data pre-processing and Section 5 shortly explains how to support cross-boarder production. Section 7 concludes the paper.

## 2 Related Work

There have been a number of papers published using machine learning methods for predictive maintenance. Here is just a selection of some.

Mining from dataset to extract correlation is a task that has been well treated in literature [1]. However, it has been shown recently that sequential patterns are not sufficiently informative in several application fields such as network alarm [2] or analysis of human activity [2]. Therefore, the chronicle pattern model, that is an extension of sequential patterns, has been introduced [3]. In [4], Dousson and Duong made the foundation of what has been later known as chronicle mining. In [5], Cram et al. introduced the HCDA algorithm, to mine the complete set of chronicles. Finally, in [6], Dauxais et al. proposed a new approach to extract discriminant chronicles in to the context of pharmaco-epidemiology.

A typical approach is to build an IoT infrastructure and collect data from machines and use neural networks to build a forcast model, as Kanawaday and Sane [7], in their paper did, by modelling slitting machines by AutoRegressive Integrated Moving Average (ARIMA).

In the paper from Sezer et.al [8], they outline the base concepts, materials and methods used to develop an Industry 4.0 architecture focused on predictive maintenance, while relying on low-cost principles to be affordable by Small Manufacturing Enterprises. The result of this research work was a low-cost, easy-to-develop cyber-physical system architecture that measures the temperature and vibration variables of a machining process in a Haas CNC turning centre, while storing such data in the cloud where Recursive Partitioning and Regression Tree model technique is run for predicting the rejection of machined parts based on a quality threshold.

## 3 Pattern Mining for Failure Understanding

To optimise machines' performance, critical events should be anticipated. Predictive maintenance is based on this principle. It consists of collecting and analysing the data from industrial equipment. Then, an alert system learn from previous event sequences and prevent from imminent failures. Intelligent systems that allow this kind of maintenance are based on analysing collected signals, that are generally a set of timestamped events. For this aim, data mining techniques are particularly suited for this task, especially sequential data mining and frequent sequential pattern mining to extract failure cause from collected data. Moreover, patterns mining output could be difficult to read and interpret even for domain experts. A graph based and richer pattern exists that break this limit and is called *chronicle*. Chronicle are a kind of patterns that represents events related with time constraint within a same model.

*Example 1.* Assuming a sequential database that collect data from machine regularly. The graphical model below details a chronicle pattern extracted from the database of frequent events with their time constraint intervals.

| Sequence Id | Events |
|---|---|
| 1 | (A,0), (B,5), (C,7) |
| 2 | (A,2), (B,3), (C,7) |

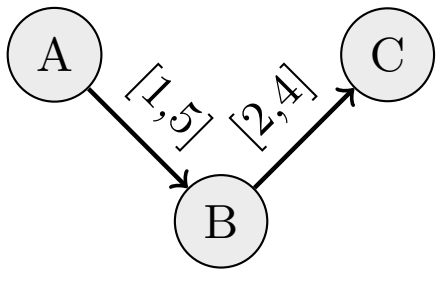

In this work, we seek to develop an approach to mine information from machine data log. These extracted data are modelled as chronicles. Our contribution solves this problem and aims to answer two distinct questions, i.e. a) Is there a correlation between sensors data values and failure? b) How can we use temporal constraints between events, and therefore chronicles, to predict anomalies before they occur?

To answer these questions, we introduced a new approach, called CPM for *Chronicle mining for Predictive Maintenance*. Our interest in this kind of temporal pattern lies not only in predicting an event, but especially in the time interval in which that event will occur (in our case a machine failure). Like any knowledge discovery process, our approach starts with a pre-processing step, a mining step and a third step for the interpretation of extracted knowledge.

The CPM approach is validated through a set of experiments performed on the mining phase as well as the prediction phase. Experiments were achieved on synthetic data as well as in a real industrial data set. Figure 1 shows the developed CPM software for chronicle mining.

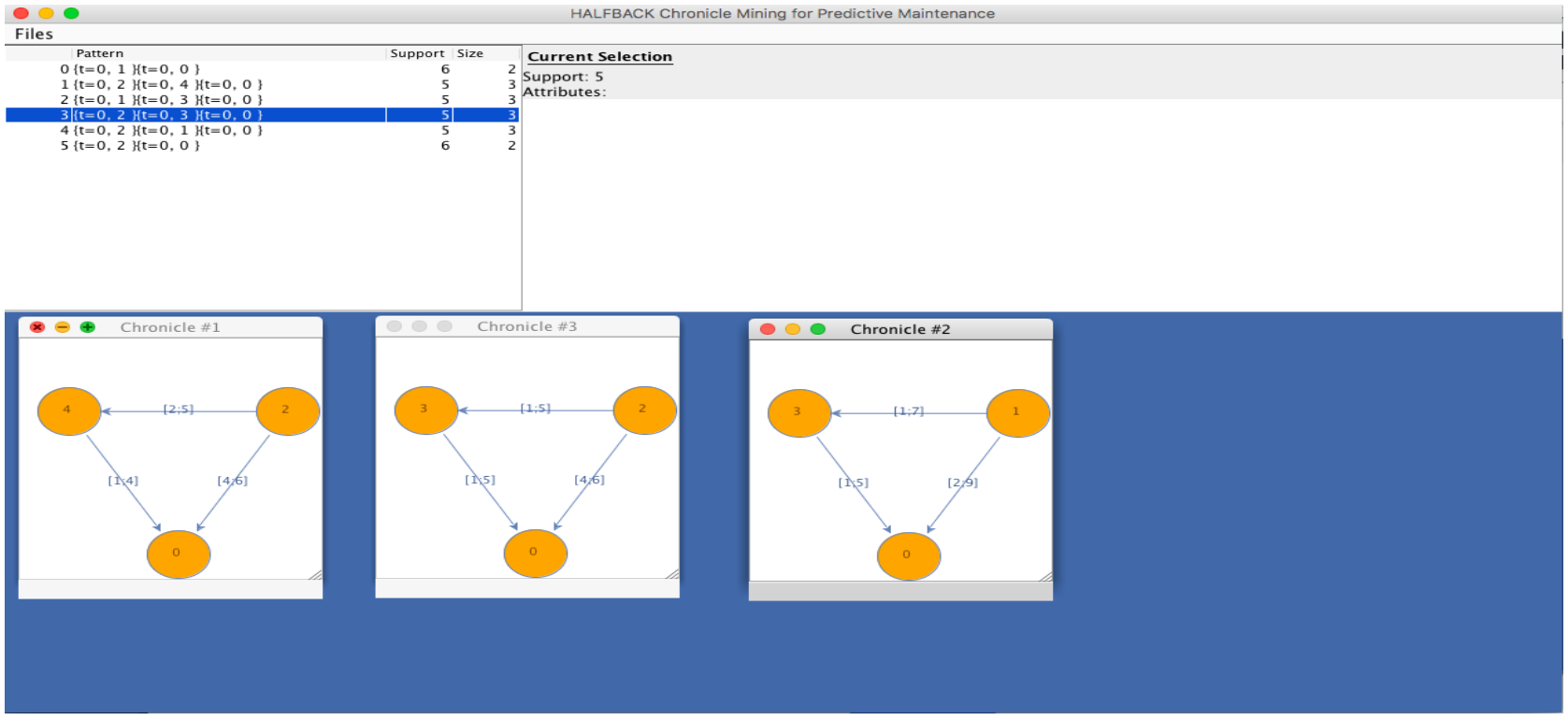

**Fig. 1.** CPM software

## 4 Predictive Maintenance and Quality Prediction with Neural Networks

Figure 2 shows the most important items to be considered, if you want to achieve high-availability for production. If one of these aspects are causing trouble the overall production will stutter. The approach of the HALFBACK project is to ensure high-availability manufacturing processes by forecasting failures of machines, tools, product quality losses, resource flow problems, etc. and through optimized and intelligent ways to plan maintenance at the right time, replace components in time, re-plan manufacturing processes and even plan production relocation to another company. This is to be achieved by collecting data on machines and tools using suitable sensors. In addition, information is collected from the manufacturing environment, the product itself, and the operator's expert knowledge. Big Data algorithms analyze the collected data in the cloud [9] to understand processes and learn from operators' experiences with the goal of avoiding machine damage, loss of quality or maintenance requirements in the future. to predict. This allows the company to act before the manufacturing process stops.

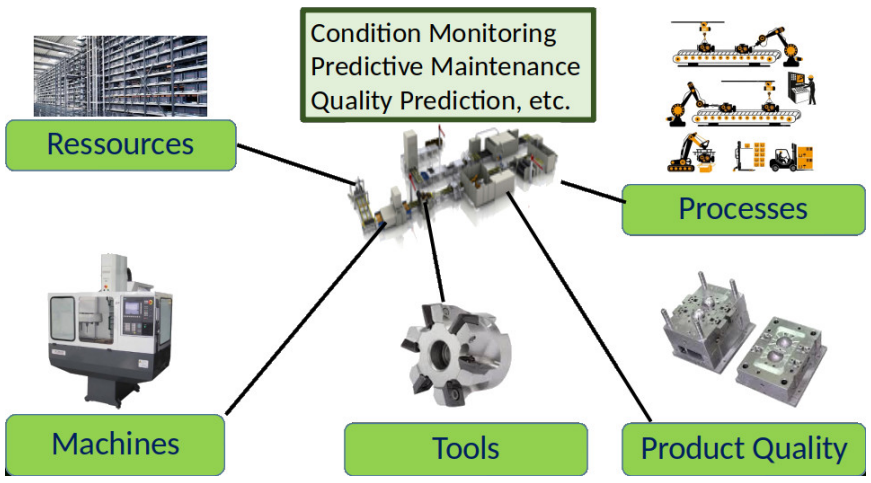

**Fig. 2.** Prediction in Production

The area of Machine Learning includes different methods / technologies to enable a computer to make decisions based on data. The computer learns to draw conclusions about the correct result on the basis of training data, without a human being (expert) having to define fixed if-then rules. For maintenance prediction, a typical use case for using neural networks (NN) for modelling a multi classifier. The classification trains the NN model to assign the input vector x to class y. The classes are predefined before the training (in case of anomaly detection the classes could be e.g. "anomaly" and "no anomaly"). The need for neural network as a justified by Susto et.al [10] in Figure 3. In Figure 3 you see two-dimensional data space with orange circles (correct behaviour) and blue plus (failure behaviour). It can easily be seen, that linear classifier is sufficient enough for a) and a non-linear for b) and a multi multi classifier for c), which can be realized by neural networks.

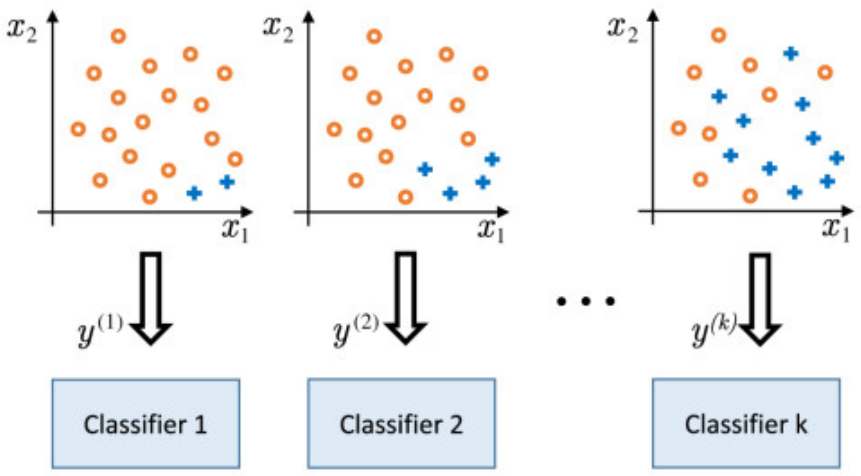

**Fig. 3.** Classifier a) one-dimensional b) two-dimensional c) multi-dimensional (see [10])

First results in the HALFBACK project have shown, that for predictive maintenance the vibration of a certain parts of the machine are important input values for a neural network. Bo Luo et.al [11] used a deep learning model to construct automatically select the impulse responses from the vibration signals to predict the next maintenance of a machine. A good start for training the NN for classifying the part which will have to be replaced during the maintenance phase are the machine parts, that historically have failed.

More often then failures are downtime because of quality loss during production, and therefore downtime caused from tool maintenance. In [12], Bai et al. compared feed forward neural networks, least squared support vector machines, deep restricted Boltzmann machines and stack autoencoders to predict quality in a manufacturing process. The dataset consisted of 19 process parameters, some are adjustable parameters and some non-adjustable, and one quality index in a range between zero and one. They trained different models via trial and error method to find the best fitting model for each archi-tecture. Furthermore, they tried different sample sizes, 100 and 1000. It turned out that the deep restricted Boltzmann machines and stack autoencoders outperformed the other model architectures. They have also shown that the bigger the samples, the better the performance.

## 5 The Need of Data Pre-Processing

An important aspect is the pre-processing after the data has been collected from the machine. Data collected from various sources will contain noise, redundancy, and inconsistency and hence it is a waste of time and resources to to store such data. Analytical methods have requirements of data quality therefore, for proper data analysis per-processing data is important. The data has to be cleaned depending upon the kind of noise or artifacts present in it also depending on the kind of analysis that will be performed on the data. To do so a highly scalable infrastructure has been build for HALFABCK, as seen in Figure 4, provided in a cloud [13]. Only with high-quality data

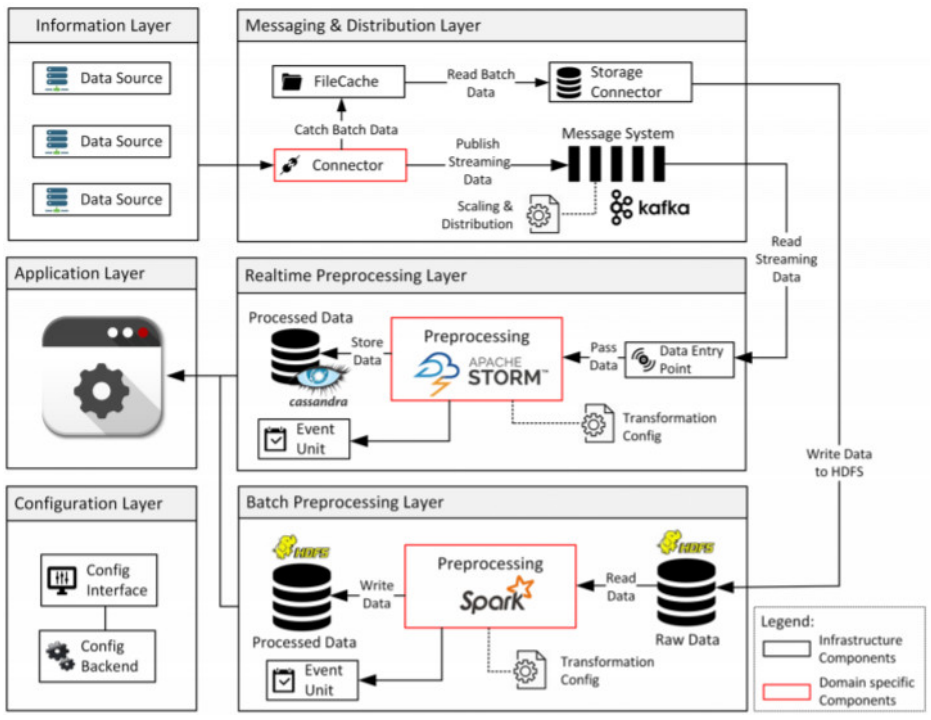

**Fig. 4.** Data Analysis and Modelling Infrastructure in the Cloud

high-quality models can be build.

## 6 High-Availability Factory by Cross-Boarder Production

In case the maintenance of a machine can not be planed without causing high cost, because of failed delivery times and therefore high penalty costs and reputation loss. A solution could be to shift the production to another factory. Prerequisites are to find a company, which has the same or an equivalent machine to overtake the production. Therefore a goal of the HALFBACK project is to describe a machine's static information (type of tool holder, etc.) and dynamic information (speed, etc.), called a virtual profile (footprint). To support the search for a suitable machine, the virtual profiles (footprints) of the machines are to be registered in the cloud with a "High Availability Machine Broker". The footprint of a machine contains the location of the machine, Machine availability, functional description, etc. The broker enables the machine to be used as a to be able to offer services to other companies, like "Machine as a Service". In case of unavoidable machine failure the HALFBACK software can use the "High Availability Machine Broker" to create an adequate search for a machine replacement and move production to another plant, as depicted in Figure 5.

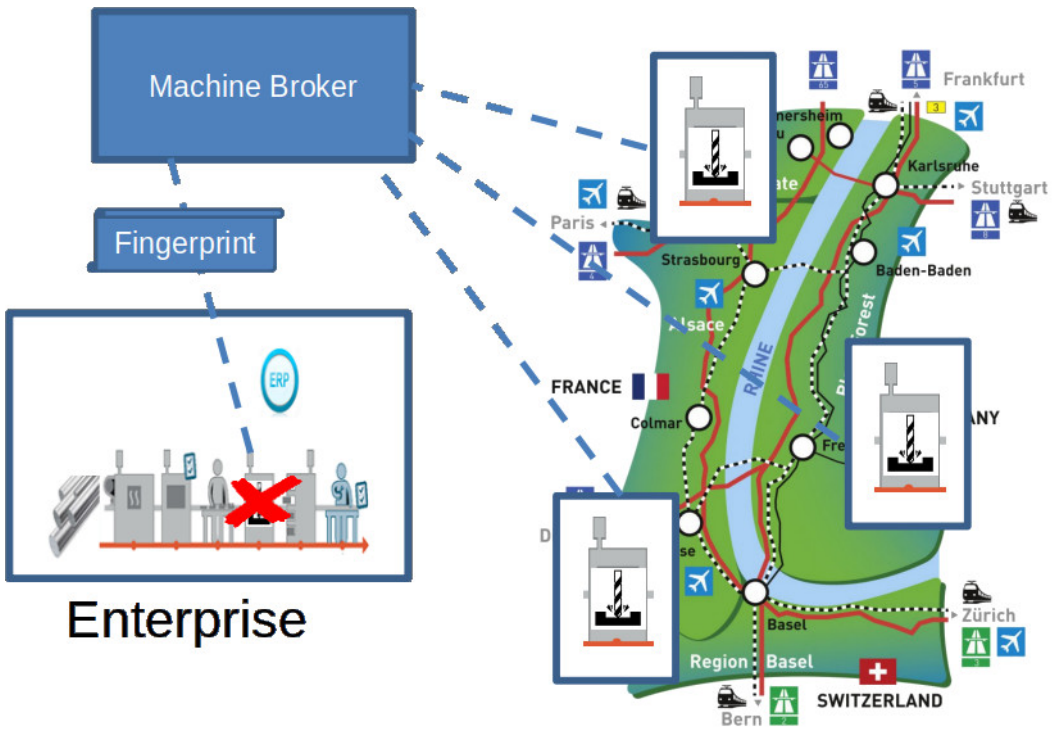

**Fig. 5.** Cross-Border Production

# 7 Conclusions

This paper discussed two machine learning approaches the *Chronicle mining for Predictive Maintenance* to analyse log events and predict the maintenance of a machine and the Neural Network approach for predicting the quality of a workpiece. Both approaches are used to plan actions to overcome the downtime of a machine. In order to avoid contractual penalties for late product delivery to the customer, the concept of Machine as a Service has been introduced. This can help the factory owner to shift the production to a factory near by.

# Feature Extraction from Raw Vibration Signal and Classification of Bearing Faults Using Convolutional Neural Networks

Tanju Gofran[1], Peter Neugebauer[1], and Dieter Schramm[2]

[1] IEEM - Institute for Energy Efficient Mobility
Karlsruhe University of Applied Sciences
gota0001@hs-karlsruhe.de
peter.neugebauer@hs-karlsruhe.de
[2] University of Duisburg-Essen
schramm@mechatronik.uni-duisburg.de

**Abstract.** Safe, efficient and uninterrupted continuous operation of an electric motor requires real-time condition monitoring of its rotating parts. Other than knowledge based signal analysis, fault feature extraction with statistical information or signal processing methods can be used to classify different fault patterns. But rule based feature extraction methods do not have domain adaptability, so the fault classification working in one system may not work for another system. A deep learning algorithm - Convolution Neural Networks approach is shown in this paper to classify different bearing faults and the trained network shows a good fault prediction capability for other systems.

**Keywords:** CNN, CMCNN, Conv Layer, Feature Map Analysis

## 1 Introduction

To avoid unwanted shut-down due to rotating machine fault in a manufacturing line , in a remote power plant and in many other applications a real-time monitoring of the machine condition is a high demand.The aim of Condition Monitoring (CM) of electric machines is to acquire a “health” indication in real-time; in order to identify possible failures in advance, thus avoiding costly and unscheduled down time, upholding accurate servicing schedules [1]. An ideal CM method should be non-destructive and should depend on easily measurable parameters[2]. In contrast to sensor-based techniques, data-driven condition monitoring methods are interesting because they do not require any knowledge about the machine parameters; instead, they only require a database of both healthy and faulty conditions of the machine for eventual feature extraction and classification. Rule based feature extraction and classification of machine conditions are studied in many publications[2–5]. The accuracy of classifying faults mainly depends on how accurate the feature extraction is.The limitation of hand crafted feature extraction is, the classification working in one domain most cases do not work in other domains. Deep Learning is a subclass of Machine Learning (ML) algorithm which can extract features directly from data without prior knowledge or mathematical background of input and then classify as required. Convolutional Neural Networks (CNN) is one Deep Learning technique believed to be the most popular ML algorithm in present time. Because of the advancement of computation technology and efficiency of CNN, it is now a widely used algorithm among researchers to solve many real time problems in various fields of natural science, computer science and engineering. In recent years many publications studied CM using CNN and Deep Neural Networks (DNN)[6–9]. [10]show a domain adaptability of classifying different bearing vibration signal using CNN . [11]proposed hierarchical deep architecture of CNN in which original data is converted into 2D data to classify bearing faults and their sizes. Many of the published work of CM with CNN approaches show very high accuracy, but these are mostly tested on the same dataset. In this work, we propose a simple CNN architecture for classifying bearing faults from 1D vibration signal, which is trained with one test bench dataset and tested on different test bench data. We propose an analysing approach of extracted features by the trained CNN model.

## 2 Convolutional Neural Networks (CNN)

Convolutional Neural Networks were inspired from the biological process of connectivity pattern of cells of visual cortex of cat and monkey published by Hubel and Wiesel in 1962 and 1968. They showed how without moving the eyes individual cortical neurons respond to stimuli only in a restricted region of the visual field known as receptive field. Modern CNN models also use the similar kind of approach to extract simple to more complex features of input image for classification or detection.

## 3 Proposed CNN Approach

The main constraint of training CNN models for classifying faults of motor is to have access of a large dataset. In our research project we develop a test-bench (IEEM-CMTestBench) to create different faulty bearing vibration data, but the challenge remains to implement real life faults on bearings. While the project is still in process, to continue our investigation we used a public dataset to train CNN model to classify bearing faults. We named our CNN model as CMCNN-Condition Monitoring Convolutional Neural Network. The trained CMCNN shows high accuracy to classify different bearing faults and finally we test the trained CMCNN model with a different system dataset to verify the model accuracy. The proposed approach is described in the following chapters:

### 3.1 Training Dataset

The bearing dataset for different faults is prepared by Case Western Reserve University (CWRU)[12] and the dataset is used to train CNN models in number of publications[8, 10, 6]. CWRU dataset provides vibration data for normal bearings (No Fault) and faulty bearings. Faults are artificially implemented on Inner Race (IR Fault), Rolling Element (RE Fault) and Outer Race (OR Fault) of both Drive-End (DE) and Fan-End (FE) bearings using electro-discharge machining (EDM). Fault diameters are ranging from 0.007 inches to 0.028 inches in diameter and vibration measurements are taken at 4 different loads.In order to quantify vibration response effect with load zone, measurements were conducted for both bearings with OR Faults located directly in the load zone(OR@3), at orthogonal to the load zone (OR@3), and at 12 o'clock(OR@12).

### 3.2 Data Preparation and Data Argumentation

The CWRU dataset includes 6 classes: No Fault, IR Fault, RE Fault, OR@6 Fault, OR@3 Fault and OR@12 Fault. Each class has less than 30 or sometimes even less than 15 datasets. Generally for successful classification with a Deep Learning Algorithm a minimum of 1000 dataset per class is required. Similarly as[10] we also apply data argumentation to segment each signal data into smaller sizes to increase the number of datasets per classes. In this work, we consider to segment the signal in a way that the input data fed to the CNN model should be equivalent approximately to unit revolution of the bearing. To differentiate random vibrations to faulty or healthy vibration a set of Random Noise is also included in the Training Data.
Our work is comparable with [10] for CNN model design in some extend, where the authors used CWRU datasets to classify 10 classes. In their work each class has a certain size of fault (0.007 to 0.021 inch) and the 10 classes data are decided into 4 combinations of loads for training. In our work we classify five bearing fault classes and one No fault class, by extracting feature from data of all different fault sizes and loads of measurements.

### 3.3 CMCNN Architecture

The CMCNN contains five layers each containing Convolution (Conv) Layer, Rectified Linear Unit (ReLU) or Activation Layer and Max Pool (Pool) Layer for feature extraction and three Fully Connected Layers (FC) with 50% dropouts for classification. The model is designed and trained with MATLAB deep learning functions. The CMCNN architecture is shown in 1. The main block of a CNN is the Conv Layer, which is done by sliding a filter or kernel over the input data producing feature map for each covered location (receptive field) of the input and each layer contains multiple numbers of trainable filters. The two main parameters to modify the behaviour of each Conv layer are Stride(s) and Padding (p) and filter size. Stride controls how filters convolves around the input and padding is often use to preserve the information of original input. The size of output feature of Conv Layer can determine by Equation 1.

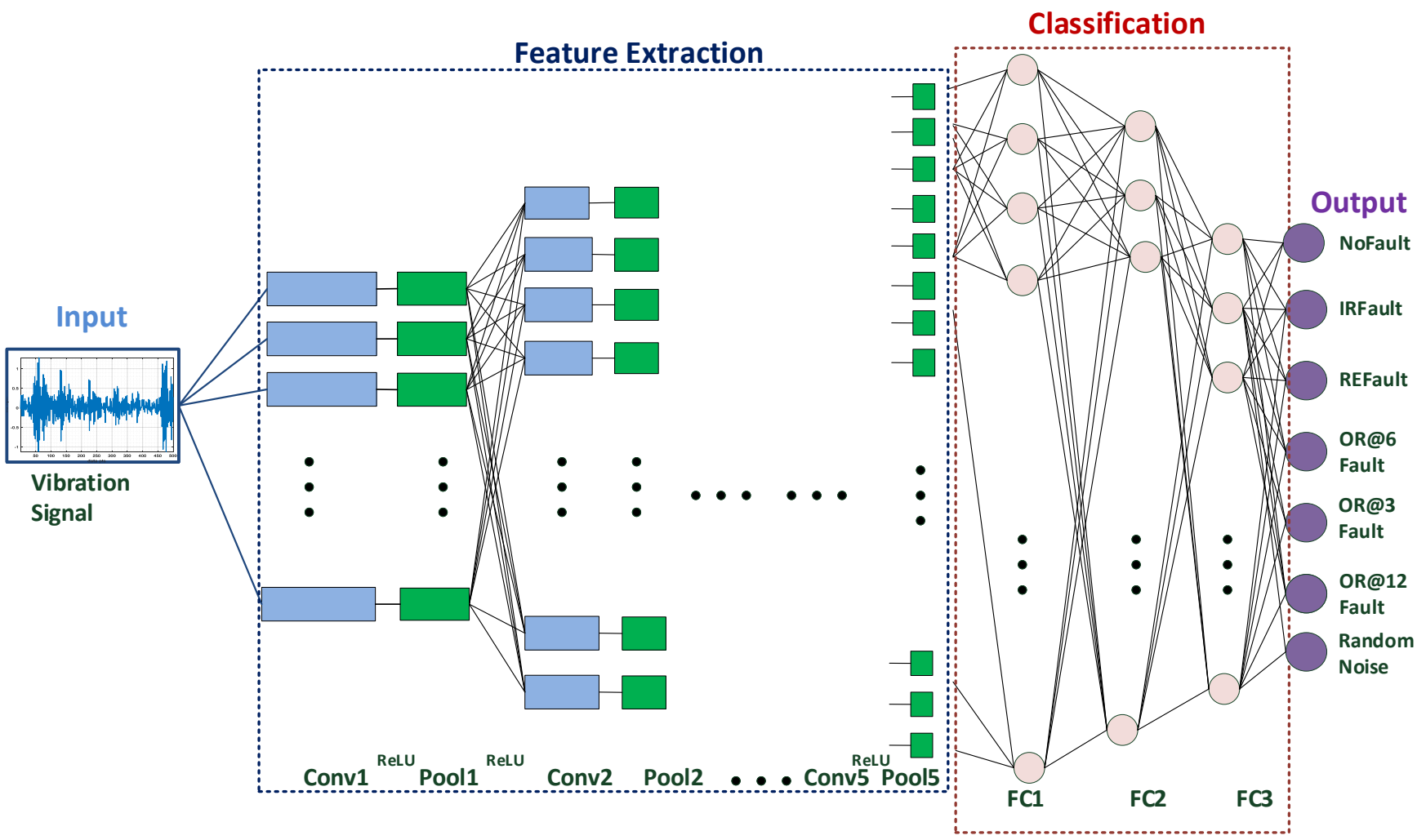

**Fig. 1.** CMCNN Architecture

$$Outout_{size} = \frac{Input_{size} + 2p - filter_{size}}{s} + 1 \tag{1}$$

Generally in image recognition problems the filter size is very small and most of the time zero padding is considered. As the target of CMCNN is to find features from raw vibration data, it does not make sense to extract a very small scaled feature and the Conv Layers should preserve original information of input as much as possible. The idea of implementing wide filters in the first convolution layer is shown in [10]. The Conv Layers of CMCNN starts with wider filters and reduced gradually in later layers. Stride and Padding are parametrized in a way that the original input information is kept as much as possible. In CMCNN the number of filters and neurons at FC layers are chosen similar as VGG16 [13].

### 3.4 Training CMCNN

The Training dataset is divided 25% for validation and 2.5% data for testing. CMCNN network training parameters are updated with Stochastic Gradient Descent with Momentum (SGDM) algorithm. Stochastic Gradient Descent (SGD) algorithm is used to train network parameter to minimize the Error Function by converging to negative gradient loss, which might oscillate to reach optimum and convergence can be very slow. Momentum can be added to reduce the oscillation. A network parameter update by SGD can be expressed by Equation 2.

$$\Theta_{i+1} = \Theta_i - \alpha \nabla E(\Theta_i) \tag{2}$$

where $i$ is the iteration number $\alpha > 1$ is the learning rate, $\Theta$ is parameter vector, $E(\Theta_i)$ is the loss function and $\alpha \nabla E(\Theta_i)$ is the gradient of the Error Function. SGD evaluates the gradient and update the parameter using subset of the training set. Training parameter update with SGDM can be expressed by Equation 3.

$$\Theta_{i+1} = \Theta_i - \alpha \nabla E(\Theta_i) + \gamma(\Theta_i - \Theta_{i+1}) \tag{3}$$

where $\gamma$ determines the contribution of the previous gradient step to the current iteration. Training performance of CMCNN is shown in 1. True Positive Rate (TPR%) and False Positive Rate (FPR%) is calculated to determine the prediction accuracy per class. 1 also include prediction accuracy of OR and IR Fault class of two new Test data (SpectrQ). SpectrQ dataset is described in later sections.

**Table 1.** Trained CMCNN 7class performance

| | Learned Class and corresponding Labels | | | | | | |
|---|---|---|---|---|---|---|---|
| | No Fault | IR Fault | RE Fault | OR@6 Fault | OR@3 Fault | OR@12 Fault | Random Noise |
| Model: CMCNN_7class<br>Mean Acc.: 94,48%<br>Mean Error: 5,51% | 0 | 1 | 2 | 3 | 4 | 5 | 6 |
| Train. Class Dist | 4943 | 4959 | 4873 | 3131 | 3342 | 2136 | 2151 |
| Val. Class Dist | 1664 | 1647 | 1711 | 1106 | 1125 | 687 | 790 |
| Train. TPR (%) | 99,98 | 90,18 | 96,80 | 89,01 | 91,46 | 95,37 | 100 |
| Train. FPR (%) | 0,02 | 9,82 | 3,20 | 10,99 | 8,54 | 4,63 | 0 |
| Val. TPR (%) | 99,94 | 89,42 | 96,54 | 88,57 | 90,62 | 93,53 | 100 |
| Val. FPR (%) | 0,06 | 10,58 | 3,46 | 11,43 | 9,38 | 6,47 | 0 |
| SpectrQuest OR TPR (%) | - | - | - | 1,25 | 0 | 72,5 | 0 |
| SpectrQuest OR FPR (%) | 14,58 | 2,5 | 9,17 | - | 0 | - | 0 |
| SpectrQuest OR TPR (%) | - | 96,67 | - | - | - | - | - |
| SpectrQuest OR FPR (%) | 0 | - | 0 | 0,42 | 0 | 2,92 | 0 |

### 3.5 Test CMCNN

In this work CMCNN model is trained and tested for a different number of class recognition using CWRU dataset; in 2 four different Models are compared.

**Table 2.** Comparison CMCNN for different number of classes and their prediction accuracy for SpetrQ Test data:

| Model | CMCNN 7 Class | CMCNN 6 Class | CMCNN 5 Class | CMCNN 4 Class |
|---|---|---|---|---|
| No Fault | 0 | 0 | ... | 0 |
| IR Fault | 1 | 1 | 1 | 1 |
| RE Fault | 2 | 2 | 2 | 2 |
| OR@6 Fault | 3 | 3 | 3 | 3 |
| OR@3 Fault | 4 | 4 | 4 | 3 |
| OR@12 Fault | 5 | 5 | 5 | 3 |
| Random Noise | 6 | ... | ... | ... |
| Mean TrainAcc (%) | 94,48 | 93,56 | 99,51 | 99,78 |
| Mean TrainErr (%) | 5,51 | 6,44 | 0,48 | 0,23 |
| Mean ValAcc (%) | 94,00 | 93,58 | 99,37 | 99,55 |
| Mean ValErr (%) | 6,00 | 6,42 | 0,63 | 0,45 |
| Mean TestAcc (%) | 94,2 | 91,99 | 99,15 | 100,00 |
| Mean TestErr (%) | 5,80 | 8,01 | 0,85 | 0,00 |
| SpectrQ ORFault PredAcc (%) | 73,75 | 81,17 | 45,83 | 0,83 |
| SpectrQ ORFault PredErr (%) | 26,25 | 18,33 | 54,16 | 98,75 |
| SpectrQ IRFault PredAcc (%) | 96,67 | 87,50 | 100 | 100 |
| SpectrQ IRFault PredErr (%) | 3,33 | 12,50 | 0 | 0 |

To check the efficiency of CMCNN, we tested this four trained CMCNN to predict OR Fault and IR Fault for a new publicly available (SpectrQ) dataset [14] produced by SepctrQuest test bench [15]. A comparison of Train Data-OR Fault and SpectrQ-OR Fault is shown in Fig. 2. SpectrQ-OR fault location is not known, so we labelled the dataset as target class '3' and checked if the CMCNN can successfully predict class as OR Fault independent where the fault location is. For this reason we trained one CMCNN model for 4 classes where we labelled all OR Fault as '3'. The result shows though the CMCNN accuracy for 4 classes is the highest, the robustness of predicting class for new test data (SpectrQ-OR Fault) is worse.

### 3.6 Feature Map Analysis

To understand and optimize the feature extraction it is required to visualize deep layer feature map. There are several approaches available to visualize the features, one is to visualize the activations of the network and other is to visualize the convolution weights. In Image recognition problems the output of each convolution layer can be interpretable because the outputs are also another image or sometimes the sharpened edges of the input image. For vibration input signals we should convert the feature output into different domain to understand. In our work we analyse the first Conv layer output by calculating it's Fast Fourier transform (FFT) and compare each filter if significant Feature Frequencies is extracted.In [10] authors also focused feature visualization with FFT and showed feature distribution for each layer and each 10 classes using Stochastic Neighbor Embedding (t-SNE). In our work, we focused on significant feature frequencies of first Conv layer for each filters and for each classes. We found that feature frequencies

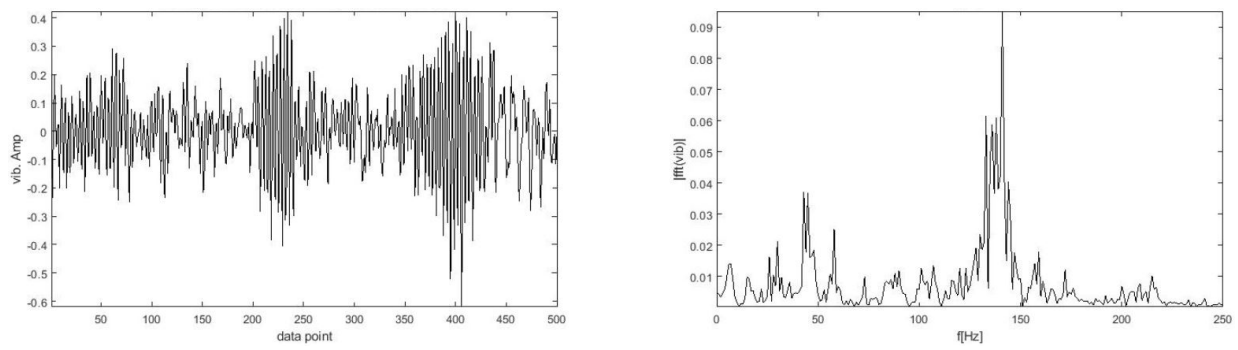

(a) Training

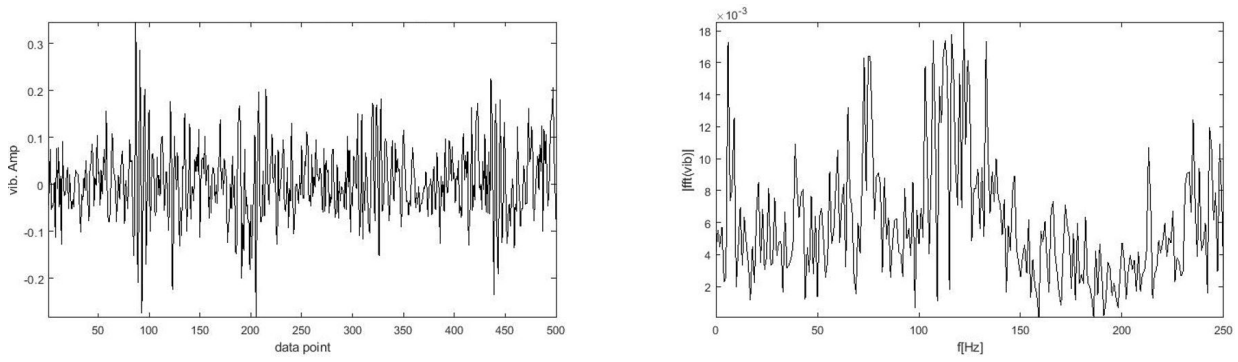

(b) Test

**Fig. 2.** Comparison of two examples of Training Data for OR Fault (a), and Test data for OR fault (b). Left: time domain data and Right: frequency domain data

are extracted in a range for each class and the extracted feature frequencies for test data (SpectQ) also in similar range of its predicted class. Figure (5) shows the comparison of feature frequencies of all 32 filters of first Conv layer of a Train Input-ORfault and Test Input SpectrQ-OR Fault.

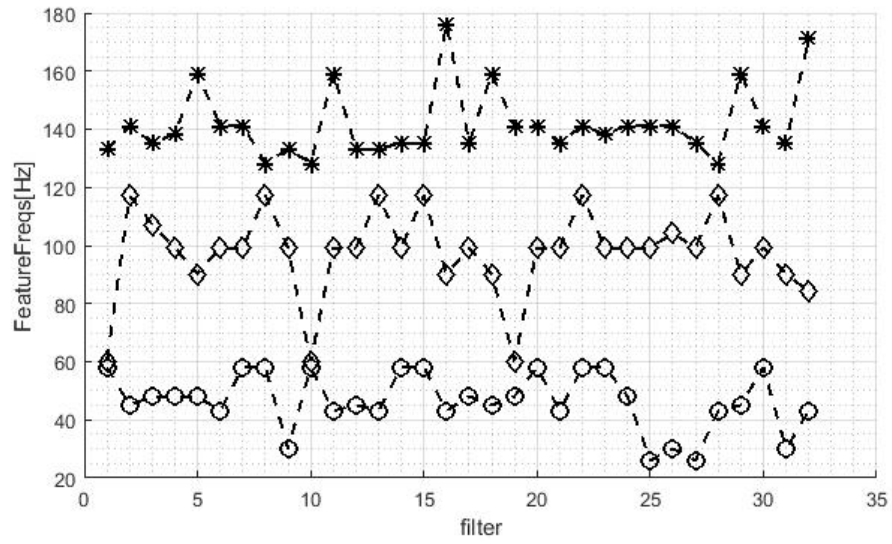

(a) Training Input

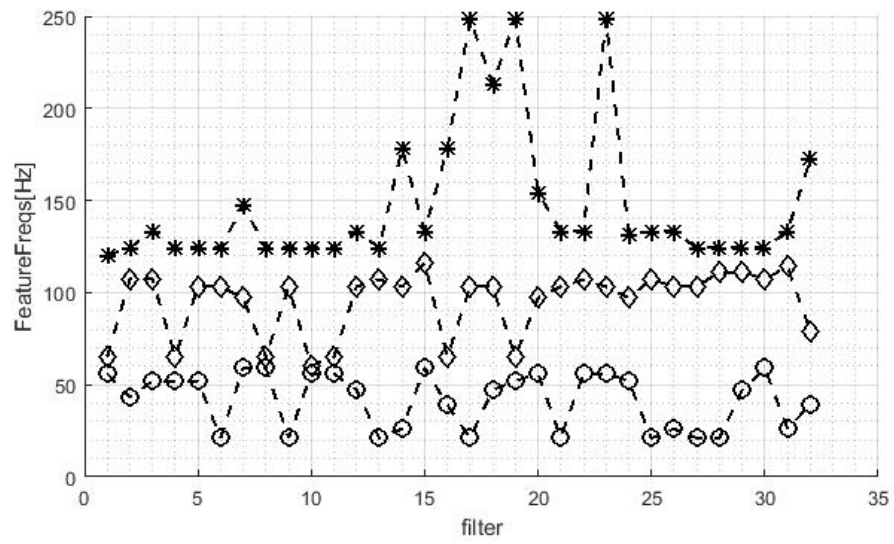

(b) Test Input

**Fig. 3.** Feature Frequencies for all trained filters of first Conv Layer for (a) Train Input: OR Fault and (b) Test Input SpectrQ: OR Fault

## 4 Conclusion

CMCNN shows good accuracy of predicting OR Fault and IR Fault for SpectrQ Test data which indicates feature extraction of CMCNN is robust for different systems. In future we will compare the filter size and input size effect on feature learning which should find the optimized architecture of the CMCNN. In our research project we aimed to implement real-life bearing faults such as surface fatigue, wear, electrical erosion, plastic deformation due to overload etc. and create dataset using IEEM-CMTestBench. Furthermore, the created dataset will be used to train the optimized CMCNN architecture to achieve robust domain adaptable feature extraction and classification of real-life fault pattern for motor bearing.

# Machine Learning Methods for State Tracking and Optimal Control of Production Processes

Norbert Link and Johannes Dornheim

Intelligent Systems Research Group, Karlsruhe University of Applied Sciences
norbert.link@hs-karlsruhe.de, johannes.dornheim@hs-karlsruhe.de

**Abstract.** Manufacturing processes act on workpieces by exerting a sequence of varying control actions. This results in a sequence of inner and outer workpiece states. The goal is to reach a final state, which has dedicated geometrical and physical properties. Variations of the input and stochastic influences must therefore be compensated during processing, while the ressource-efficiency should be maximized. For this purpose, self-optimizing Artificial Intelligence (AI) control methods were developed. The corresponding Markov Decision Problem is solved via Machine Learning methods. The cost trade-off between pre-production data sampling to learn the required models and initial low-quality production with learning from production-experience is addressed by two corresponding approaches. 1) Deep Neural auto-encoders and state trackers deliver the input of an optimizing process control, which is constructed from Approximate Dynamic Programming with integrated Neural Networks, representing the learned process dynamics. 2) An explorative AI approach with re-inforcement learning, which automatically learns an implicite model for the control policy, based on the experience with each processing result. This approach can also adapt to process drifts (e.g. from tool wear). Other than classical control methods such as Model Predictive Control, the new approaches can compensate input quality variations, stochastic state perturbations and slowly varying conditions.

**Keywords:** intelligent control, approximative dynamic programming, reinforcement learning

## 1 Introduction

The series production of parts is a repetition of processes, transforming each part from an initial state to some desired end state. Each process is executed by a finite, discrete or continuous sequence (*path*) of -in general irreversible- processing steps. Examples are plastic deformations in forming processes or material removal in milling. The optimal control of such processes is different from classical control theory problems and self-learning technologies have to come into place to adaptively solve the optimization problem under varying conditions. The process path optimization is formulated as a Markov Decision Problem (MDP) with finite horizon. The optimization strategy (*policy*) can be refined between subsequent processes, which are called *episodes* in the field of Machine Learning. This allows the adaptation of the strategy to changing process conditions, e.g. to varying state transition functions.

The episodic fixed-horizon manufacturing processes considered here are nonlinear stochastic processes. Every episode in the process consists of $T$ irreversible control steps at discrete times. Deep drawing is used as sample process throughout this paper and is depicted in fig. 1. Based on the measured quality of the process episode result, costs are assigned and transfered to the control agent by a reward signal $R_T$ at the end of each execution. The control effort (energy consumption, material use, processing time, etc.) is reflected by intermediate costs and also being subject to optimization. The goal is to find a control policy, that minimizes the total cost and thereby optimizes the process performance regarding the resulting product quality and the process effciency. The multistage optimization problem of our interest is described by a MDP and is solved by Dynamic Programming (DP) or Approximate Dynamic Programming (ADP) or by Reinforcement Learning. The DP/ADP/RL approaches in this paper involve the following concepts:

- The value or cost or reward function $J_t(\mathbf{x}_t)$ that defines the optimal value of the optimization objective (defined over the remaining time horizon) for the state $\mathbf{x}_t$ at time $t$
- In a Markov Decision Process, a transition from a precedessor state $\mathbf{x}_t$ to a successor state $\mathbf{x}_{t+1}$ taking the decision or action $u_t$ is subject to uncertainty and can be expressed by the conditional probability $P(\mathbf{x}_{t+1}|\mathbf{x}_t, u_t)$.
- The control policy $\pi_t$ that maps the current state $\mathbf{x}_t$ to the optimal action $u_{t_{opt}} = \pi_t(\mathbf{x}_t)$.

The optimization tries to select a control action, which leads to minimum expected cost in the ongoing process. It therefore requires a prediction model, which is built via machine learning and applied via AI inference. The individual model components (costs, state transition, Q-function and policy) are represented by continuous ANNs or by discrete value tables in combination with a k-Nearest Neighbor (k-NN) predictor. The mapping between the actual state $\mathbf{x}_t$ and the (optimal) action $u_{t_{opt}}$ is referred to as the (optimal) policy (i.e., control law). The decision is made based on the stagewise costs $C_t(\mathbf{x}_t, u_t)$ which depend on the selected state transition. To optimize the entire process consisting of multiple time stages, the total costs $\langle \sum_{t=1}^{T} \gamma_t C_t(\mathbf{x}_t, u_t) \rangle$ need to be considered, where $\gamma_t$ is a discount factor over time ($0 < \gamma_t \leq 1$). The result of the optimization defines the optimal decision for each considered state at every point in time. According to Bellman's Principle of Optimality [4], the multistage optimization problem can be reformulated in terms of the Bellman Equation [5]:

$$J_t(\mathbf{x}_t) = \min_{u_t \in U_t} \{ C_t(\mathbf{x}_t, u_t) + \gamma_t \langle J_{t+1}(\mathbf{x}_{t+1}) \rangle \} \tag{1}$$

with $\langle J_{t+1}(\mathbf{x}_{t+1}) \rangle = \sum_{\mathbf{x}_{t+1} \in X_{t+1})} P(\mathbf{x}_{t+1}|\mathbf{x}_t, u_t) J_{t+1}(\mathbf{x}_{t+1})$.

$J_t$ describes the costs that are associated with the current state $\mathbf{x}_t$, and $J_{t+1}$ stands for the costs that are associated with the remaining time steps and depends on the successor state $\mathbf{x}_{t+1}$ of the current decision. Since the state transition is subject to uncertainty, the expectation of $J_{t+1}$ is taken with respect to all possible successor states $X_{t+1}$ and their conditional probabilities $P(\mathbf{x}_{t+1}|\mathbf{x}_t, u_t)$. Two approaches are possible: The first learns explicit initial models of the process state dynamics and of the process state measurement in advance, based on experimental data from laboratory or simulation. The second starts building an optimization model from scratch during processing, learning directly from process data in a trial-and-error fashion. In the first case, much experimental and modelling effort is required, before production can start. In the second case, the initial production will be sub-optimal until the models are sufficiently learned. Nevertheless both AI approaches result in highly optimized production strategies, which can both compensate stochastic influences, while the second can also automatically adapt to changing input quality, to tool wear and other influences, thus maintaining continuous high product quality.

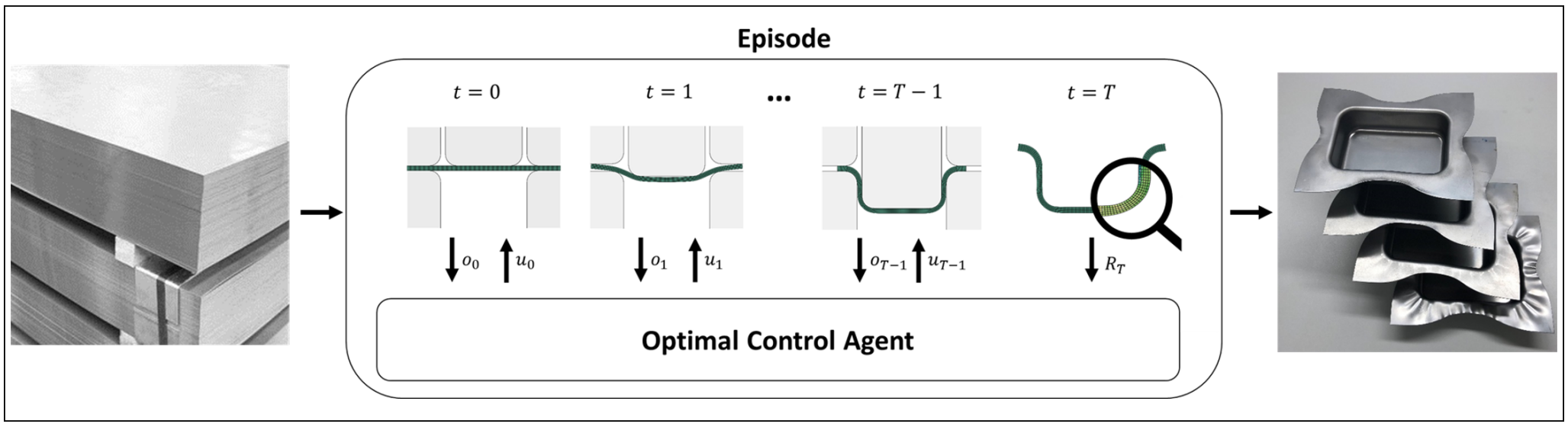

**Fig. 1.** Deep drawing optimal control as episodic fixed-horizon process. For every control step $t$, the optimal control task is to determine the control-action $u_t$ that maximizes the expected episode reward $R_T$, based on the observations $[o_0, ..., o_t]$ in the current episode and on data from previous episodes. (source: [1])

We use the optimal control of a deep drawing process with forced punch speed as application example and evaluation case. In cup deep drawing, a metal sheet is clamped between a blank holder and a die. A punch presses the sheet into the die such that we obtain a cup-shaped workpiece. If the blank holder force is chosen too high, material cracking will occur, while a too small blank holder force will lead to wrinkles in the sheet edge. This necessitates the control of the blank holder force during the execution of the deep drawing process for a finite time horizon. Besides given input properties, like the initial blank shape and thickness variation, the time-variation of the blank holder forces is crucial for the resulting process quality. The control goal is to set the blank holder force at each processing step, depending on the current, partially observable process state in a way, which yields the optimal process result. The influence of space and time variation schemes of blank holder forces on the process results is examined e.g. in [2], [3] and [4]. The state of a workpiece during processing is described by the stress distribution within the material, but which is not directly observable during processing. Only the reaction force in the punch and the intake of the material between die and blankholder can be measured. This makes the optimization a so-called *partially observable Markov decision problem* (POMDP).

## 2 Optimal control for continuous state spaces by incremental learning

The first approach presented in [5] uses model-based optimal control methods where a process model, an state-observation model and a cost function are available from previous work [6]. Optimal control is then acheived by dynamic programming [7]. Such approaches are subject to the so-called curse of dimensionality in high-dimensional state spaces, leading to diffculties from huge sample sizes required for modelling and from the resulting computational complexity. In the case of continuous (and thus infinite-dimensional) state spaces, the optimal control solution by dynamic programming requires discretization of the state space, leading to suboptimal solutions. These problems are addressed in the field of approximate dynamic programming, combining dynamic programming with function approximation [8]. In applying ADP to the problem, Artificial Neural Networks (ANNs) are created as global parametric function approximators to represent the value functions as well as the state transitions. For each time step of the finite time horizon, time-indexed function approximations are built. We use a backward ADP approach with batch learning of the ANNs. Here, the ANNs are trained from temporary value tables constructed by exhaustive search backwards in time. The obtained value function approximations are obtained with good performance, where the models for the state transition and the value function are determined with batch learning ANNs from simulation data. The control policy is given by the solution of the stochastic Bellman Equation with approximated successor costs $\tilde{J}_{t+1}$ with respect to the pre-decision state $\tilde{\mathbf{x}}_{t+1}$) (Eq. 2).

$$J_{t_{opt}}(\mathbf{x}_t) = \min_{u_t \in U_t} \{C_t(\mathbf{x}_t, u_t) + \langle \tilde{J}_{t+1}(\tilde{\mathbf{x}}_{t+1}) \rangle\} \tag{2}$$

The optimization is performed backwards in time as depicted in Fig. 2.

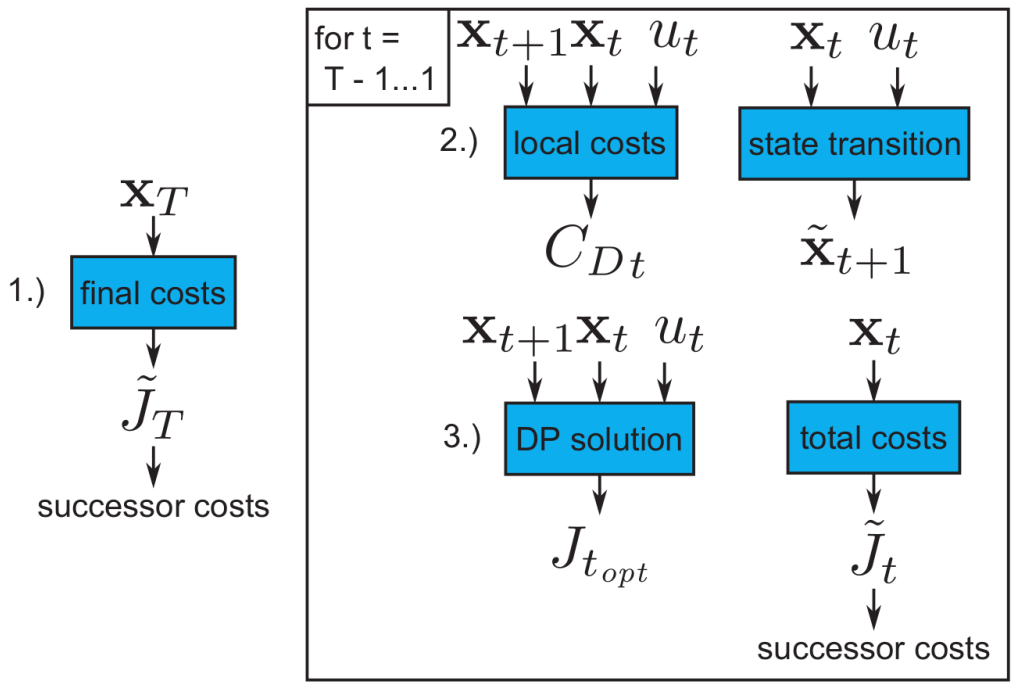

**Fig. 2.** Scheme of backward ADP (source: [5])

We start by determining the final costs $J_T = C_T = C_F$ for each final state $\mathbf{x}_T$ and create the approximation $\tilde{J}_T$ in terms of a batch learning ANN. This serves as the successor costs $\tilde{J}_{t+1}$ in the first loop run for $t = T-1$. For each loop run, we determine the local costs $C_t$ and create a batch learning ANN to predict the state transition from actual states $\mathbf{x}_t$ with decisions $u_t$ to the corresponding successor states $\tilde{\mathbf{x}}_{t+1}$. In a next step, the Bellman Equation (with approximated successor costs $\tilde{J}_{t+1}$ is solved with respect to the pre-decision state

$$J_{topt}(\mathbf{x}_t) = \max_{u_t \in U_t} \{C_t(\mathbf{x}_t, u_t) + \langle \tilde{J}_{t+1}(\tilde{\mathbf{x}}_{t+1}) \rangle\} \tag{3}$$

for each state $\mathbf{x}_t$ and the optimal solution $J_{t_{opt}}$ is stored in a temporary value table. The approximated successor costs $\tilde{J}_{t+1}$ are either given as an ANN from the previous loop run or by the approximated final costs $\tilde{J}_T$ at the beginning. The expectation $\langle \tilde{J}_{t+1} \rangle$ is calculated as a numerical integral. The content of the temporary value table then serves for training a batch learning ANN $\tilde{f}_t$ based on the states $\mathbf{x}_t$. This approximation corresponds to the successor costs in the following loop run. The Backward ADP gives an optimal value-function from which the optimal control strategy can be derived, but which is not adaptive to process drifts.

## 3 Reinforcement Learning Approach

The processes considered here are not fully observable. The sole sensor observations at a time are not suffcient for deriving optimal control decisions. Instead of learning measurement models (mappings of sensor data history on states) as in the previous approach, the accumulated sensor data and control history is taken into account to avoid any pre-production modelling.

The model-free adaptive approach, proposed in [1], requires no prior information, (like reference trajectories, a process model or an observation model), since the optimal control strategy is learned during execution. The approach can thereby be used, if no accurate process model is available or the use of the given process model for optimization is impractical.

In this approach, following the standard notation of reinforcement learning, the system and the optimization problem are modeled as an MDP, as introduced in the previous chapter. In MDPs, instead of the cost-function $J$, a reward function $R$ is used, leading to a maximization problem instead of the minimization in eq. 3. The Bellman equation is then given by the value function

$$V^*(\mathbf{x}) = \max_{u \in U} \mathbb{E}_P \Big[ R_u(\mathbf{x}, \mathbf{x}_{t+1}) + \gamma V^*(\mathbf{x}_{t+1}) \Big], \tag{4}$$

where the probability of $\mathbf{x}_{t+1}$ is given by the transition probability function $P_u(\mathbf{x}, \mathbf{x}_{t+1})$, capturing stochastic process conditions. The optimization is achieved by selecting the action, which leads to the state that maximizes the value function $V^*$. This would require knowledge of the state transition probability function. Instead, we use the optimal Q-function, assigning an expected reward value to each state-action tuple

$$Q^*(\mathbf{x}, u) = \mathbb{E}_P \Big[ R_u(\mathbf{x}, \mathbf{x}_{t+1}) + \gamma \max_{u_{t+1} \in U} Q^*(\mathbf{x}_{t+1}, u_{t+1}) \Big]. \tag{5}$$

For discrete action-spaces, the optimal policy can then be determined from the Q-function simply by

$$\pi^*(\mathbf{x}) = \arg\max_{u \in U} Q^*(\mathbf{x}, u). \tag{6}$$

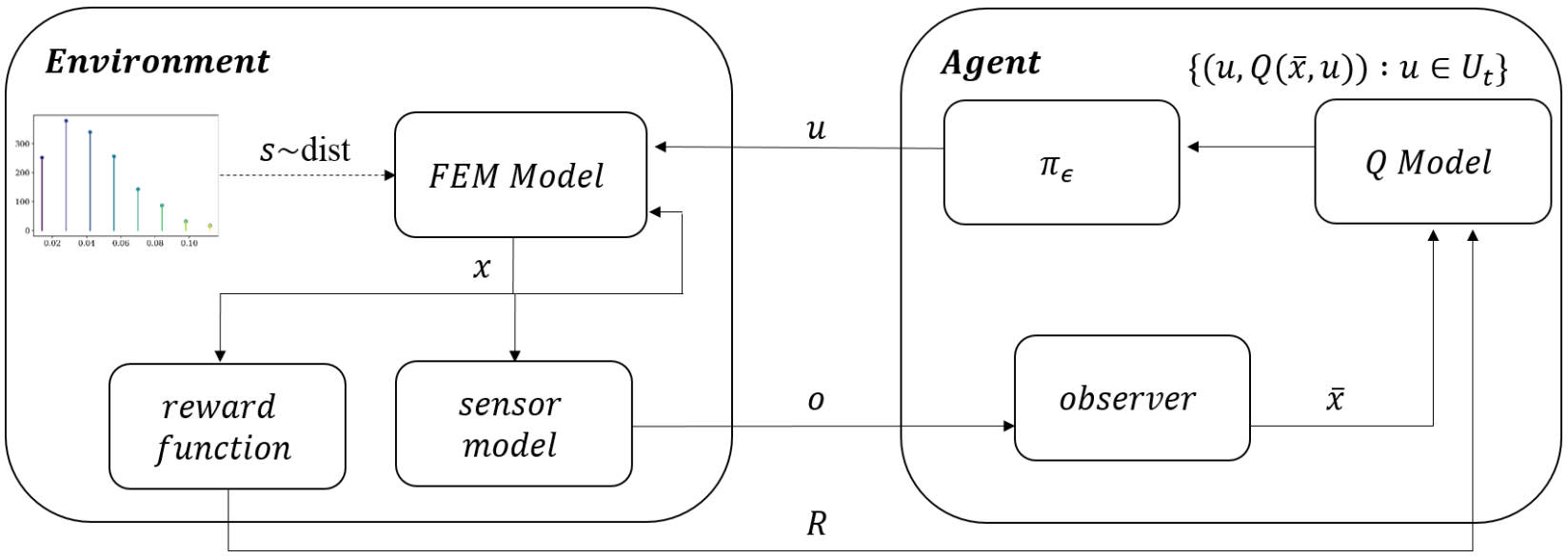

**Fig. 3.** Scheme of the interaction of the optimal online control agent with the process environment. (source: [1])

By taking the actions into account, the $Q$-function implicitly captures the system dynamics, and no additional system model is needed for optimal control. A generic version of the $Q$-learning control agent is depicted in fig. 3.

In $Q$-learning-based algorithms, $Q^*$ is found by constantly updating a $Q$-approximation by the update step in eq. 7, using experience tuples $(\mathbf{x}, u, \mathbf{x}_{t+1}, R)$ and a given learning rate $\alpha \in [0, 1]$, while interacting with the process in an explorative manner.

$$Q'(\mathbf{x}, u) = \big(1 - \alpha\big) Q(\mathbf{x}, u) + \alpha \big(R + \gamma \max_{u_{t+1} \in U} Q(\mathbf{x}_{t+1}, u_{t+1})\big) \tag{7}$$

An observer derives surrogate state descriptions $\bar{\mathbf{x}}$ from the observable values $o$, and previous control actions $u$. Control actions are determined based on a policy $\pi$, which itself is derived from a $Q$-function. In the approach, proposed in [1], the $Q$-function is learned from the processing samples via batch-wise retraining of the respective function approximation, following the incremental variant of the neural fitted Q iteration approach ([9]). For exploration, an $\epsilon$-greedy policy is used, acting randomly in an $\epsilon$-fraction of control actions. To derive the optimal action, the current approximation of the $Q$-function is used (exploitation). The exploration factor $\epsilon \in [0, 1]$ is decreased over time to improve the optimization convergence and to reduce the number of sub-optimal control trials. We use an exponential decay over the episodes $i$ according to $\epsilon_i = e^{-\lambda i}$ , with decay rate $\lambda$. Since $x$ is only partially observable, we use the full information about observables and actions for the current episode by concatenating all these values into a surrogate state description $\bar{\mathbf{x}}$. Thus, the dimension of $\bar{\mathbf{x}} \in R^n$ is time-dependent according to $n_t = [\dim(O) + \dim(U)] * t$. When using $Q$-function approximation, the approximation model input dimension is therefore also dependent on $t$. If function approximation methods with fixed input dimensions (like standard artificial neural networks) are used, a dedicated model for each control step is required. The extension of the model-free approach for applications with multiple weighted and eventually contrary objectives is described in [10].

## 4 Results and Conclusion

Both approaches, proposed in [5] and [1], are applied to the standard industrial process of cup deep drawing (soda cans, machine covers, pots, etc.) with volumes of millions of batches per day. The process goal is to produce a cup with low internal stress and low material usage, but with sufficient material thickness. The three optimization criteria are combined into a single Reward value by calculating the weighted harmonic mean of the minimum wall thickness, the negative residual stress and the left-over material. The weight values reflect the emphasis, which is given to the respective goals. The first approach requires requires a high effort in sampling experimental data and training of the models, before it can be applied to the process, but produces nearly optimal results from scratch despite being subject to processing noise. The second approach requires no sampling or training before the start of production. The price is lower performance in the early phase of production, while the models are learnt. It was shown in [1], that in the case of variantions in the friction between material and matrix/blank-holder/punch during processing, which occur due to roughness variations of the sheet material, the second approch learnt to cope with the variations and out-performed the first approach by 20% on the long run.

## 5 Acknowledgements

The authors would like to thank the DFG and the German Federal Ministry of Education and Research (BMBF) for funding the presented work carried out within the Research Training Group 1483 "Process chains in manufacturing" (DFG) and under grant #03FH061PX5 (BMBF).

# Current regulatory issues of AI in the field of medical technology

Martin Haimerl[1 2]

[1] Hochschule Furtwangen University (HFU)
[2] Innovation and Research Center (IFC) Tuttlingen of the HFU
martin.haimerl@hs-furtwangen.de

**Abstract.** This paper describes current issues regarding regulatory requirements in medical devices with a focus on data-driven / AI based approaches. It shows that the EU Medical Device Regulation (MDR) sets high requirements to assess product performance based on systematically collected data, whereas the collection of data is difficult in the EU. Contrary, it demonstrates that the FDA is currently very active in supporting the development of software based systems in the US with dedicated regulatory programs. In particular, it pursues more dynamic approaches for releasing software devices. The overall situation favors developments in the US. Thus, the paper surveys a program to support local entities on adapting AI technologies.

**Keywords:** regulatory requirements, Medical Device Regulation (MDR), FDA, data-driven approaches, artificial intelligence

## 1 Introduction

Artificial intelligence (AI) is considered to have a huge potential in the field of medicine and medical devices. This includes areas like automated diagnosis of pathologies or personalized therapies using individual patient data from different sources. Additionally, optimization of processes in the development, production, and marketing of devices based on systematically acquired business data may substantially benefit from AI technologies.

But, medical devices also have wide-ranging regulatory requirements, e.g. regarding performance, reliability and traceability throughout the entire life-cycle of the device. On the one hand, these requirements set limitations on a rapid availability of AI based products. On the other hand, regulations may/should be a key to ensure better control and sustainability of the development. Already highly regulated domains like medical devices may be important to establish appropriate rules for this.

The main objective of this paper is to clarify the applicable regulations and to describe challenges and opportunities for AI based devices. In a first step, the requirements and current trends in regulatory processes will be analyzed which apply to AI technologies. In a second step, a way to enforce discussions about and develop best practices is outlined, which aims to improve competencies at local entities.

## 2 Current evolution of the regulatory environment

The following section delineates the current regulatory landscape and development trends w.r.t. software and AI based medical devices. It focuses on a comparison between the European Union (EU) and the United States (US).

## 2.1 Current regulatory activities in the EU

In the EU, the Medical Device Regulation (MDR) [1] is the main reference regarding regulatory requirements for medical devices. It became effective in May 2017 and will get compulsory in May 2020. The MDR includes software in terms of stand-alone systems as well as software components. Thus, it also is the main reference for using AI in the field of medical devices or medicine in general.

However, the development of the MDR was mainly driven by major issues caused by medical implants (e.g. breast and hip implants). There was less focus on devices, which are based on software. Specific rules or recommendations for such devices were only addressed in a limited way. Additional regulations or guidances are not yet available to the best of the author's knowledge. In particular, this applies to AI based methods which do not have to be considered as fixed devices, but as adaptive systems that dynamically change according to additionally provided data.

In the development of the MDR, an evidence based approach to perform clinical validation as well as a comprehensive post-market surveillance were two of the pivotal points. According to the MDR, clinical validation has to be based on a systematic analysis of clinical data which prove the success of the devices. These data are considered to be provided through high-level clinical studies. Post-market surveillance includes an assessment about the entire chain of the development, production, and application of the device in the field. This requires comprehensive collection and analysis of data in a dedicated evaluation process. The usage of AI technologies seems to be a natural fit to address these requirements. A high-level approach w.r.t. data analysis may be considered as a crucial point in the implementation of the MDR.

In contrast, the EU General Data Protection Regulation (GDPR) [2] sets high restrictions on collecting data, which are necessary to pursue an evidence based approach for assessing the performance of medical treatments. Besides the right for protection of personal data, there is a certain need for the availability of clinical data to achieve this. In some aspects, medical applications require special rules, which have to be carefully balanced between personal rights and public demands.

In general, the GDPR sets substantial impediments for developing innovative applications using AI in the EU. The EU considers AI as a field, where the development should be more deliberatively controlled to achieve sustainability. See also the recently published EU guideline for trustworthy AI [3], which tries to establish best practices to include ethical aspects into AI development. In [3], medical applications are mentioned but the considerations were not aligned with the MDR, e.g. regarding high-level standards on clinical validation.

In summary, there is a substantial discrepancy between the high requirements on the one hand and the hurdles and uncertainties regarding the practical implementation on the other hand. Further clarification is important to develop appropriate ways for the evolution of requirements and best practices.

## 2.2 Current regulatory activities in the US

While the relevant players in the EU are still struggling with a clear interpretation and implementation of the MDR rules, the situation in the US is much different. There is considerable effort to develop an appropriate regulatory environment, which fosters software and AI technologies.

In general in the US, there is a substantially higher willingness to provide data for the development of AI based technologies as well as a better established knowledge how to implement and market AI systems. This includes an agile mind set which considers product development as a dynamic learning based approach, that incrementally improves product performance. On the one hand, this attitude and environment matches very well with the basic goals of the MDR in terms of a continuously applied evaluation process. On the other hand, less restrictions are placed w.r.t. development and implementation of data-driven applications including AI methods.

The US Food & Drug Administration (FDA) as the regulatory authority for medical devices in the US is currently proactively supporting the development of software and AI technologies. It recently published a series of of guidances and programs which are dedicated to this. This includes guidances about "Clinical and patient decision support software" [4], which clarifies the status of software tools preparing information for clinical decision making, as well as "The use or real-world evidence to support regulatory decision-making for medical devices" [5]. The latter provides opportunities to derive clinical validation data from real world scenarios and release medical devices based on this without the necessity of setting up dedicated clinical studies. Real world data are considered to be any data related to patient health status that could e.g. be collected from electronic health records, product and disease registries, or patient-generated data collected from in-home-use settings.

Basic FDA strategies to foster the development of software devices are described in the "Digital health innovation action plan" [6] as well as the "Digital health software precertification (Pre-Cert) program" [7]. In the latter, the FDA tries to establish a regulatory process, which can be considered as a dynamic systems adapting its rules according to the assessment of its own performance. Thus, it realizes a learning-based approach, which is capable to dynamically react on changes in the development landscape. In particular, this addresses the quicker development cycles which usually occur in software devices. This is in major contrast to the development of the MDR which took approximately ten years to define a rather strict and extensive set of rules for governing the regulatory pathways.

### 2.3 AI based technologies as a dynamic process

These endeavors show that the FDA considers data-driven approaches and dynamic development processes as an important way to improve not only patient health but also opportunities for software-based devices. In general, AI systems have to be considered as dynamic systems. They are designed to get improved when providing additional data. Thus, AI based devices are not intended to have a fixed status, but are continuously changing in nature.

In contrast, medical device regulations require a fixed status for the release. This status has to be comprehensively validated before release. This would require a re-release of the device each time new data come in. This inhibits continuous changes of the data base during application. A validation and release of the entire process (incl. on-site data collection, training, and validation) would be necessary (see fig. 1). This is not yet covered in the regulations.

In principle, such a process-oriented validation not only applies to products but also to other software based tools within the quality system of an organization. Thus, it is also applicable for the continuous evaluation of medical devices, as required by the MDR, when they are addressed based on AI tools.

The FDA currently takes basic steps towards such a strategy in its pre-cert program [7]. It allows companies to quickly release medical devices once the validity of basic development steps have been proven. The final step, to allow continuous validation during the application phase of hr product is not covered.

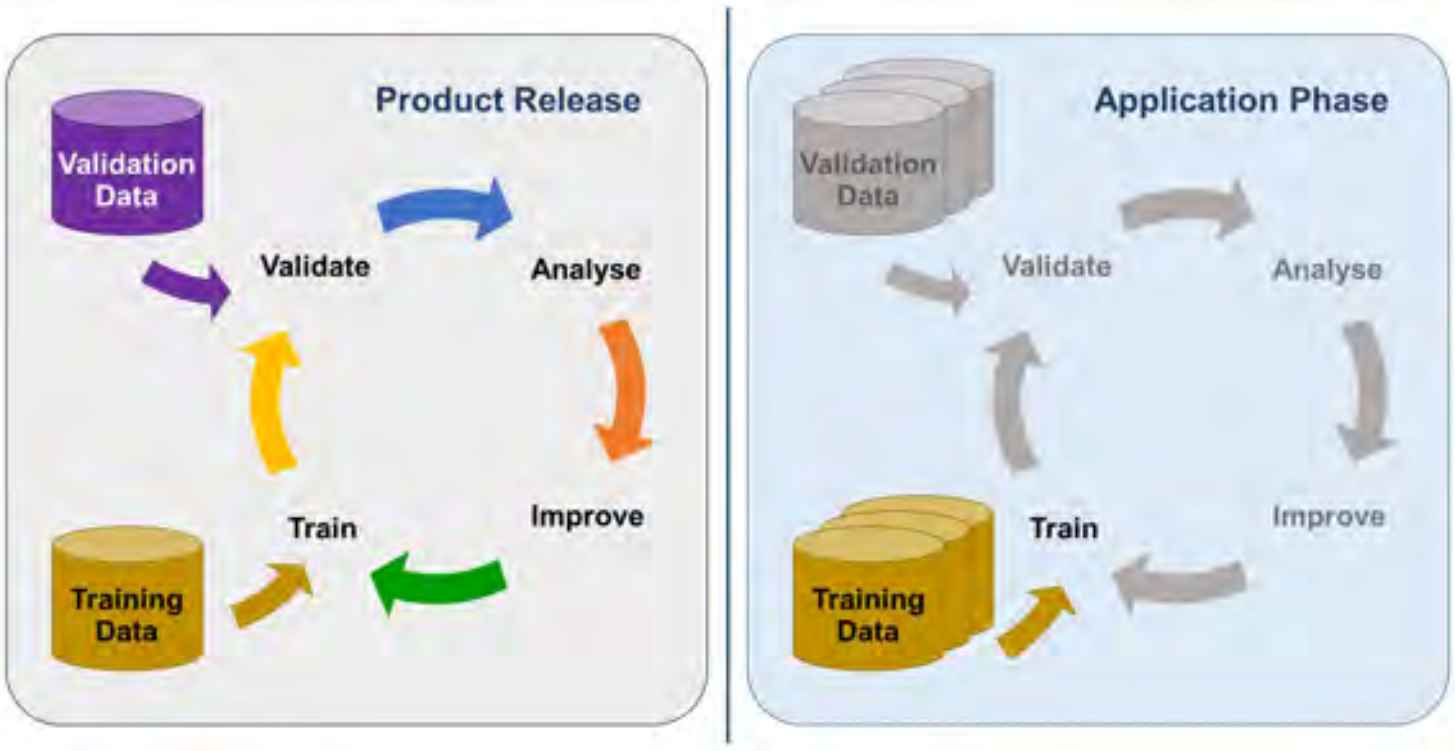

**Fig. 1.** Development cycles of AI based devices. On the left, a standard development cycle is shown which leads to the release of the device. On the right, a dynamic increment of the data basis is sketched, which would require re-releases of the device or a process-based validation of the overall system.

## 3 Conclusion

In summary, there are unclarified issues in the regulatory landscape for AI based medical technologies, including product as well as process development steps. Additionally, there are major differences between EU and US, which can have a substantial impact on the progress in each area. This is intensified by the MDR, which requires a substantially data-driven approach to assess the performance of the devices.

This discrepancy may compromise the currently still strong position of EU companies in this field in comparison to US counterparts. A similar situation was observed in the second half of the 20th century in the pharma industry when the standards for developing drugs were substantially raised by requiring comprehensive validation in terms of clinical studies before releasing a product. This resulted in a substantial shift of major players towards the US.

Thus, strategies should be developed to overcome these issues. On the one hand, practical hurdles should be eliminated and requirements should be clarified. On the other hand, a high quality approach should be pursued which keeps a major focus on reliability and sustainability of the development.

## 4 Outlook

At the Hochschule Furtwangen University (HFU), this is intended to be addressed in a combination of educational and research activities. It is considered that such a development needs a close cooperation between the academic and the industry side. This will start with general informative talks which address a broad audience before it is continued in smaller workshops and expert tables. Detailed discussions about common issues and opportunities are pursued in these smaller groups to develop best practices for addressing integration of AI methods in compliance with regulatory requirements. The final objective is to transfer particular activities into dedicated research projects.

Currently, this program is not directly part of a funded research program. It is considered as part of the transfer strategy. The Innovation and Research Centre (IFC) Tuttlingen is the central institution where such cooperation oriented programs are implemented at the HFU. The IFC's mission is to provide impulses for a transformation of the local industry towards innovative technologies.

# Machine Learning: Menschen Lernen Maschinelles Lernen

Ralph Isenmann, Jürgen Prinzbach, Stephan Trahasch, Volker Sänger, Tobias Lauer, Tobias Hagen, and Klaus Dorer

Institute for Machine Learning and Analytics
Offenburg University of Applied Sciences
{ralph.isenmann/ juergen.prinzbach/ stephan.trahasch/ volker.saenger/ tobias.lauer/ tobias.hagen/ klaus.dorer} @hs-offenburg.de

**Abstract.** This paper describes the concept and some results of the project "Menschen Lernen Maschinelles Lernen" (Humans Learn Machine Learning, ML2)[1] of the University of Applied Sciences Offenburg. It brings together students of different courses of study and practitioners from companies on the subject of Machine Learning. A mixture of blended learning and practical projects ensures a tight coupling of machine learning theory and application. The paper details the phases of ML2 and mentions two successful example projects.

**Keywords:** machine learning, data science, blended learning

## 1 General Description

Machine learning is a multidisciplinary field of study that is increasingly relevant to a wide range of practical applications. Unfortunately, company employees usually do not have the necessary knowledge to use machine learning in their respective areas, while computer scientists, for example, do not have the necessary domain knowledge [1]. This problem is addressed by the interdisciplinary research project ML2. It is funded by the Federal Ministry of Education and Research (BMBF) and implemented by the Institute for Machine Learning and Analytics (IMLA)[2] at Offenburg University of Applied Sciences. The central research question here is how to raise the potential of machine learning for medium-sized companies and at the same time ensure application-oriented education for the students of the university. The answer is a new teaching and learning concept, which has been specially developed for both target groups within the framework of this project and provides for two blended learning phases. These can optimally link theory and practice as well as the learning locations university and company as the later place of work for the students.

## 2 Project Objective

In addition to the basic subjects of computer science and mathematics, concrete machine learning projects always include an area of application as a third discipline. Basically all areas in which large amounts of complex data accumulate can be relevant. However, especially for the respective domain experts, who have little computer science or mathematical background, the field of machine learning is often unknown. At the same time, it is precisely these users who can identify those areas in which machine learning can offer new approaches to solving problems.

[1] `https://ml2.hs-offenburg.de`
[2] `https://imla.hs-offenburg.de`

The primary goal must therefore be to resolve this dilemma by training experts from companies in machine learning, for example in industry 4.0, trade and logistics or social media. They are addressed in particular because the project was preceded by the assumption that new problems and applications of machine learning are generally not technology-driven. Rather, the users know the problems hidden in the data as well as possible solutions. It is therefore important for them to understand the thinking and approach of machine learning and to expand their understanding of its functions and effects through targeted qualification.

In addition to employees from companies, students from Bachelor's and Master's degree programs are also trained to study in various fields of application that are related to machine learning. As graduates, these students can ultimately transfer their knowledge and skills to companies and thus play an important multiplier role in solving problems through machine learning. The innovative core of the project, however, lies not only in the close integration of theory and applications, but also in the joint qualification of the target groups mentioned. This enables interdisciplinary work across subject, course and department boundaries in order to develop ideas for intelligent data products and services and implement them with the help of machine learning. A further goal is therefore imparting the professional and cross- disciplinary competences.

## 3 Qualification Concept

The aim of the ML2 qualification program is to qualify company employees and HSO students in the field of machine learning. The two target groups require and enable interdisciplinary work across subject, course and department boundaries in order to develop ideas for intelligent data products and services and implement them with the help of machine learning. The program lasts nine months and is divided into a theoretical phase "Machine Learning for Practice" and a practical phase "Machine Learning in Practice" (see Figure 1). The figure reflects the general structure of the program.

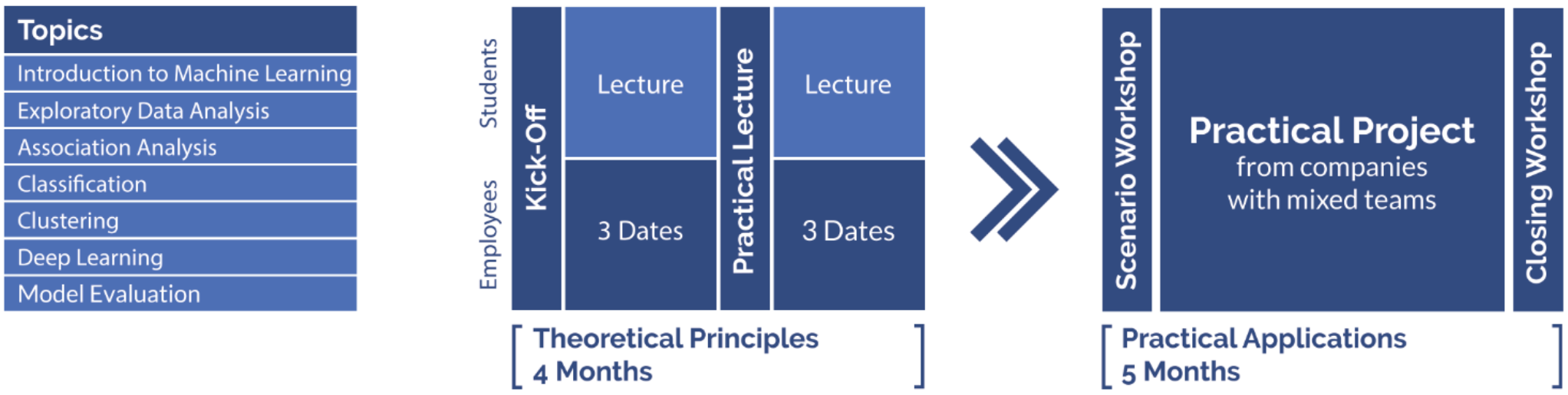

**Fig. 1.** Project Stages

### 3.1 Phase 1: Machine Learning for Practice

In the first qualification section, the basics and procedures of machine learning for practical use are taught. The aim is that at the end of the first phase, company employees and students have comparable machine learning skills and a common frame of reference and level of knowledge. Since the prerequisites and the time budget of the two target groups are heterogeneous, the first qualification phase for the users and the students takes place in different qualification paths with target group-specific learning content that takes the different prerequisites into account. In the following, the paths are referred to as student and employee tracks.

In the student track, an interdisciplinary course "Machine Learning" consisting of a lecture and laboratory/exercises is organized. In this course, students learn the same content as employees in the employee track, whereby the student track also deals with some advanced content and more strongly with theory. The students conclude this phase with an examination according to the study and examination regulations.

At the same time, a qualification on machine learning also takes place in the employee track. The qualification concept for employees is based on a blended learning approach [2] that combines the advantages of classroom-based courses with digital forms of learning that are independent of time and place. It consists of a kick-off meeting and a subsequent phase with interlinked on-line training and face-to-face meetings as well as a final meeting. The on-line phases enable participants to acquire knowledge flexibly in terms of time and place. The intermediate presence events set the pace and support the learning processes through the professional exchange between the participants and the teachers.

In lectures given by data scientists from industry and research, the participants and students can gain further experience already in the theoretical phase and combine it with learned knowledge. At the same time, they provide an opportunity for exchange and networking with the lecturers.

### 3.2 Phase 2: Machine Learning in Practice

In the practical phase, the linking of theory and practice as well as the linking of the learning locations university and company as the later place of work is of particular relevance for the students. The practical phase involves the transfer and deepening of the participants' machine learning expertise by applying and developing the knowledge acquired in the first phase in a professional and entrepreneurial context. It is necessary that the participants apply the Machine Learning know-how according to the situation and adapt it to the existing conditions of practice and to company goals. For example, a method that is best suited in theory might not be applicable in practice due to the amount of data, its nature or availability, or due to the time or resource budget.

At the end of the first phase, a selection of practice-relevant projects from companies of the participating employees will be available, which will be processed in this second phase. A presentation of these projects will take place at a final attendance date where both the employees and the students will be present. This allows participants and students to form preferences as to which projects they would like to work on. Thus, interdisciplinary teams are formed for the practical phase. The lecturers determine in a final vote which projects will be carried out and how the teams will be formed, taking preference into account. The practical phase starts with a classroom event in which the teams meet for the first time and discuss the course and the milestones of the project.

The practical phase is carried out agilely in four sprints of four weeks each, which are based on the CRISP-DM [3] standard. Each sprint begins with the development of the goals of the sprint with regard to the overall project goal from a business perspective (Business Understanding), followed by data acquisition (Data Understanding) and its preparation (Data Preparation). In the next step, a data modeling technique is selected (modeling), applied to the data, the results carefully evaluated (evaluation) and presented to the company (deployment).

The aim of the first sprint is to be able to decide whether the desired project objective can be achieved within the framework of the qualification program. If this is not the case, the experience gained and the remaining three sprints will allow sufficient time to adjust the objective. The further sprints serve to broaden the database, to refine the preparation of the data or to apply and evaluate several different machine learning methods. The practical phase ends with a final workshop at which all project teams discuss their project status. This allows experience to be gained from all projects across the board.

A special aspect of the practical phase is that not only machine learning algorithms are applied, but also the typical problems during the project work occur, such as availability and quality of the data. In addition, the applicability of the algorithms can be better experienced and learned in a real project. In a real project, a statement can also be made about the economic efficiency, which is an essential success factor.

## 4 Practical Implementation

Already in the first round, which took place between March 2018 and January 2019, 26 employees from 17 different companies as well as 19 students of the University of Applied Sciences Offenburg took part. They were able to apply the skills they had acquired in the theoretical phase in 13 different projects in the practical phase. Even though there were some hurdles to overcome and the hoped-for goals could not always be achieved in all cases, the projects led to new insights and experiences. Two projects and their results are presented below.

### 4.1 Lead Classification of Test Users

During the test phase of a software product, it would be good if the users could be evaluated on the basis of their behavior and thus better understood. The question of whether the user becomes a customer is essential. The aim of the project was therefore to use machine learning methods to identify users who are likely to purchase the software and then to use individual measures such as on-board e-mails and calls as well as discount campaigns. This was achieved by creating a prototype based on Python and Amazon Web Services that first excludes the data relevant for classification from the database and then classifies different potentials from low to high using a previously learned model. The results are then made available to the employees responsible for marketing campaigns in order to support them in their work. The prototype is currently being tested and, if successful, further developed into a productive system.

### 4.2 Advertising Media Framing

In general, the business case is about providing the user with promotional prices from various advertising media. The project was specifically about automatically recognizing sales, prices and advertising texts in previously scanned advertising brochures. For this purpose, Deep Learning methods are used to automatically draw frames around the characteristics to be classified. In concrete terms, manually framed and classified advertising media have already been used to further train the existing MobileNet SSD network from Google with the help of the Machine Learning Framework Tensorflow. The learned network could thus be used to localize and classify the already mentioned features in new advertising media by means of object detection. During the activities, numerous additional insights were gained. These include, for example, that a lower resolution of the images can significantly accelerate the training of the networks while maintaining the same quality. Finally, it can be stated that all previously defined goals have been achieved.

## 5 Conclusions

In summary, it can be said that the interdisciplinary research project "Menschen Lernen Maschinelles Lernen" achieved the desired goals and enabled the close interlocking of theory and applications in a joint qualification of company employees and students. This means that interdisciplinary work is carried out across disciplines, courses and departments in order to develop ideas for intelligent products and services, including machine learning. The experience gained in this way helps to sensitize and inspire the participants to the subject area. This success is currently to be repeated in a second round of the project and will lead again to exciting projects.

This research has been funded by the Federal Ministry of Education and Research of Germany in the framework of *IKT 2020 - Forschung für Innovationen* (project number 01IS17080).

# Object Detection for the Audi Autonomous Driving Cup

Felix Wagner, Christoph Lehmann, and Klaus Dorer

Hochschule Offenburg
{felix.wagner/christoph.lehmann/klaus.dorer}@hs-offenburg.de

**Abstract.** One of the challenges for autonomous driving in general is to detect objects in the car's camera images. In the Audi Autonomous Driving Cup (AADC)[1], among those objects are other cars, adult and child pedestrians and emergency vehicle lighting. We show that with recent deep learning networks we are able to detect these objects reliably on the limited Hardware of the model cars. Also, the same deep network is used to detect road features like mid lines, stop lines and even complete crossings. Best results are achieved using Faster R-CNN with Inception v2 showing an overall accuracy of 0.84 at 7 Hz.

**Keywords:** object detection, deep learning, autonomous driving

## 1 Introduction

Since 2015, the German car manufacturer Audi runs a competition in the area of autonomous driving, in which student teams compete in autonomous driving challenges. One of the challenges is to detect objects in the car's camera images while driving. Among those objects are other cars, adult and child pedestrians for which barbie dolls are used to fit to the scale of the cars and, new in 2018, blue emergency vehicle lighting. The cars are instructed to drive significantly slower, if a child is in the proximity of the road. They have to stop if a pedestrian wants to cross at a pedestrian crossing. Also, the cars have to drive to the right side of the street and stop if they detect an emergency car with flashing emergency light, or give way to them on crossings.

To detect these objects, the cars are equipped with a Basler daA1280-54uc wide angle front camera (170 °) with a resolution of 1280 x 960px at 45Hz. A GeForce GTX 1050Ti on board can be used for image processing. The ceiling and bonnet in the images are cut off in this work to a final resolution of 1240 x 330px.

## 2 Object Detection

To determine the best solution for object detection, several deep learning techniques were tested. There are two kinds of approaches for object detection. Candidates for one stage approaches are SSD and YoloV3. Two stage approaches are R-FCN, R-CNN and it's optimizations Fast R-CNN and Faster R-CNN.

The Single Shot MultiBox Detector (SSD) is based on the VGG-16 network [1] which is slightly modified. The architecture can be explained in three parts by its name. Single Shot describes the task of object localization and classification in a single forward pass through the network. MultiBox is the technique to determine the bounding boxes. The Detector detects objects and does a classification of those [2].

---
[1] https://www.audi-autonomous-driving-cup.com

The basic approach of detection in the You Only Look Once (YOLO) architecture is based on a fully convolutional neural network which predicts the bounding boxes and classification of the image in one stage [3]. Version 3 of this architecture (YoloV3) is an optimization with the goal to reach a better accuracy sacrificing performance. The improved accuracy is based on three different sized detection kernels in the network instead of one [4].

The R-CNN (Region-based Convolutional Network) is a two stage object detection architecture. In stage one, the image is separated in 2000 region proposals. For every region, a convolutional neural network extracts a 4096-dimensional feature vector. In stage two, a Support Vector Machine (SVM) is used to calculate the probability of a searched object in each feature vector [5]. Fast R-CNN is an improvement of R-CNN. The CNN is now getting the input image instead of region proposals and generating a convolutional feature map. In the next step, a RoI pooling layer extracts the regions, which are finally passed to fully connected layers for classification and bounding box regression [6]. Both approaches are based on selective search to find the region proposals. This time-consuming process is eliminated in the Faster R-CNN architecture. Instead of selective search, a separate network is used to learn and predict the regions [7].

The Region-based Fully Convolutional Network (R-FCN) works similarly to the R-CNN approaches in finding the region proposals. The main difference is the removed fully connected layer after the RoI pooling. Score maps are generated before the RoI Pooling layer. After the pooling, classification and localization are based on average voting which results in a faster architecture than Faster R-CNN [8].

Every mentioned object detector has an underlying feature extractor, a deep neural network. There are several architectures of networks which can be used. The ResNet architecture is characterized by the fact that the layers are not only feeding the following layer, but each layer also has connections to layers with 2 to 3 hops distance. Blocks with this property are called residual blocks. Due to the fact that the optimal number of layers required in the network isn't known, this method results in the ability of the network to skip layers during training that do not add value to the overall accuracy [9].

Another architecture is called Inception. One problem in designing of neural networks is choosing the right kernel size. Instead of making the network deeper to fit this problem, the inception architecture uses filters of multiple sizes on the same level and makes the network wider [10]. Inception v2 is intended to upgrade the accuracy and reduce the computational complexity by avoiding to alter the dimensions of the input and using smart factorization methods [11].

## 3 Results

A set of 15.343 manually labeled images or synthetically generated and labeled images [12] were used to train the networks pre-trained on coco[2] or kitti[3]. The images contain 38791 objects with a class distribution shown in Figure 1.

Table 1 shows the results of the combinations of region proposal algorithm and feature extractor evaluated. The best precision is achieved using R-FCN with ResNet101 achieving 0.86 mean average precision on at least 0.5 intersection over union (mAP@0.5IoU). At the slow speeds the model cars are driving, the 4.8 fps are acceptable. The best result due to its higher performance of 7 fps with an almost equal precision of 0.84 was achieved using Faster R-CNN with Inception v2. An example image is shown in Figure 2.

[2] http://cocodataset.org

[3] http://www.cvlibs.net/datasets/kitti/

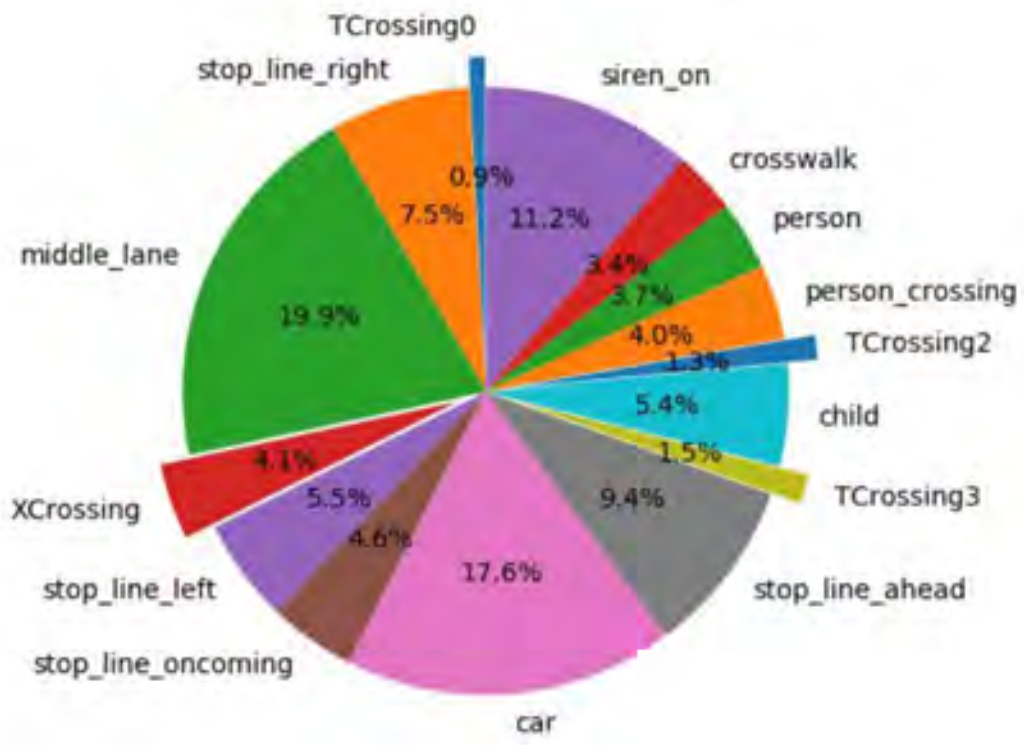

**Fig. 1.** Distribution of input images by class [13]

**Table 1.** Results of different deep learning networks [13]

| Detector | Feature Extractor | max proposals | mAP@0.5IoU | FPS |
|---|---|---|---|---|
| Faster R-CNN | ResNet101 | 300 | 0.84 | 2.1 |
| Faster R-CNN | ResNet101 | 200 | 0.83 | 2.6 |
| Faster R-CNN | ResNet101 | 100 | 0.70 | 3.8 |
| R-FCN | ResNet101 | - | 0.86 | 4.8 |
| Faster R-CNN | ResNet50 | 300 | 0.84 | 3.7 |
| Faster R-CNN | Inception v2 | - | 0.84 | 7.0 |

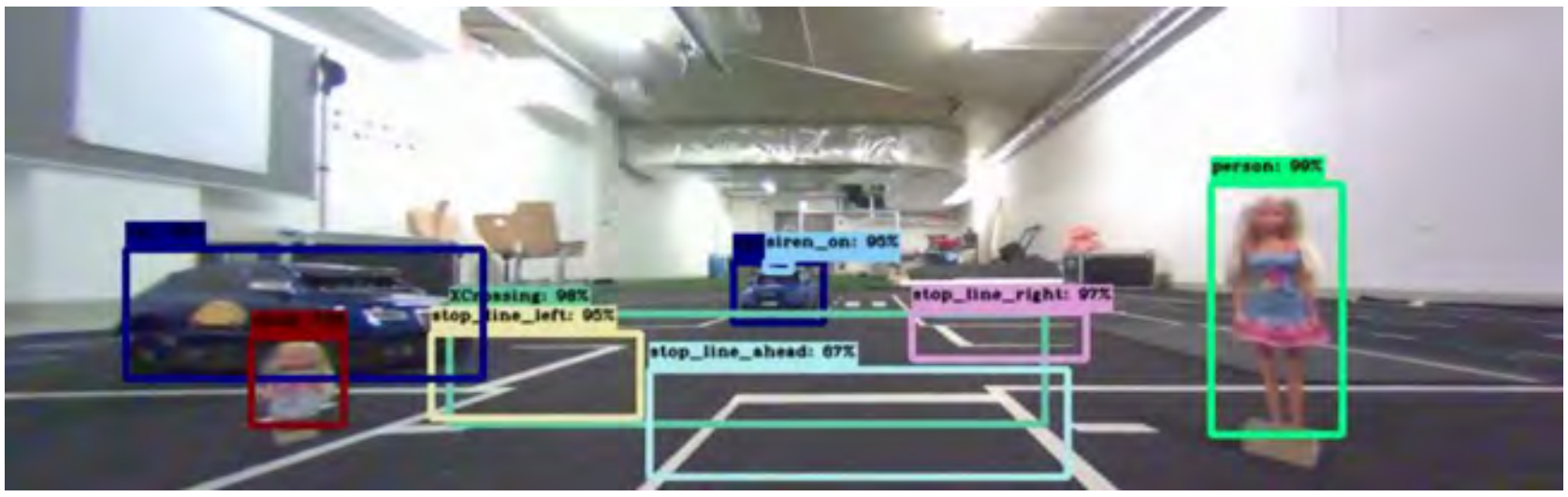

**Fig. 2.** Example detections with Faster R-CNN on Inception V2 [13]

# 4 Conclusions

In this paper we have shown that recent deep networks can run on the Audi Autonomous Driving Cup cars' GTX 1050Ti with reasonable frame and detection rates. This is remarkable given that 15 different classes of objects have been trained into the network.

A considerably higher accuracy can be achieved when reducing the number of classes to detect. Running a network with full crossing detection only achieves, for example, an accuracy of 0.987 despite the complex structure of X (see Figure 2) and T crossings. However, running multiple networks concurrently is not feasible on the single graphics card of the cars performance wise.

Currently, at higher speeds of the car, detection rates drop probably due to motion blur. This is not a general limitation of the deep learning networks, but indicates that more training images are required at higher speeds of the car.

# Design and Implementation of a Deep-Learning Course to Address AI Industry Needs

Bastian Beggel and Johannes Schmitt

Kaiserslautern University of Applied Sciences
bastian.beggel@hs-kl.de, johannes.schmitt@hs-kl.de

**Abstract.** Industry in Europe and worldwide currently has high interest in deep learning, but many medium-sized and smaller companies fear significant investments in specialized hardware and shortage of qualified personal. The goal of our work is to address these two issues. The University of Applied Sciences in Kaiserslautern is starting to offer a deep learning course in its new computer science master program. The course specifically addresses industry needs by focusing on the practical applications of deep learning in the fields of computer vision and natural language processing. We enable our students to perform deep learning projects by teaching hands-on technical expertise and programming tools that are required to do so using cloud systems. We believe that our work will help to enable local industry partners to develop deep learning supported artificial intelligence solutions in the near future.

**Keywords:** Deep Learning, Artificial Intelligence, Course, Tensorflow, Kerras

## 1 Introduction

Artificial intelligence (AI) has made an astonishing development in recent years and surpassed human intelligence at a diverse set of difficult tasks ([1], [2] and [3]). Deep learning is the key technology that facilitates this progress. Deep learning uses a cascade of multiple layers of nonlinear processing units for feature extraction and transformation to solve supervised and unsupervised learning problems.

To prepare students for upcoming AI challenge, universities and universities of applied sciences started to offer AI and deep learning focused courses. These courses often focus on the mathematical foundations and technical details. An indication of such a theory-oriented course is examination in form of a verbal or written exam instead of project work. Theory-oriented deep learning courses may constitute several drawbacks. For instance, graduates may actually miss necessary hands-on programming, data handling and model optimization experiences to successfully apply deep learning in real-world industry applications. Furthermore, programming-oriented but mathematics-averse students may not have the mathematical background and might shy away from these courses and, thus, do not acquire deep learning capabilities at all.

Thus, the focus of this work is the design and implementation of a deep learning course with following preconditions and targets:

- Focus on practical exercises using cloud systems,
- As few mathematics as possible,
- Project work as examination,
- Non prior knowledge of machine learning,
- Focus on supervised learning applications, in particular image classification and natural language processing,
- Inverted classroom.

# 2 Methods and Materials

## 2.1 Computing Resources

Deep learning requires specific computing resources namely graphics processing unit (GPU). GPUs operate at lower frequencies but have many times the number of cores compared with traditional central processing unit (CPU). Thus, GPUs can process far more data per second than CPUs at lower purchase and operation costs. The University of Applied Sciences in Kaiserslautern has invested in such a GPU equiped computing resource (Table 1 for details) to facilitate deep learning projects and courses.

**Table 1.** Details of the deep learning server at Kaiserslautern University of Applied Sciences (price point about 20k€ in 2018).

| Component | Description |
|---|---|
| CPU | 28-Cores - 2x Intel Xeon E5-2680v4 (2.40-3.30GHz, 14-Cores) |
| RAM | 256GB DDR4-RAM 2400MHz ECC Memory Reg. (16x32GB) |
| GPU | 4x NVIDIA GTX 1080 Ti (11GB RAM 3.584-Cores) |
| SSD | 2TB SSD SATA 6Gb/s |
| HDD | 40TB HDD (Raid 5) |
| OS | Linux Ubuntu 16.04 LTS |
| POWER | 4 * 1900W redundant power supply |

In addition to on-premise computing resources, cloud resources were also employed. The Google Cloud Platform supplied education grants for both the teaching stuff (2 times 100 US$) and the students (25 times 50 US$). The Google Cloud Platform offers ready-to-use deep learning virtual machines with configurable computing resources and computing libraries that can be made available within a few minutes at nearly arbitrary scale. Student grants of 50 US$ provide about 50 to 100 hours of suitable computing resources.

## 2.2 Libraries and Toolset

The field of deep learning is a very active but immature research field, that besides computing power heavily relies on capable toolsets and software libraries. So-called deep learning libraries (e.g. Tensorflow ([4]), Kerras ([5]), Pytorch ([6]) and Caffe ([7]) among others) are of special importance as they facilitate the programming, training and inference of deep neural networks using very high (and comfortable) abstraction layers. For the purpose of an introductory lecture Tensorflow and Kerras were selected to be used. Tensorflow is the standard library in research and industry for deep learning applications. Furthermore, Tensorflow forms the basis of almost any other deep learning as backend. Kerras provides a more compact programming interface (compared with Tensorflow) that allows even more rapid programming but uses Tensorflow for all low-level computations.

Further, Numpy, SciPy ([8]) and Pandas ([9]) are required for process programming data preparation.

Jupyter notebooks ([10]) are another essential programming tool. These notebooks provide access to a web-based interactive code execution engine with markup language support for documentation and plotting capabilities.

## 2.3 Literature and Lecture Materials

Deep learning has a profound mathematical background that comprises elements of the fields of linear algebra, probability theory, numerical optimization and machine learning. Ian Goodfellow et al. provides an comprehensive text book on these preliminaries tailored to deep learning ([11]) and is freely available as HTML version.

The Massachusetts Institute of Technology (MIT) provides excellent deep learning online courses. The introductory course [12] contains seven lectures of the most relevant topics of deep learning and includes slides and videos of very high quality. Three of these lectures (detailed in Table 2) were used to implement the inverted classroom concept.

**Table 2.** Contents of the Massachusetts Institute of Technology lectures 6S191_L1, 6S191_L2 and 6S191_L3. These lectures are used for the inverted classroom concept [12].

| Name | Content | Video Length |
|---|---|---|
| 6S191_L1 | The Perceptron<br>Forward propagation<br>Neural networks<br>Loss optimization<br>Backpropagation<br>Regularization | 45 minutes |
| 6S191_L2 | Sequence modells<br>Recurrent neural networks<br>LSTM networks | 36 minutes |
| 6S191_L3 | Tasks of computer vision<br>Convolutional neural network<br>CNN architecture<br>Beyond classification | 41 minutes |

To implement the practical parts of the course several programming exercises in the form of Jupyter notebooks were prepared [13]. Multiple public datasets were utilized including MNIST ([14]), CIFAR-10 ([15]) for convolutional neural networks (CNNs) and several freely available text books for recurrent neural networks (RNNs). These notebooks provide working examples of deep learning programming examples implementing the following procedure:

1. Data preparation with missing parts as programming exercises,
2. Presentation of a suboptimal neural network as the basis of programming exercises to analyze and optimize the network performance,
3. Presentation of a more suitable neural network as the basis of programming exercises to analyze the network performance.

## 2.4 Project work

Student projects for examination are based on reproducing, understanding and documenting results from existing and challenging deep learning application papers. A list of proposal papers for CNNs ([16],[17] ,[18],[19] ,[20]), RNNs ([21],[22],[23]) and Generative Adversarial Networks (GAN) ([24],[25]) were assembled.

# 3 Results

A deep learning course was designed to meet the targets and respect the preconditions as described in the introduction.

**Table 3.** Detailed outline of the full course. Each row describes a 180 minutes block divided into 45 minutes for theory upfront teaching and 135 minutes for hands-on practical programming exercises.

| Nr. | Theory | Exercise & Homework |
|---|---|---|
| 1 | Motivation for deep learning<br>Outline course<br>Present student projects | Python, numpy, pandas<br>Data preparation programming<br>Homework:<br>Finalize programming examples |
| 2 | Machine Learning Basics<br>Regression and classification<br>Test/train split<br>Regularization | Machine Learning<br>programming exercise<br>Homework:<br>Watch MIT lecture 6S191_L1 |
| 3 | Recap of MIT lecture 6S191_L1<br>Model Optimization for DL<br>Test/train split for DL | Dense networks<br>Tensorflow<br>Homework:<br>Watch MIT lecture 6S191_L2 |
| 4 | Recap of MIT lecture 6S191_L2<br>Introduction to Kerras | RNNs<br>Kerras<br>Homework:<br>Watch MIT lecture 6S191_L3 |
| 5 | Recap of MIT lecture 6S191_L3 | Programming exercises<br>CNNs<br>Kerras |
| 6 | Teaser of further DL topics<br>Project proposals | Project assignment |
| 7-12 | Status meetings<br>with project teams | |

In the first two lectures, the basic concepts of machine learning using the python machine learning technology stack are introduced and strengthened using multiple programming exercises. These two sessions are not specific to deep learning in particular and could also be placed at the beginning of a traditional machine learning course. Nevertheless, it is important to make the students familiar with the programming environment and to build a rudimentary understanding of the challenges of machine learning with respect to prediction, regularization and overtraining.

The next three lectures then use course materials (slides and videos) from the MIT introductory deep learning course (see Table 2 for details) to implement the inverted classroom concept. In these three sessions Neural Networks, RNNs and CNNs are introduced in this order, accompanied by extensive and suitable programming exercises. The MIT course material needs to be extended by theory chapters on model optimization techniques and procedures suitable for deep learning models. It is not expected that students fully understand all mathematical and technical details of the complex deep learning network architectures and inference procedures. The focus of the lectures needs to be on the programming exercises. Students utilize cloud systems to perform programming exercises. Cloud systems are required to enable more than a hand full of students to work in parallel on deep learning challenges.

With the sixth session the course enters its project phase. Students in groups of two to four are assigned to a deep learning project. The preparation of practically relevant and interesting student projects with a suitable level of complexity is a major challenge that was addressed. Project progress is monitored and project support is provided with bi-weekly status meetings.

## 4 Conclusions

A concept for a deep learning course was presented that addresses current industry needs and is suitable for a computer science master program of a university of applied sciences.

The authors believe that this work will help to enable local industry partners to develop deep learning supported AI solutions in the near future. Finally, the work presented here might motivate and help other education facilities to offer similar programs to support European AI initiatives at a wider scale. Feel free to contact the authors for friendly exchange of experience, advice or course materials.

# Reinforcement Learning - Agents Learn to Interact in Unknown Environments

Matthias Haselmaier[1], Alexander Schwarz[2], and Tim Hallyburton

[1] Kaiserslautern University of Applied Sciences
matthias.haselmaier@hs-kl.de
[2] Kaiserslautern University of Applied Sciences
alexander.schwarz@hs-kl.de

**Abstract.** Reinforcement Learning describes a machine learning paradigm which is applied to find an optimal sequence of actions to achieve a given goal. An agent receives an reward for performed actions and has to find a policy which maximizes the expected value of the sum of cumulative rewards. In this paper linear function approximation and SARSA Learning are explained and implemented. They are applied to the computer game Breakout. Experiments with different lengths of training episodes and different hyperparemeters were performed and the results presented. Because of the selected features very good results were achieved after only a short training periode.

**Keywords:** Machine Learning, Reinforcement Learning, SARSA Learning, Function Approximation, Temporal-Difference Learning

## 1 Introduction

Reinforcement Learning is a paradigm of machine learning. It is based on the idea to reward good actions and punish bad actions and tries to train an agent to perform sequential actions to achieve a given goal. One of its defining characteristics is the interaction between the agent and his surrounding environment. It is shown in figure **??**.

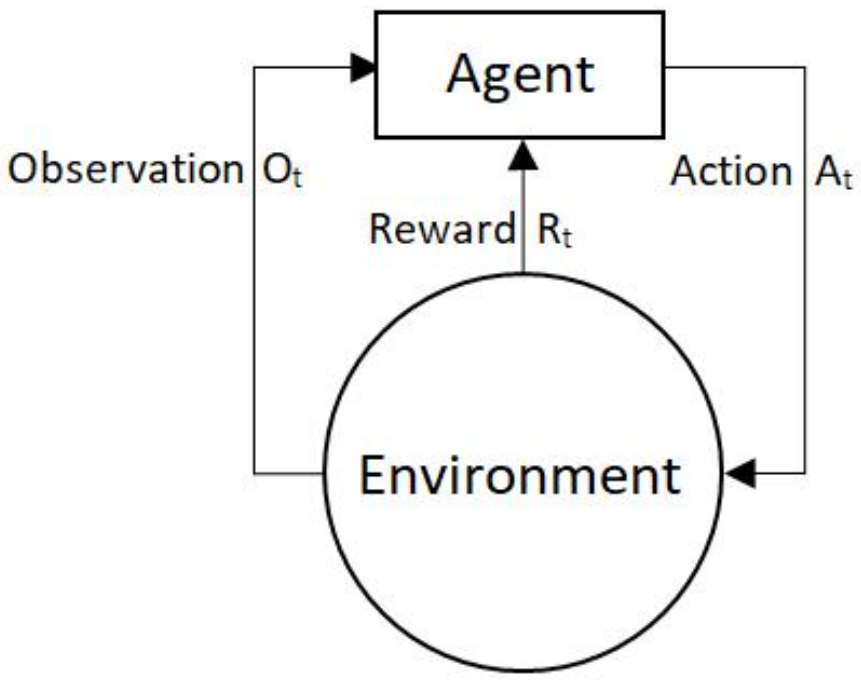

**Fig. 1.** The agent in his environment

With every action $A_t$ the agent takes he is given a scalar *reward* $R_t$. He must use this *reward* to improve the *policy* he is using to choose his actions. Compared to other machine learning paradigms time is another defining charatistics of Reinforcement Learning, as the *reward* the agent receives for a specific action may possibly be received many timesteps after the agent performed the action. Furthermore the information or observation $O_t$ the agent receives about its surroudings, which he has to make his decisions on, is not i.i.d. as it is influenced by his previous actions. If e.g. a small robot with a camera explores the right side of a room it may see drasticly different things as if he was to explore the left side of the room. To achieve his goal the agent has to find the *policy* that maximizes the expected value of the cumulative sum of *rewards*.

What happens in the environment surrounding the agent is based on the current state $S_t$. This state can be formulated as a function of the historie $H_t$, where $H_t$ is the sequence of observations, rewards and actions until timestep $t$.

$$H_t = O_1,\ R_1,\ A_1,\ \ldots,\ A_{t-1},\ O_t,\ R_t \tag{1}$$

$$S_t = f(H_t) \tag{2}$$

This state $S_t$ can be seperatet into the environment state and the agent state. The environment state may contain information that is unimportent or not accessible to the agent. The agent state includes everthing he needs to chose his next action. This also is the information that is used by the Reinforcement Learning algorithm. If the state $S_t$ has the Markov property only the most recent state may be saved as there is no information lost.

Reinforcement Learning was proven to be successfull in a variaty of different problems, e.g. [?], [?], [?]. In this paper we apply Reinforcement Learning to the computer game Breakout and present a small software framework which can be used to implement a Reinforcement Learning algorithm. We will present the problem space in more detail first and explain the theoretical background of Reinforcement Learning. Afterwards we present our implementation and finally discuss the result of the applied Reinforcement Learning algorithm.

## 2 Breakout Application

The game Breakout was chosen as learning task. Breakout is a relatively simple game which was first published by Atari in 1976 as an arcade game. Basically Breakout is a grid of blocks at the top of the window, which have to be destroyed by a ball. The player tries to hit the ball with a paddle, which is located at the lower border of the window. The paddle can be moved left or right. The goal is to destroy all the blocks without losing your lives.

For this application a clone was implemented in Java. The grid consists of 90 blocks, which are arranged in 6 rows and 15 columns. If the ball hits the left, right or upper window border or one of the blocks, the exit angle corresponds to the entry angle. The player has the possibility to control the ball minimally. This happens when the paddle is in motion at the time of the collision with the ball. In this case the angle of the ball is deflected by 5 degrees in the respective direction. If the ball hits the resting paddle, the bounce behaves as with the walls and the blocks. At the beginning of each game, the player starts with 5 lives. If the ball is not returned upwards and hits the lower border of the window, the player loses one life. Each time a new life is started, both the ball and the paddle are placed in the middle of the window. At the beginning the ball always moves straight along the Y-axis to the bottom and in a random direction on the X axis. For each block hit by the ball, the player receives 5 points. Once all the blocks have been destroyed, there is a bonus of 100 points and the game is reset like a life loss. At any time the player has the 3 possible actions stop, move left or move right.

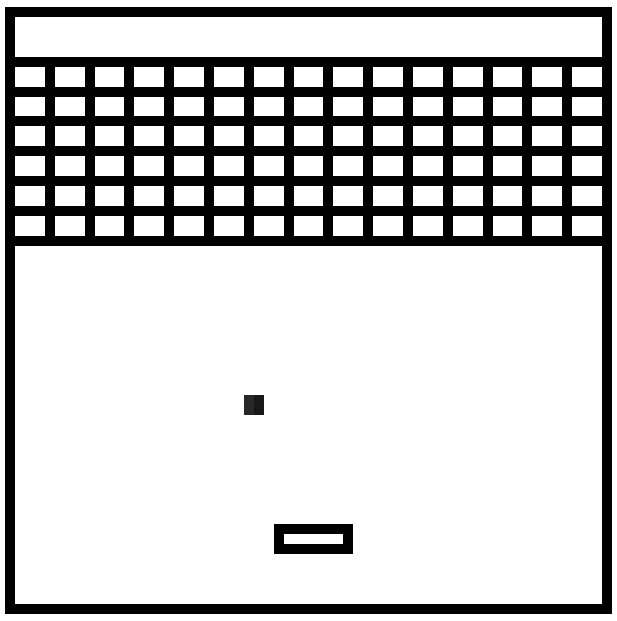

**Fig. 2.** Breakout-Application

## 3 Used Method and Algorithm

Since the state space of Breakout is too large for a tabular solution, the method of Value Function Approximation was used together with the SARSA Learning Algorithm. The SARSA algorithm belongs to the Temporal Difference Learning algorithms and uses the transitions from state-action pair to state-action. It is described by naming the tuple $(S_t, A_t, R_{t+1}, S_{t+1}, A_{t+1})$. $S$ is the current state, $A$ the currently executed action, $R$ the reward for this action, $S_{t+1}$ the subsequent state and $A_{t+1}$ the action in the subsequent state. The update rule used is the following:

$$Q(S_t, A_t) \leftarrow Q(S_t, A_t) + \alpha\big[R_{t+1} + \gamma Q(S_{t+1}, A_{t+1}) - Q(S_t, A_t)\big] \tag{3}$$

The updates are performed after each transition from a non-terminal state. If $S_{t+1}$ is a terminal state, $Q(S_{t+1}, A_{t+1})$ is set to zero.

**Algorithm 1** SARSA-Algorithm

Initiazlize $Q(s,\ a), \forall s \in S,\ a \in A(s)$, arbitrarily, and $Q(\text{terminal-state},\ .) = 0$
Repeat (for each episode):
Initialize $S$
Choose $A$ from $S$ using the policy derived from $Q(e.g.,\ \epsilon\text{-greedy})$
Repeat (for each step of episode):
Take action $A$, observe $R,\ S'$
Choose $A'$ from $S'$ using the policy derived from $Q(e.g.,\ \epsilon\text{-greedy})$
$Q(S,\ A) \leftarrow Q(S,\ A) + a[R + \gamma Q(S',\ A') - Q(S,\ A)]$
$S \leftarrow S';\ A \leftarrow A';$
until $S$ is terminal

The Value Function Approximation approximates a state-action function $q_\pi$ as a parameterizable function $\hat{q}$ with a feature vector $w$. The action-value function $\hat{q}$ is a linear combination of features:

$$\hat{q}(S, A, w) = x(S, A)^T w = \sum_{j=1}^{n} x_j(S, A) w_j \tag{4}$$

A feature is the description of a state action pair of the state space, e.g. how the distance from the paddle to the ball in the breakout example in state $S$ changes due to action $A$. A state action pair is described by several features which are collected in a feature vector.

$$x(S, A) = \begin{pmatrix} x_1(S, A) \\ \vdots \\ x_n(S, A) \end{pmatrix} \tag{5}$$

The update function for the feature vector $w$, using the SARSA learning algorithm looks as follows.

$$\Delta w = \alpha(R_{t+1} + \gamma \hat{q}(S_{t+1}, A_{t+1}, w) - \hat{q}(S_t, A_t, w))\nabla_w \hat{q}(S_t, A_t, w) \tag{6}$$

The weight vector is now updated in each step.

## 4 Used Features

As a rather simple approach, the goal was to hit the ball with the paddle. The control over the direction of the ball is quite small, but if the ball is kept in play as long as possible, it is only a matter of time until all blocks are cleared. To achieve this goal, a feature has been implemented that maps the distance between the center of the ball and the paddle in the interval from 1 to 0. The closer the ball is to the paddle, the lower is the value of the feature. This caused the angle of the ball to become flatter over time, and since the ball is slightly faster than the paddle, the agent played himself in situations where he could no longer reach the ball. For this reason, an attempt was made to position the paddle in the middle of the playing field as soon as the ball was up to the blocks. For this a feature was implemented, which indicates the distance to the center of the playing field. The combination of these two features resulted in the agent being torn between the decision to minimize the distance to the ball or to the center, and usually just standing still. Several different combinations with other features were tested. The combination that performed best was formed from the first feature, the distance between the ball and the paddle, and a feature that maps the angle of the ball to an interval from 0 to 1, where 1 is a very steep and therefore preferred angle. The steep angle makes it easier for the agent to hit the ball again, as the paddle only needs to move minimally. Also, the distance feature has been adjusted to take a value of 0 if the ball will hit the paddle in its current trajectory if it doesn't move. Otherwise, the agent would try to further reduce the distance, making the angle of the ball flatter in many cases. The results from the next sections were obtained with this feature combination.

## 5 Results

The implemented framework served as a basis for several test series, which are now examined in more detail. In the first experiment the number of learning episodes was varied. With the resulting weights $w_1, w_2$ of the features $f_1, f_2$, 100 episodes were played. The average reward achieved during these episodes serves as a quality measure for the combination of $w_1$ and $w_2$. The remaining parameters were freely chosen and remained constant during the test series. They were $\alpha = 0,2$, $\gamma = 0,3$ and $\epsilon = 0,05$.

After just a few episodes, less than 10, good results were already achieved. This short training period was achieved because the chosen features were created with knowledge about the problem. Especially the distance from the ball to the paddle effected the fast learning process which resulted in the ball being hit almost every time. On average a reward of 1856 was achieved.

In further experiments, the parameters $\alpha, \gamma$ and $\epsilon$ were varied and the number of training episodes remained constant. Particularly interesting is the variation of $\epsilon$ in the range $\epsilon \in [0.02, 0.05]$. A higher $\epsilon$-value can achieve better results if only a limited number of training episodes (here: 1000 episodes) are run through.

## 6 Conclusions

In summary, it can be said that the selected features can achieve very good results even after short training phases. However, the way these results are achieved can still be improved. Currently the agent avoids losing a ball and thus a "life" - but the remaining blocks are not considered when selecting the next action. The breaking of the blocks, and the associated reward, is therefore only a by-product of striving not to lose the ball. This causes the $\frac{Reward}{Time}$-ratio to deteriorate as the number of remaining blocks decreases. It would be more effective to destroy the blocks by targeting them. To achieve this a feature could be used that indicates whether the ball is currently under a block. The goal of such a feature would be to show the agent where a flat or steep angle in the field promises a better success.

# Two Examples of Using Machine Learning for Processing Sensor Data of Capacitive Proximity Sensors

Björn Hein[1]* and Hosam Alagi[2]

[1] Karlsruhe University of Applied Science
bjoern.hein@hs-karlsruhe.de
[2] Karlsruhe Institute of Technology
hosam.alagi@kit.edu

**Abstract.** Recently we have shown developments on capacitive tactile proximity sensors (CTPS) in combination with machine learning techniques to extract further information out of the sensor signals. In this work we summarize two examples of the applications we have presented. In the first approach we have investigated distance classification based on the proximity information of the sensors. In the second approach the possibility of material recognition was investigated. The latter is done by variating the spatial resolution and the exciter frequency of our sensors. For both approaches, distance classification and material recognition, an artificial neural network was set up and fed with various data sets of different electrode combination. The influence of the electrode combinations and shapes on the recognition accuracy was investigated and some promising results could be achieved.

## 1 Introduction

An established sensor technology, at least in research, are sensor skins for robots. These can detect touch and provide information about location and force of this touch. These skins are used for a variety of applications ranging from human-robot-interaction (HRI) and grasping concepts. More recently the ability to detect objects/events in the near proximity of the robot are under investigation as they provide the interesting possibility to get information right before the touch event even happens. This extends even further the possibilities in HRI, grasping and could significantly contribute to safety concepts. Proximity sensing closes in a unique way the perception gap between classical vision based sensing and touch sensing.

For the implementation of this proximity sensor modality different measurement principles have been proposed based on different physical effects, mainly acoustic, optical and capacitive; examples are [1–3].

Our line of work has been dedicated to designing hardware and applications for capacitive tactile and proximity sensing. Capacitive sensing is affected by the electrical properties of the object (conductivity, permittivity when non-conducting), its size and shape and of course its distance to the sensor. Unfortunately this implies complicated dependencies in the sensor signals, which are not easy modeled.

The research presented here is a continuation of our previous work [4], where we introduced a flexible, easy to integrate capacitive tactile proximity sensor for applications in robotics (s. Fig. 1).

* This research was funded by the German Research Foundation (DFG) under the grants HE 7000/1-1 and WO 720/43-1.

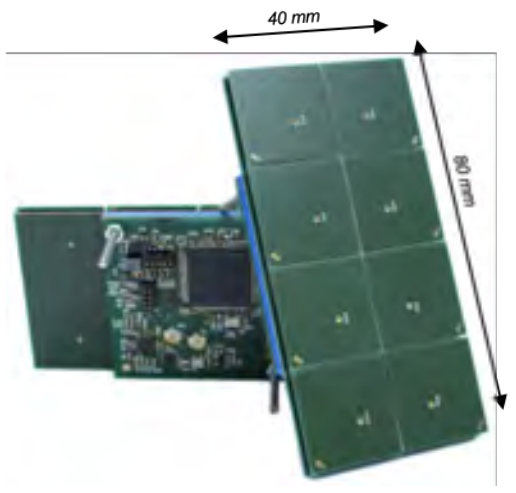

**Fig. 1.** Sensor module and electrodes of the sensor presented in [4]. Up-to eight electrodes can be combined individually.

In this work we especially summarize two aspects of our previous research, which did focus on using machine learning techniques for extending distance measurement [5] and material recognition [6].

The rest of the paper is structured as follows: After this introduction the state of the art is presented in Section 2. In Section 3 we present our sensing system used to collect the data about the different objects. In Section 4.1 we discuss the experiments in material recognition and their results. Finally, in Section 5, we provide a summary and give conclusions for this paper.

## 2 Related Work

A very interesting application scenario for proximity sensing is preshaping of a robot end-effector, i.e. aligning the gripper of a robot to an object prior to finally grasping it. Preshaping applications for robotic gripper have been shown based on proximity sensing with capacitive sensors in [7–9]. Another important application for proximity sensing is collision avoidance. Early works are [10] and the milestone work by Lumelsky and Cheun [11]. As mentioned above, capacitive sensing is susceptible to shape, material and size of the objects it perceives and suffers from strong non-linearity. Therefore machine learning seems to be a valid way to address these issues.

Like it was stated in the introduction, the second aspect to be presented in this work is material classification. Capacitive proximity measurements have the potential to be applicable also for this kind of scenario. Materials can basically categorized and distinguished by their relative permittivity $\varepsilon_r$. In general this can be done by gathering sample data and then applying classical classification methods. Kirchner et al. proposed data generated by using three discrete frequencies [12] to classify four classes of objects (concrete, human, metal and wood). The use of different frequencies is decisive for their results. By driving the sensor towards the object, a distance dependent curve was recorded which was then used for the material ranging. In our work we also use different frequencies, but we propose an approach which only depends on the sensor values. In other words, a single frame of sensor measurements is used to predict the material.

Another interesting approach for an impedance analyzer using magnetic and electric component is shown by Yunus et al. and Wang et al. [13] and [14] respectively. Both works target the same application domain regarding water pollution detection. Wang et al. specifically investigated different electrode designs in [14]. Due to the fact that significant changes in the signal have been observed in permittivity at frequencies below 500 kHz and that the shape and the order of the electrodes are major for the measurement, we see the analogy to our sensor in the multi-frequency measurement and the dynamic electrode configuration. However, our approach is less complex using two discrete frequencies and the electrode reshaping was realized by the flexible spatial resolution of the sensor.

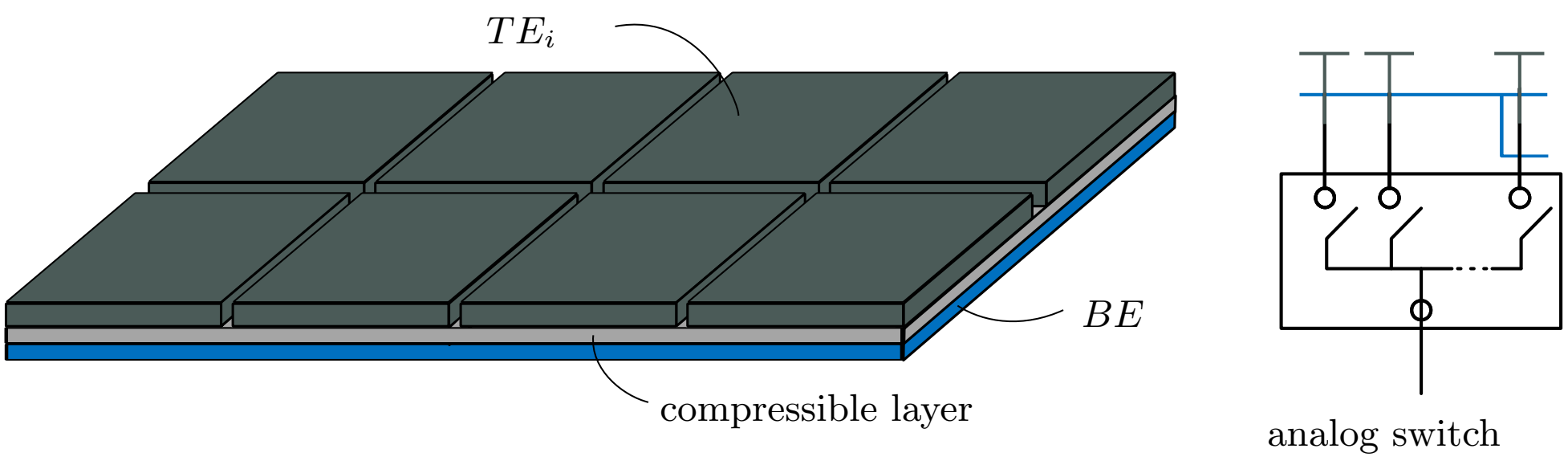

**Fig. 2.** The assembly of a sensing element ($TE_i$ - Top Electrode $i$, $BE$ bottom electrode. An analog switch allows selection and combination of electrodes [4].

A simple capacitive proximity sensor with one exciter frequency and fixed electrodes would not provide enough information about detected objects. Therefore, A.Kimoto in [15] combined optical sensor with capacitive one. He was able to distinguish between Acrylic, PTFE, Class and Aluminum in different surface properties. In the context of robotics and especially in grasping applications collecting information about the object is essential. In such scenarios a gripper or manipulator equipped with suitable capacitive sensors can provide internal properties of the object's material. Respectively, the control system can then decide which object should be grasped [16]. In this case the capacitive sensor provides information which is complementary to those of vision or haptics.

## 3 System Description

The sensor design, which was presented in detail in [4] (s. Fig. 1), can be used in a very flexible manner, in this section we therefore start with presenting the general concept of our sensor design and then explain the hardware implementation needed for the two approaches: distances measuring and material classification.

### 3.1 Sensor Design

One specialty of our sensor is its ability to reconfigure its spatial resolution on-line by merging the electrodes using an analog switch (s. Fig. 2). This increases the measurement area of the resulting electrode and thus the sensitivity of the measurement. Additionally the sensor is able to use different frequencies for the measurement and multiple sensors can by used in a synchronized fashion. Fig. 5 shows in the material classification task three of our sensor modules as part of an end-effector that is approaching a wooden ball.

The sensor can be used in *self-capacitive* (send) and *mutual-capacitive* (receive) mode. Both modes are alternatively called *single-ended* and *double-ended* in literature. In the self-capacitive mode the top electrodes ($TE_i$) are driven with an AC signal passing through the analog switch, which can select or join electrodes arbitrarily. The signal causes the periodic charging and discharging of the electrodes. The capacitive coupling of the configuration with the environment can be measured by finding the amplitude of the current flowing through the circuit. To guide the coupling towards the "outside" side of the sensor, the bottom electrode $BE$ is used to actively shield the $TE_i$ from any components below. An object will affect the coupling according to its material properties and its proximity to the sensor. In the mutual-capacitive receive mode the $TE_i$ are connected to ground through a measurement circuit. In the presence of a sender there will be an electric field lines showing from the sender towards the receiver, inducing a current at the receiver side. An object near the electrodes will affect the field. This effect is reflected in the current being measured and again depends on the object properties. This double-ended measurement is suited for detecting insulated objects with a high enough relative permittivity. With our sensor a tactile measurement is also possible due to the compressible, insulating layer between $TE_i$ and $BE$, but it is not used in this work. Finally, the frequency of the exciting signal can be adjusted, which is also an important aspect, i.e. for material recognition, since the permittivity can be a frequency dependent value.

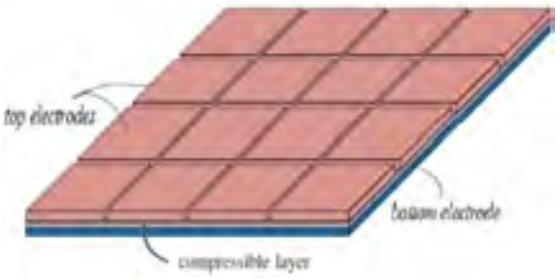

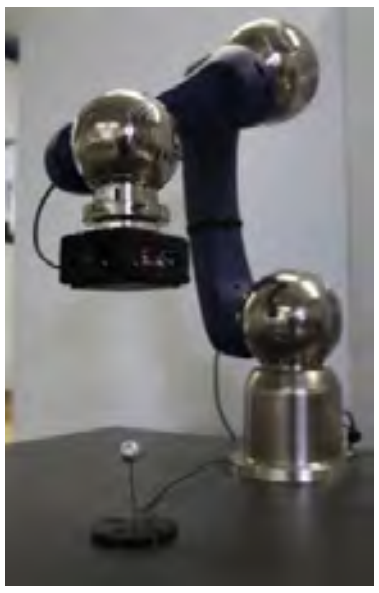

**Fig. 3.** First sensor setup for investigating distance measurement based on machine learning. Left: $4x4$-sensor configuration using 2 sensor modules controlling $2x4$ electrodes at a time; Right: Schunk-LWA with $4x4$-sensor configuration attached as end-effector.

For both setups the end-effectors with the sensors were mounted on small 6-Axes-Robots. Distance measurement tests were done with a SCHUNK-LWA (s. Fig 3, right), material classification tests where done using an UR5 (s. Fig. 5, top right) and the Robot Operating System *(ROS)* was used as a middleware. For generating and executing the trajectories *MoveIt* and *ROS Control* were used.

### 3.2 Sensor configuration

**Sensor setup for evaluating distance measurements via machine learning**
For the machine learning approach measuring distances 2 sensor modules with 2x8 electrodes were used as sensing elements (s. Fig. 3). In this setup the end-effector consists of 16 electrodes and the spatial resolution is used to take measurements with different electrode and sizes, i.e. signals from one single electrode or signals from 4 combined electrodes were recorded. So, a total of 22 different signals were collected. The basic idea behind this approach is that with a higher resolution (single electrodes) we get shorter measurement range and a higher uncertainty about the capacitances and therefore distances, but the object's size and localization is clearer. Conversely, with lower resolution (combined electrodes) we get a higher measurement range due to a better representation of the capacitance and therefore distances at the cost of loosing accuracy in the localization of the object. A frame combining both potentially unifies the advantages of both worlds.

The used end-effector for material detection is basically quite similar to the one for distance measurements and can be seen in Fig. 5. Two $4 \times 2$ modules combined make up the receiver array of the end-effector as already shown in the distance measurement set-up (s. Fig. 3). In addition a third module was added and acts this time as dedicated sender. Six of its electrodes are installed at both sides of the receiver array, but only the ones closest to the receivers have been activated (represented by blue stripes). The flexible spatial resolution is used to take measurements with different electrode sizes and shapes, as seen Fig. 5($C1_n$, $C2_m$, $C3_k$). At each resolution the set of measurements can be thought of as a capacitive image. The collection of all measurements then is a multi-resolution capacitive image, which we call a *frame*. The hardware sequentially generates a frame within some fraction of a second by time-multiplexing through each configuration of electrodes.

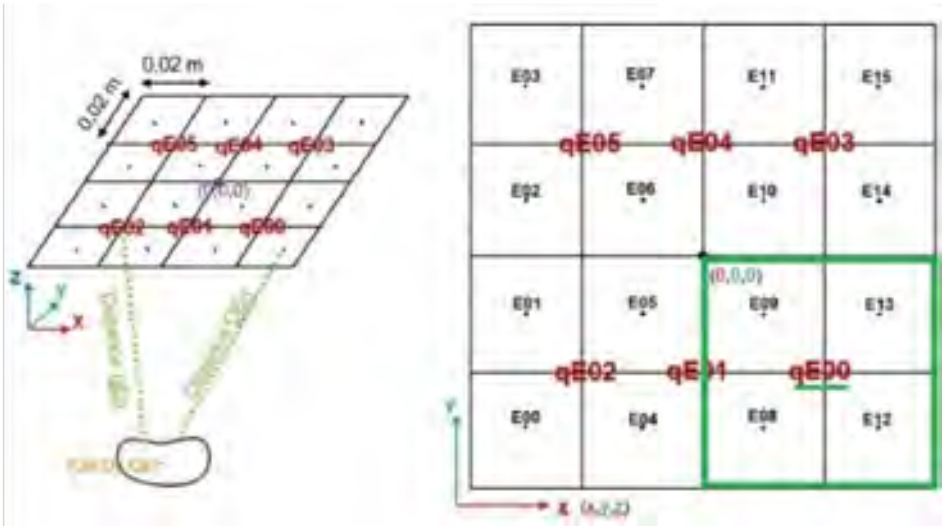

**Fig. 4.** Sensor configuration used for testing distance measuring [5]. Due to the flexibility in the combination of electrodes (s. Fig. 2) the signal of single electrodes ($E00 - E16$) and combined electrodes ($qE00 - qE05$, 4 combined electrodes) are fed to the neuronal network.

**Sensor setup for evaluating material detection via machine learning**
By reshaping the electrodes we assume to get different configuration of the potential field and thus the penetration or polarization of the dielectric objects. in addition two exciter frequencies were used to obtain information about the object material due to the fact that the relative permittivity shows a frequency dependency. Figure 5 shows the configurations used for the measurements, where three basic electrode combinations are highlighted.

# 4 Evaluation and Results

## 4.1 Experimental Setup: Distance measurements

Basic idea of this experiment was to classify distances according to 1cm steps. To generate the training data the end-effector (s. Fig. 3) was moved over the object in the pattern shown in Fig. 6.

In this section we discuss the experimental setup, the framework and setup for learning with artificial neural networks (ANNs) and the results obtained using different combinations of electrode configurations and signal frequencies. The latter - changing frequencies - was only used for the material classification test.

**Artificial Neural Network**
The distance classification neural network has been implemented using the *Tensorflow* framework. The recorded data was trained using a feed-forward network with 3 hidden layers. Each hidden layer consists of a set of fully connected neurons. All hidden neurons used the rectified linear unit (ReLu) activation function [17], the network was trained using backpropagation [18] method and the Adam optimizer [19].

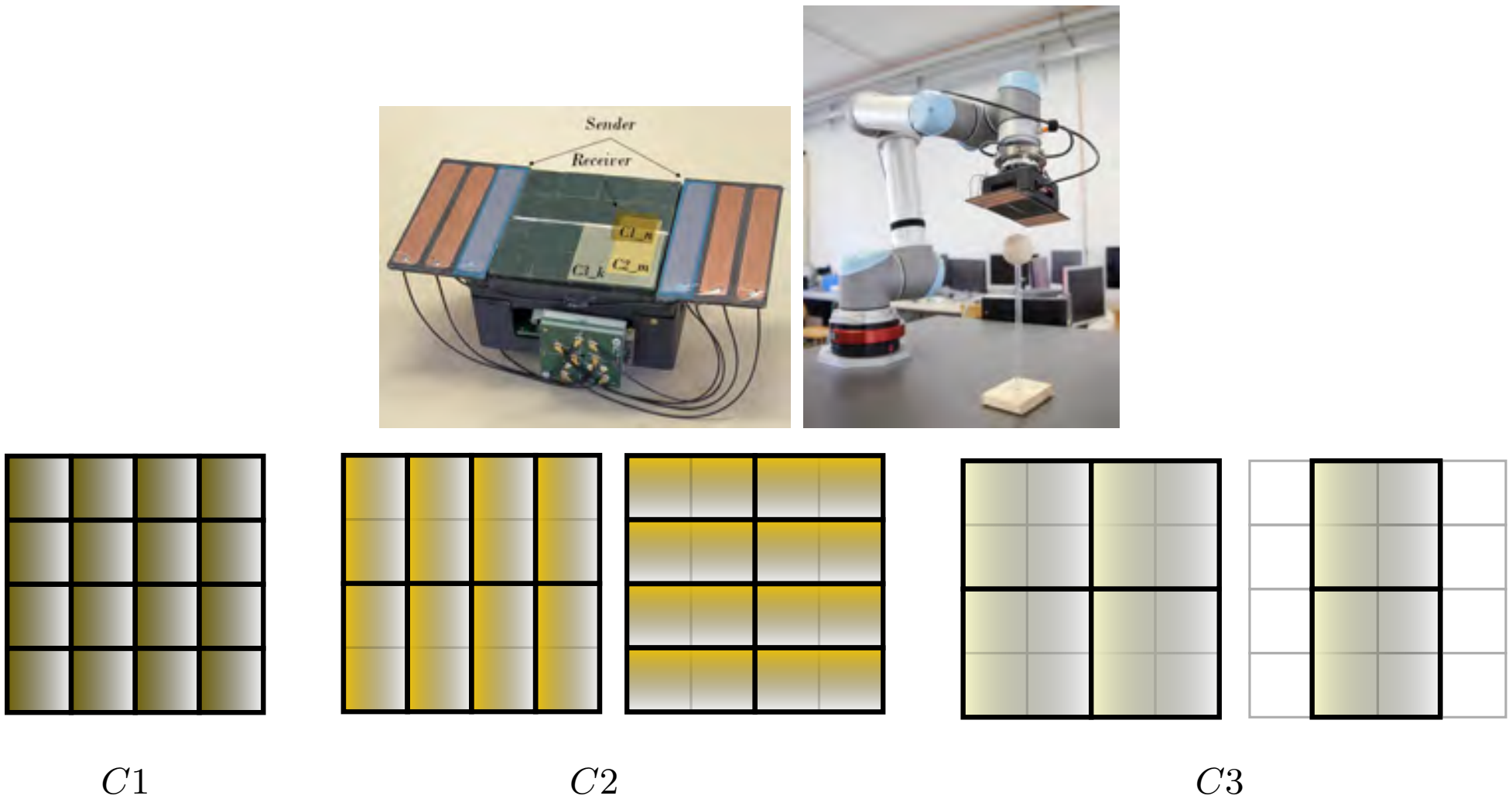

**Fig. 5.** Second sensor setup for classification of materials using machine learning. Top Left: The end-effector featuring three sensor modules: two $4 \times 2$ modules in the middle configured in receive mode and one module driving up to six electrodes in send mode. The PCB with the analog and digital electronics for the third module is visible in the front. The three modules are connected to the same I2C bus; Top Right: Setup with an Universal Robot with the sensors mounted on an end-effector. Similar to set-up shown in Fig. 3, the robot is used to perceive objects in its workspace. In this set-up it is used to classify them according to their permittivity or conductivity; Bottom: Illustration how the flexible spatial resolution is used to select single electrodes $C1_n$ with $n \in \{1, \ldots, 16\}$, configurations of two electrodes combined $C2_m$ with $m \in \{1, \ldots, 16\}$ and configurations of four electrodes combined $C3_k$ with $k \in \{1, \ldots, 6\}$.

The data, that was collected during the movement (s. Fig. 6) was shuffled and split into training and testing sets, the cross validation technique [20, 21] was performed with 5 folds (k=5), meaning one of the k subsets is used as the test set and the other k-1 subsets were combined to form a training set. The training algorithm had to run from scratch k different times. In the end, the average root mean squared errors was computed across all k folds.

**Results**

Multiple configuration were tested and per configuration 18246 data samples were generated. In [5] this is discussed more in detail. To give an idea two configuration are exemplary discussed and presented here. In a first notion the sensor signal of 3 different electrodes were used as feature input. This is based on the idea that triangulation requires at least 3 measurements to identify a location. The score that could be reached by this idea was 51.49%. Figure 7 on the left clearly shows, that distances close to the sensor can be distinguished quite good, but far distances led to misclassifications.

In summary we ended up using all available sensor measurements, i.e. all single sensor electrodes and all the combined ones ($E00 - E15$ and $qE00 - qE05$, s. Fig. 4) as input features. We got therefore a feed-forward backpropagation network with 22 inputs, 14 outputs for corresponding distances and 3 hidden layers with 80-40-20 neurons respectively. It yields after using cross-validation and tuning manually the hyper-parameters the highest classification score of all configurations with a value of 94.87%.

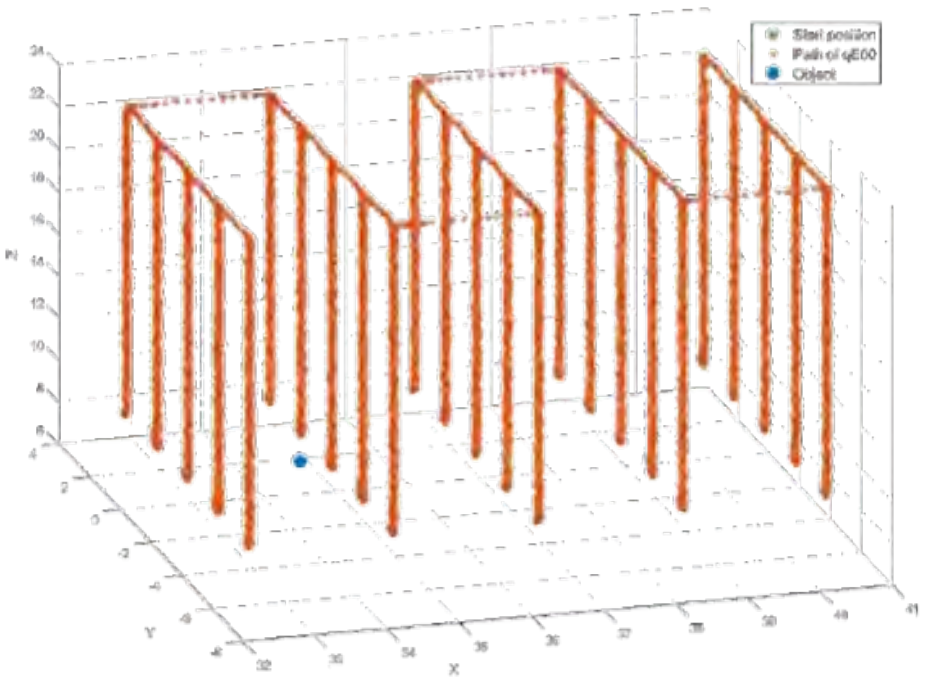

**Fig. 6.** Recording of the training data: The end-effector is moved over the object. At equidistant positions the end-effector moves downwards in direction of the object (s. Fig. 3). Based on the ability of the sensor to change the spatial resolution by combining electrodes, the sensor-data of single electrodes and of combined electrodes (s. Fig. 4) are recorded while moving.

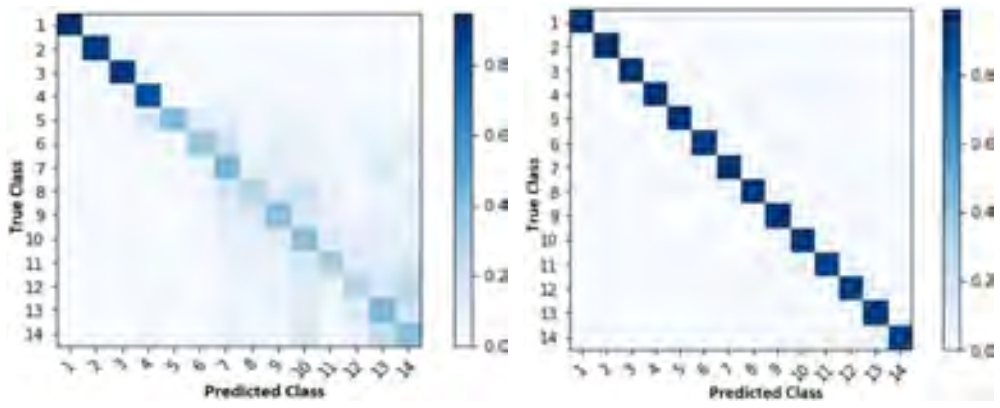

**Fig. 7.** Left: Using sensor information of three electrodes / electrode combinations as input allows a good classification for small distances but fails for far distances; Right: using all electrodes and all combinations (total of 22 sensor streams) lead to good classification results even for far distances.

Being this the first time for us to apply an ANN to our sensor data, the results were quite promising. Especially as the range for distance measurement was nearly doubled. Overall - for the given setup - the distances could be much more precisely measured and identified using ANNs than by using the original model-based approach.

### 4.2 Experimental Setup: Material classification

Encouraged by the previous results using ANNs for distance classification, a second way of exploiting ANNs and their classification abilities was investigated. Basic idea of this experiment was to classify materials based on their relative permittivity. The relative permittivity $\varepsilon_r(\omega)$ is depending on the frequency of an exciting electrical field they are in. As our sensors create such an electric field and are capable to use different exciter frequencies it was obvious step to try to exploit this capability.

Objects consisting of different materials (s. Table 1) were used to test the concept.[3]. All experiments were done in the laboratory by constant ambient temperature and humidity.

The distance between the electrode array and the top of the objects was kept to 2 mm during the measurements. Similar to the distance calculation in the previous Section the end-effector was moved in a meander path over the objects enclosing the borders of the receiver array (s. Fig. 9). This path was iterated ten times while the data was collected. In the training phase 74033 samples of all material were split in proportion 8/2 for training/validation. Additional 18512 samples were recorded for test. Every sample is a frame that contains the sensor values for the various electrode combinations and frequencies. Its size could variate depending on the selected combinations.

[3] The table is an excerpt of the values found in http://www.kayelaby.npl.co.uk/general_physics/2_6/2_6_5.html

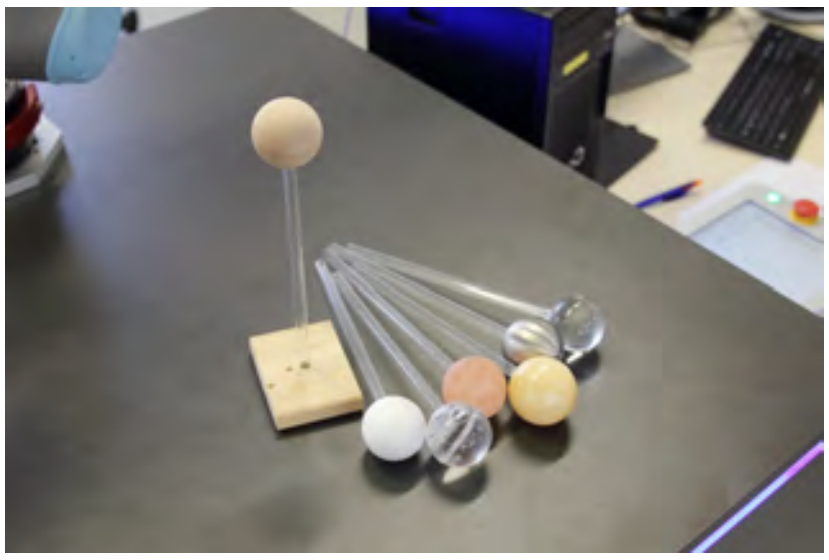

**Fig. 8.** Balls with diameter of $5cm$; Test objects of different Materials, one ball each

| Material | Relative Permittivity $\varepsilon_r$ |
|---|---|
| Styrofoam[22] | 1.03 |
| Glass (quartz) | 3.8 |
| Salt (NaCl) | 6.1/5.9 |
| Marble | 8.0 |
| Wood (Beech 16% water) | 9.4/8.5 |
| Tap water | 80.1 |

**Table 1.** relative permittivity of the used test materials

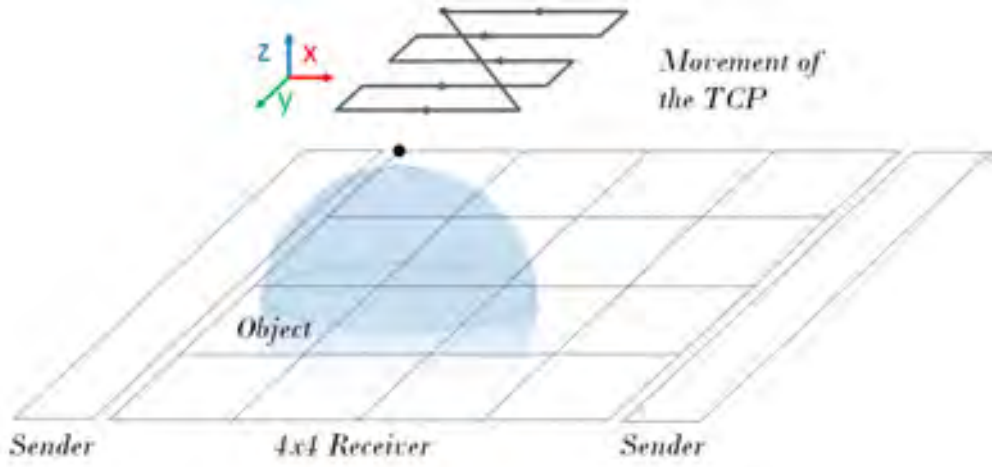

**Fig. 9.** Initial position of the end-effector and its path

## Artificial Neural Network

The material recognition neural network was also implemented in *Tensorflow*. The recorded data was trained using a feed-forward neural network (ANN). For the ANN, the number of the input neurons was defined as the size of the data frames, hence by the selected electrode combinations and frequencies for each data set. The input was then forwarded through a series of *ReLU-layers*, where a dropout rate of 10% had been set. Lastly, the data was fed to a *softmax* output layer, whose size equaled the amount of materials trained for. In addition to introducing dropout to all hidden layers, the weights were also modified through L2 regularization to further decrease overfitting. Updates to weights were done in mini-batches of 300 and optimized by the Adam optimizer [19]. To further prevent overfitting, every five epochs, the validation set's accuracy was calculated and training was prematurely ended when this accuracy exceeded a threshold of 0.98. Furthermore, a patience period of ten epochs to stop training was set, which counted up if the validation loss did not decrease compared to the last best loss. The best results were achieved by setting the hidden layer size to two layers with 400 units each. The initial learning rate was set to 0.001. The Hyperparameters are listed in Table 2. Those were determined in a previous experimental run and showed best results for all data sets. The hyper-parameter do not variate between the data sets.

| | |
|---|---|
| Hidden layers | 400, 400 |
| Batch size | 300 |
| Learning rate | 0,001 |
| L2 beta | 0,1 |
| Dropout | 0,1 |

**Table 2.** Hyperparameters of the ANN

**Results**

The measurement was performed for all seven objects[4]. For each object about 13100 data frames were collected. Due to the air-like relative permittivity of "styrofoam" this object acted as air.

| | |
|---|---|
| Set | 3 |
| Variation | $C2_m$ |
| | 89 kHz,178 kHz |
| Input | 32 |
| Epochs trained | 25 |
| Test accuracy | 94.2% |

$$\begin{bmatrix} me & sa & ma & wo & gl & wa & sf \\ 2665 & 20 & 0 & 0 & 8 & 2 & 0 \\ 2 & 1835 & 822 & 0 & 1 & 0 & 0 \\ 0 & 223 & 2436 & 0 & 0 & 0 & 0 \\ 0 & 0 & 0 & 2632 & 0 & 0 & 0 \\ 0 & 0 & 0 & 0 & 2640 & 0 & 0 \\ 0 & 0 & 0 & 0 & 0 & 2607 & 0 \\ 0 & 0 & 0 & 0 & 3 & 0 & 2627 \end{bmatrix}$$

| | |
|---|---|
| Accuracy | 94.2% |
| Precision | [99.3 69.0 91.6 100 100 100 100] |
| Recall | [99.9 88.3 74.8 100 100 100 100] |

The network was fed with different combinations of the measurement regarding the electrode combinations and the exciter frequencies. In [6] the different combinations are discussed in detail. It turned out that the $C2_k$ combination showed the best results. We believe that the $C1_n$ and $C3_k$ do not provide enough information. While $C1_n$ has the highest spatial resolution, its electrodes are relatively small and less sensitive, with a lower Signal-to-Noise ratio (SNR). On the other hand, $C3_k$ has the highest sensitivity,

[4] me: metal, sa: salt, ma: marble, wo: wood, gl: glass, wa: water, sf: Styrofoam

but also the lowest spatial resolution. The $C2_m$ configuration is placed in between $C1_n$ and $C3_k$ and therefore seems to combine for the given setup the needed spatial resolution and increased sensitivity in an optimal way.

## 5 Conclusion

In this work we have summarized and compared our previous work about two examples of beneficially using ANNs in combination with our tactile proximity sensors. Specially we did focus on the proximity sensing ability of these sensors.

In the first approach [5] we could show, that distance classification can be done using an ANN in a very straightforward way. For the given setup this approach yields better results regarding the measurable distances. With the original model-based approach we could detect and identify distances up to somehow 8cm. The ANN approach described here, nearly doubled the range and allowed us to extended the feasible measuring range up to 14cm.

The second approach [6] is targeting material recognition. It makes use of the abilities of our sensors to measure with multiple frequencies and to flexible combine electrodes allowing to configure the spacial resolution of the sensors. Correspondingly to the distance approach an artificial neural network for the classification was used. The essence of the work was to perform the measurement with various combination of the sensor features which can provide sufficient information to recognize material without further information. For the evaluation, data has been collected by measuring seven object of different materials with variation of electrode combinations and two driving frequencies. The best recognition result with an accuracy of 94% was reached through the $C2_m$ electrode combination and both frequencies. The experiment shows that the combination of various electrode shapes and driving frequencies is promising for material recognition.

Since all experiments were done in the laboratory by constant ambient temperature and humidity. Further investigation in operating conditions should also be done.

# Automated Item Picking - Machine Learning Approach to Recognize Sports and Fashion Articles

Moritz Weisenböhler, Michael Walz, and Christian Wurll

Karlsruhe University of Applied Sciences
moritz.weisenboehler@hs-karlsruhe.de
christian.wurll@hs-karlsruhe.de

**Abstract.** Automated Item Picking (AIP) systems can be used to improve the quality of order picking processes and to reduce the cost. This work presents an AIP solution for the sports and fashion sector. The system consists of an industry robot, a customized gripper and an AI-based perception. A convolutional neural network is trained for identification and localization of various apparel articles and shoeboxes under real conditions.

**Keywords:** automated item picking, computer vision, object detection, machine learning, neural networks

## 1 Introduction

With the worldwide increase in sales figures for e-commerce, the challenges for associated logistics are also rising. Nevertheless, many logistical fulfillment processes are still largely manual and labor-intensive. The use of automated order picking systems for repetitive processes can increase the quality and flexibility of the process, while reducing the costs. However, these systems are rarely used due to the great challenges in terms of perception and the required flexibility. In many applications a large variety of objects has to be handled. The rapid progress in AI can offer appropriate solutions for these challenges.

In a cooperation project between the Karlsruhe University of Applied Sciences, Adidas AG, KUKA AG, Zimmer GmbH and Roboception GmbH a prototypical solution is developed to automate the order picking process for sports and fashion articles. The primary goal is to fully automatically detect and pick shoeboxes and plastic-wrapped apparel from a tote. Special challenges here are handling shoeboxes with opened lids, the chaotic arrangement in the tote and the high number of inhomogeneous articles. This paper focuses on the use of machine learning methods for object detection.

## 2 Related Work

The breakthrough of machine learning in image processing was achieved by Krizhevsky et al. in 2012. With AlexNet, a Convolutional Neural Network (CNN) for image classification, they clearly won the prestigious competition ImageNet Large Scale Visual Recognition Challenge (ILSVRC) [1]. Numerous further developments followed, such as the *GoogLeNet* designed by Szegedy et al. with the so-called *"Inception"* architecture, which won at the ILSVRC 2014 [2].

In addition to image classification, object localization is a central task in computer vision. While image classification only assign a label to the whole image, localization approaches detect the position and class of multiple objects in an image. The region-based CNN (R-CNN) developed by Girshick et al. laid the foundation for modern object localization [3]. Further developments such as the *Fast R-CNN* [4] and *Faster R-CNN* [5] improved in particular the recognition performance and the speed of the method. Besides the classification and object localization, segmentation is another possible application of CNNs. He et al. developed an architecture known as *Mask R-CNN*, allowing segmentation at pixel level. *Mask R-CNN* is an extension of the *Faster-R-CNN* algorithm [6].

Today CNNs have become the standard for most object detection tasks, as the winners of well-known Amazon Robotics Challenge (ARC) illustrate [7]. For example the winning teams of 2016 (TU Delft) and 2017 (Queensland UT) both used CNNs for perception [8] [9].

## 3 System Overview

The developed automated item picking system consists of the following components: the robot (control) and the PLC, the hybrid gripper and the perception system, all managed by the central flow control (see Figure 1).

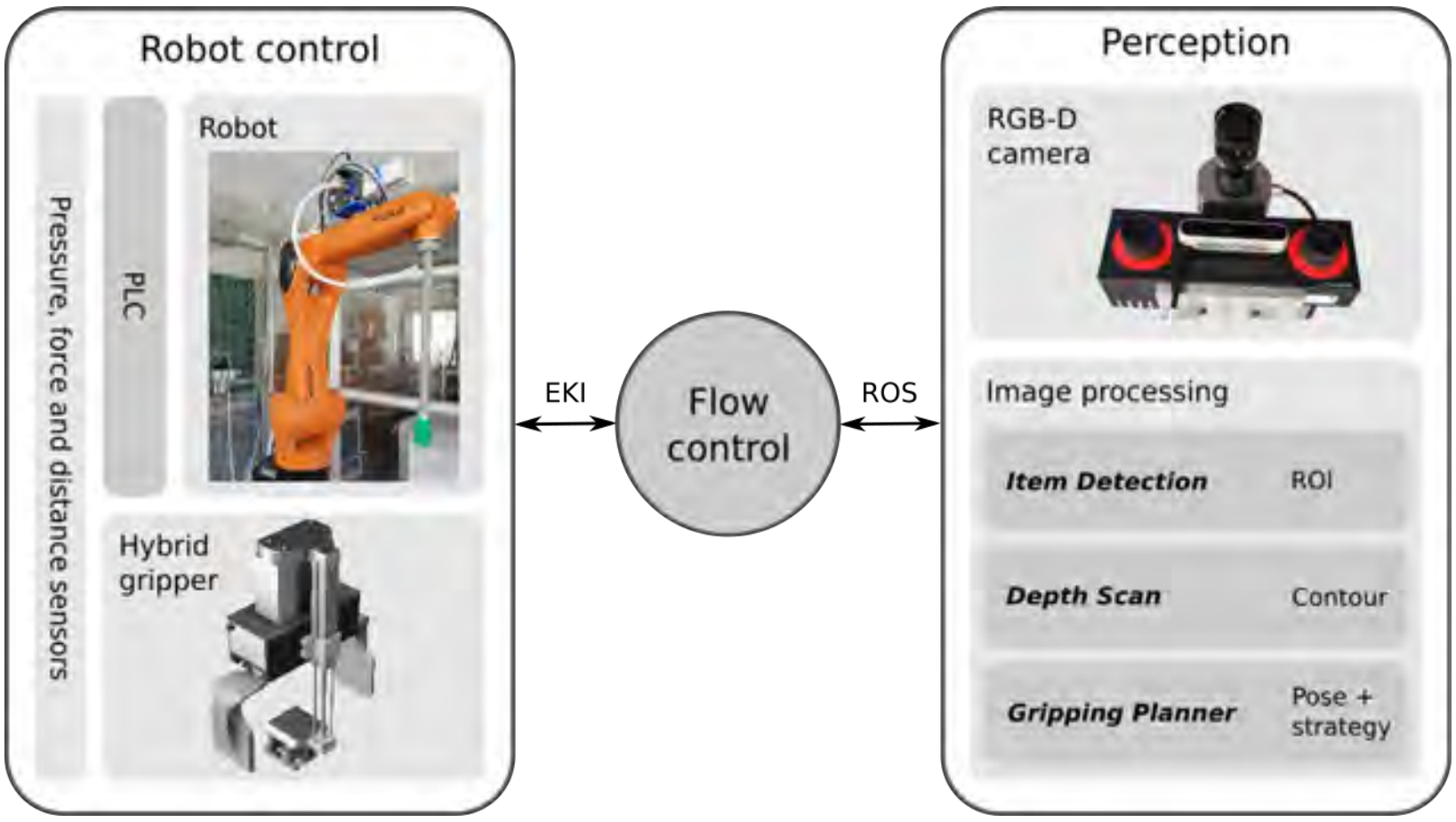

Fig. 1: Schematic system overview.

A KUKA *KR 10 AGILUS* is selected as the main actuator. With a payload of 10 kg, this robot is able to handle all relevant products and is fast enough to reach the desired cycle times, while having a good cost-benefit ratio. All sensors, including pressure, force and distance control, are connected to a Beckhoff PLC directly interfacing with the robot. The robots movements are not programmed directly in the robot control, but provided by the flow control over an Ethernet interface (EKI).

The gripper is designed for the problem specific items: plastic bags and shoeboxes. Especially the handling of the shoeboxes with unfixed lids was a main challenge for the system. Based on the *Zimmer* pneumatic 2-jaw gripper *GP400*, a hybrid gripping system with an extendable cylinder, multiple suction cups and two coated clamping jaws is developed and evaluated. The resulting gripper can handle various types of plastic-wrapped clothes and perform orientation related shoebox picks without opening lids.

Being designed very modular, the perception system allows using any depth sensor, providing RGB images and a Robot Operating System (ROS) interface for image acquisition. Mainly used and tested are the *Roboception rc_visard 160* with a RandomDot projector and the *Intel RealSense™ D415*. To calculate the optimal gripping item and gripping position the acquired images are passed through a multi step process. In a first stage the *Item Detection* module proposes a region of interest (ROI) by a CNN detecting individual items in the RGB image. The ROI is transferred to the *Depth Scan* module, which uses the depth image to determine exact contour if the item. Finally, the *Gripping Planner* selects the best positioned item, calculates its pose and determines an appropriate movement strategy depending on the environment.

# 4 Item detection with machine learning

The goal of the perception system is to identify and locate items in a tote. Therefore, different CNNs are trained and evaluated by a total of 15 different sports articles. The machine learning solution is based on state-of-the-art models and realized with Google's *TensorFlow* framework.

## 4.1 Data collection

The data collection process is based on the experience of the Delft Robotics team, winner of the ARC 2016. Hernandez et al. divide the process into two different subgroups, a basic and a tote set, in order to train the CNN in two stages [10]. The basic set only contains images of individual objects and is used to train the CNN on all generic object characteristics. Afterwards, the pre-trained CNN is fine-tuned with the tote set consisting of scenes of multiple objects placed in one container. A total of more than 8,500 RGB images (8,200 basic and 300 tote) are produced. The setup for data generation contains a *Roboception rc_visard 160* and an *Intel Real Sense SR300* 3D camera, as well as individual controllable lighting modules.

## 4.2 Training

Various state-of-the-art neural networks are trained and tested, with three models being considered in more detail: SSD MobileNet, Faster R-CNN Inception and Faster R-CNN Inception ResNet. SSD MobileNet is a high speed model specially developed for mobile applications. The latter two are based on the processing-intensive but more precise Faster R-CNN approach. They differ by using different submodels to classify the regions of interest.

As explained in subsection 4.1, all training processes are divided into a pre-training and a fine-tuning part. The first part typically takes several hours up to a day. Due to the comparatively small training set the latter part finishes in less than one hour. All neural networks are trained on a single *Nvidia GeForce GTX 1080 Ti* graphics card.

# 5 Results

The three most relevant models are compared in terms of inference speed (duration on GPU and CPU) and accuracy (see Table 1).

Table 1: Comparison of three CNN-models for object localization

| Criteria | SSD MobileNet | Faster R-CNN Inception | Faster R-CNN Inception ResNet |
|---|---|---|---|
| Duration CPU | 0,174 s | 1,532 s | 20,353 s |
| Duration GPU | 0,020 s | 0,109 s | 0,495 s |
| Precision mAP | 41,19 % | 82,12 % | 92,01 % |

In terms of inference speed, SSD MobileNet shows the strength of its comparatively lightweight design. Even without a GPU, the inference takes less than 0.18 seconds processing time per image. The Faster R-CNN Inception takes about 1.5 seconds for a calculation, which is almost nine times longer than SSD MobileNet. On average, Faster R-CNN Inception ResNet takes more than 20 seconds for a calculation. This is obviously too slow in terms of a targeted detection cycle time of about one second. Here the use of a GPU is required for shortening the processing time to about half a second.

For the evaluation of the detection accuracy, an excluded test set of roughly 10% of the total images are used. With a mAP of only 41%, the fast SSD MobileNet is not suitable for a secure item detection. Better results are achieved using the Faster R-CNN networks. Inception ResNet achieves the best mAP of 92%, while Inception scores slightly lower (mAP of 82%).

The Faster R-CNN Inception ResNet provides good forecasts for scenes containing deformable plastic-packed articles (see Figure 2a). A high recognition reliability can be achieved in particular if the objects lay clearly visible in the tote. In the case of concealed or strongly deformed objects, the CNN shows weaknesses. However, even for humans, it is difficult to discern all objects in these constellations. For scenes with several shoeboxes the Faster R-CNN Inception ResNet shows also good results. These are reliably localized even for challenging arrangements with boxes lying very close to each other or opened lids (see Figure 2b).

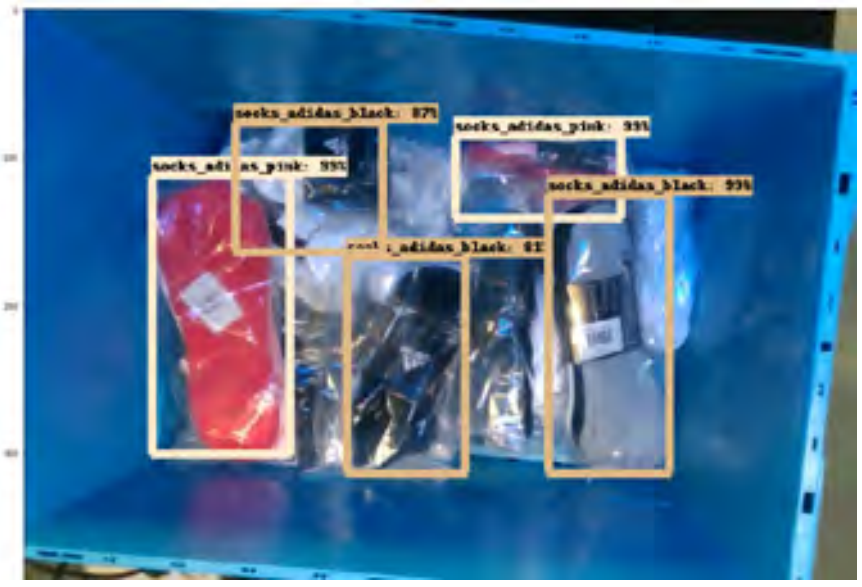

(a) Tote scene with socks.

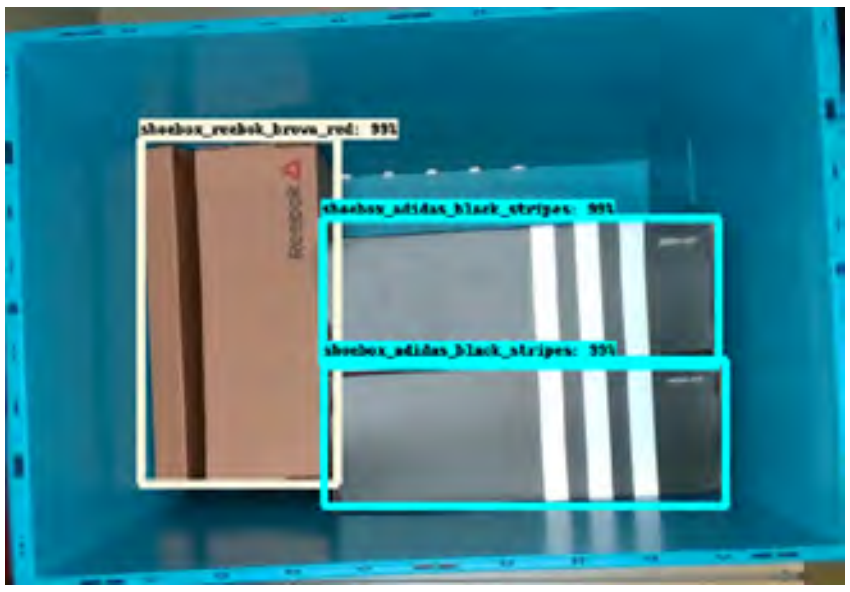

(b) Tote scene with shoeboxes.

Fig. 2: Example results for Faster R-CNN Inception ResNet.

The use of a orthogonal bounding box is a disadvantage for the precise contour detection of inclined shoeboxes. Although the detected object is completely enclosed by the bounding box, it does not provide enough information about the exact location and rotation of the object. This is why the downstream Depth Scan module is required (see section 3). The algorithm uses the corresponding depth image to determine the contour of an item inside the detected ROI.

## 6 Conclusion

The present paper examined machine learning techniques to detect shoeboxes and deformable plastic-packed articles in realistic scenarios. Different neural networks were trained and evaluated to overcome the analyzed challenges. Based on the Faster R-CNN Inception ResNet, the developed solution was able to reliably detect and locate different objects by means of bounding boxes. The reproducible learning process can be quickly transferred to new objects with sufficient training data available.

Further investigations will concentrate on the reliabilities of different degrees of item distinction categories, the detection of unmanageable arrangements and items (e.g. opened shoeboxes) and the detection of shoebox poses. Especially the latter task requires the detection of the lid and the direct involvement of the corresponding depth data.

# Intelligent Grinding Process via Artificial Neural Networks

Mohammadali Kadivar and Bahman Azarhoushang

Institute of Precision Machining (KSF), Furtwangen University of Applied Sciences
**kamo@hs-furtwangen.de**
**aza@hs-furtwangen.de**

**Abstract.** Surface roughness of the ground parts and the grinding forces are two important factors for the assessment of the grinding process. The surface roughness directly influences the functional requirements of the workpieces and the grinding forces are an important criterion for the achievable material removal rate. The establishment of a model for the reliable prediction of surface roughness and grinding forces is a key issue. This work deals with design of appropriate control strategy for prediction of grinding forces and surface roughness as one of the important indicators of the machined surface quality via applying Artificial Neural Networks (ANNs) through special sensors integrated into the machine tool. A micro-grinding process of Ti6Al4V was chosen. The model was verified by various experimental tests with different grinding and dressing parameters. It was found that the predictions made by the ANN model matched well with the experimental results.

**Keywords:** Artificial Neural Networks (ANNs), Micro-grinding, Surface roughness, Modeling, Intelligent grinding

## 1 Introduction

Mechanical micro-cutting is one of the key technologies to enable the realization of high accuracy complex micro-products made from a variety of engineering materials. Amount mechanical micro-machining process, micro-grinding has been an effective process to achieve high dimensionally accurate parts in machining process with superior surface finishes. However, modelling of the micro-grinding, especially predicting micro-grinding forces and surface roughness in a very small size is complicated and is still at its early stage. Most of the analytical models are adapted from conventional approaches but taking one or more size effects into consideration. The size effects which have been modelled to predict micro-cutting forces include ratio of feed rate to tool radius; cutter edge radius [1]; minimum chip thickness [2]; and micro-structure effect.

Research has been carried out in micro-cutting mechanics for decades and experimental studies still dominate the micro-cutting research. Limited researches, which are dealing with the fundamental understanding of the material removal mechanisms in the micro-scale (single grain-workpiece interaction), are available [3–5]. Moreover, the micro-scale numerical modelling methods are developed to describe the plastic behavior of the workpiece material at the high temperature and strain-rates linked with the grinding process. Cheng et al. [6] presented a mathematical model for the prediction of the micro-drill-grinding force. Park and Liang [7] modelled the micro-grinding forces based on the physical analysis of the process.

In all modeling studies the prediction error for the surface roughens in grinding process is very high since the grinding process is a complex process. The grinding grits on the surface of the grinding tool are stochastically distributed and it is almost impossible to find two same abrasive grit with the same shape and cutting edges, making the process very complex to be modeled. There are also several parameters which are influential in the modeling of the grinding forces and surface roughness such as: vibration, the precision of the machine tool, the tool specification, dressing parameters and other parameters which is so complicated to import all of them into the modeling process and consider their effects.

Using the ANNs is a desirable way to model such complex systems. Vrabel et al. [8] used the ANNs to predict the surface integrity in the machining process. They developed and tested a neural network which was able to predict the drill flank wear to prevent anomalies occurring on machined surface. Beatrice et al. [9] studied the ability of modeling the surface roughness (Ra) in terms of cutting parameters during hard turning of AISI H13 tool steel with minimal cutting fluid application. They showed that the ANN model can be a useful tool to select the cutting parameters for achieving desired surface finish. Tomaz Irgolic et al. [10] used the feed-forward backpropagation neural network to predict the cutting forces in milling operation. They proved that the prediction of the cutting forces with the ANNs was very reliable; the error in predicting cutting forces was smaller than 10%.

In this work several experiments at different conditions with various grinding parameters, i.e. cutting speed, feed rate and depth of cut and dressing parameters such as dressing overlap ratio and dressing speed ratio were carried out. Using the experimental data, obtained from the grinding process, two different neural networks were trained to model and predict the grinding forces and surface roughens. Finally, the models were validated and tested with some other experimental data.

## 2 Methodology

First the desire sensors were integrated into the machine tool for measuring the required out-put parameters during the experiments. This was an important part of this work since the errors which occur in the measuring of the outputs will directly influence the prediction of the outputs via the ANNs. To this end the titanium alloy “Ti-6V-4Al” was chosen as the workpiece material. Block samples with the dimensions of 30x20x10 mm were ground using vitrified-bonded diamond grinding pin (D46C150 V) with the diameter of 2 mm. The grinding pin was dressed using a diamond dressing roll with the diameter of 100 mm. Oil was utilized as the grinding fluid. The micro-grinding tests were carried out at different cutting speeds ($v_c$), dressing-speed ratios ($q_d$), and dressing overlap ratios ($U_d$) to investigate the effect of these parameters on grinding forces and surface roughness. A high-precision 5-axis CNC machining center was used for the experiments (Fig.1).

**Table 1.** Process Parameters

| **Grinding Parameters:** | **Values** |
|---|---|
| Cutting Speed $v_c$ (m/s) | 6, 10, 12 and 14 |
| Feed Rate $v_{ft}$ (mm/min) | 200 and 1000 |
| Depth of Cut $a_e$ (µm) | 4 and 10 |
| Width of Cut $a_p$ (mm) | 2.5 |
| **Dressing Parameters:** | **Values** |
| Dressing Depth $a_{ed}$ (µm) | 2 |
| Dressing Speed Ratio $q_d$ | -0.4, +0.4 and +0.8 |
| Dressing Overlap Ratio $U_d$ | 45, 90, 270, 305, 910, 1830 |

To measure the forces and surface roughness, a type 9256C2 dynamometer and a tactile surface roughness tester (Hommel-Werke model T-1000) were used, respectively. The surface roughness measurements were taken perpendicular to the grinding direction at three positions: at the beginning, at the middle, and at the end of the grinding path. A confocal microscope (µsurf mobile plus) was used to obtain the confocal pictures from the ground surface. Each test was repeated three times. Table 1 lists the utilized process parameters for the experiments.

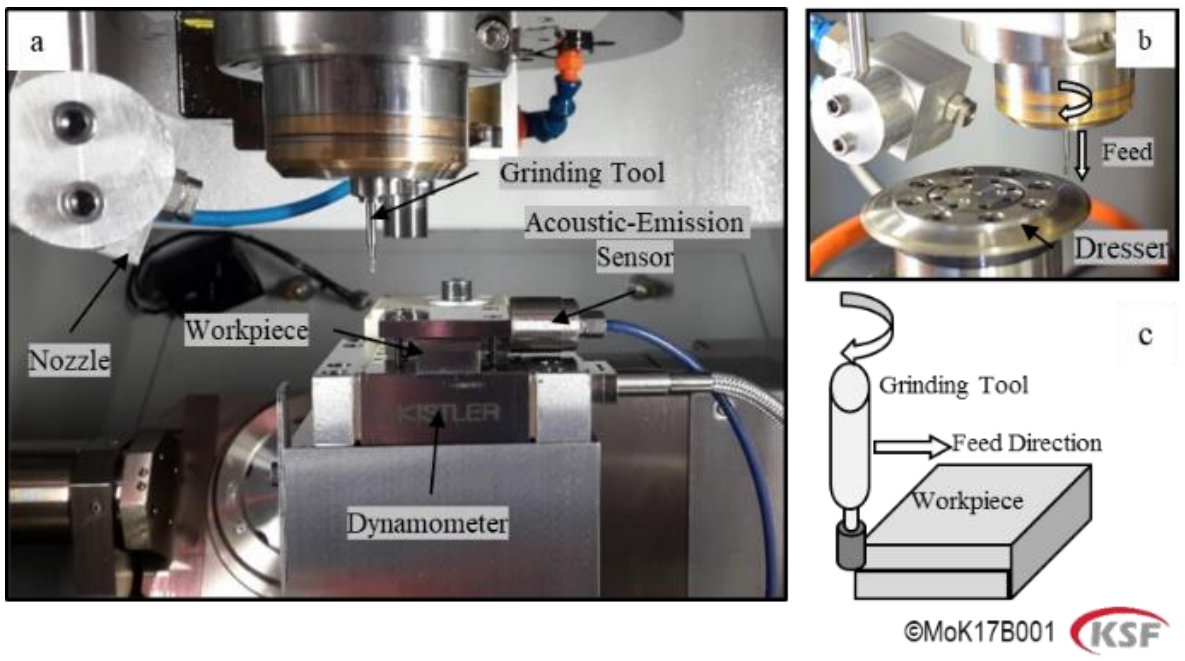

**Fig. 1.** a) Experimental setup, b) Dressing setup, c) Illustration of the micro-grinding process

## 3 Neural Networks

The appropriate architecture for the ANN was selected through a comprehensive examination of several network configurations. This was accomplished by changing the number of hidden layers and number of neurons in the hidden layers. The hidden layer plays an important role in modeling of the process via the neural networks and has an optimal quantity. The low number of the neurons in the hidden lower may cause the high Sum Square Errors (SSE) and increasing their number reduces the SSE up to a certain point that it becomes stable. After that the SSE can be fluctuated and even increases by increasing the number of the neurons [11].

A routine that utilizes a feed forward back propagation algorithm was used to develop the model as it is widely used by other researchers, since this method generally leads to the most accurate results. The feed forward back propagation is considered to be a powerful technique for constructing non-linear functions between several inputs (such as cutting speed, feed, depth of cut) and one or more corresponding outputs (such as the cutting forces). The back-propagation network typically has an input layer, an output layer and at least one hidden layer, with each layer fully connected to the succeeding layer. During learning, information is also propagated back through the network and used to update the connection weights. The following expressions give the basic relationships used for this analysis [12]:

$x_q^{[s]}$ = current output state of the $q^{th}$ neuron in layers.

$w_{qp}^{[s]}$= weight on the connection joining the $p^{th}$ neuron in layer s-1 to the $q^{th}$ neuron in layer s.

$I_q^{[s]}$= weighted summation of inputs to the $q^{th}$ neuron in layer s.

A back-propagation element therefore propagates its inputs as:

$$x_q^{[s]} = f\left(\sum_p^{[s]}(w_{qp}^{[s]} * x_q^{[s-1]})\right) = f\left(I_q^{[s]}\right) \qquad (1)$$

The MATLAB ANN toolbox was used for easily updating the value of weights and biases of the algorithm. Networks with different architecture were trained for a fixed number of cycles and were tested using a set of input and output parameters. The sigmoid function is used in this study. Fig. 2 shows a neural network structure for learning process with a sigmoid function for the hidden layer and a linear transfer function for the output layer.

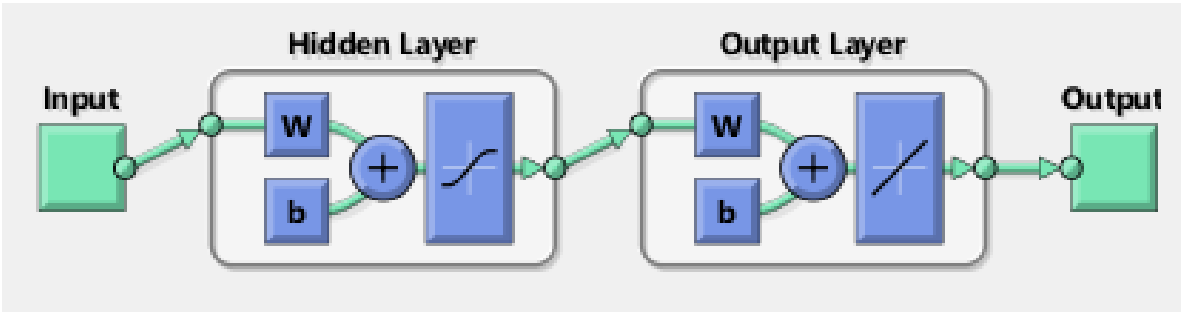

**Fig. 2.** The schematic of a neural network for learning

## 4 Results and discussion

In the modelling 80 percent of the gathered data has been chosen for training, 10 percent for validation and 10 percent for testing of the network. Different hidden layers with different number of neurons was chosen to obtain the optimal hidden layer.

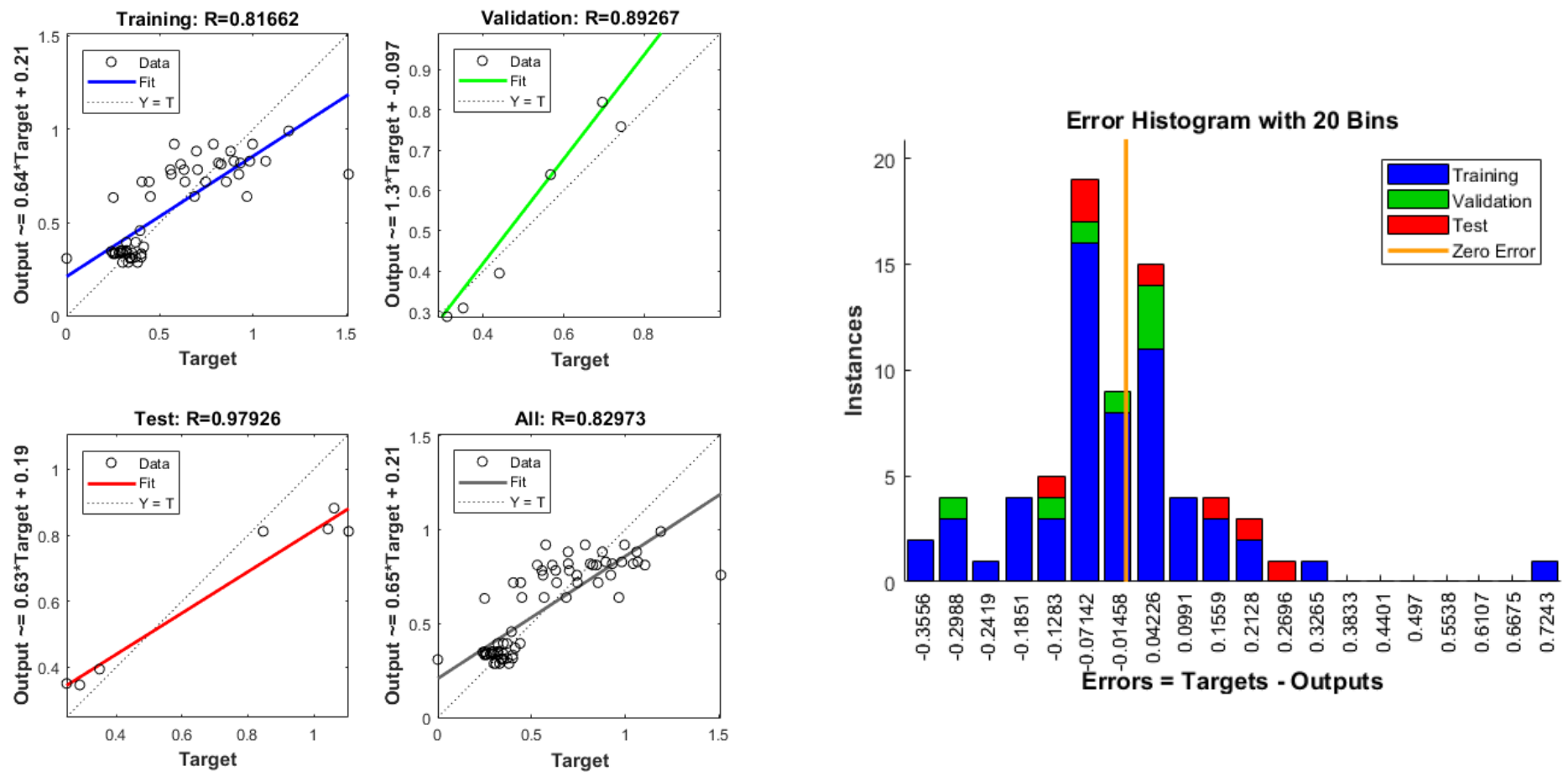

**Fig. 3.** left) The regression for the trained and test data for $R_a$, right) the Error Histogram for $R_a$

For modeling of the surface roughness the number of the hidden layers has been set equal to the number of the inputs and the results have been given in Fig. 3-left. As it can be seen in the figure, all the phases (training, testing, and validation) have an acceptable error which shows that the developed ANN model can precisely predict the surface roughness in the term of $R_a$. The training phase has an SSE of 0.82 which shows that all data are trained sufficiently. After the training the trained model has been validated with the 10 percent of the gathered data. The validation has also an acceptable error (SSE of 0.89). The model was tested furthermore with another 10 percent of the gathered data which showed a very good ability of predicting of the surface roughness via ANNs (SSE of 0.97).

Fig. 3-right represents the Error Histogram which is calculated as the difference between the targets of the neural networks and the actual outputs. The graph also shows that most of the errors in all stages are around the middle of the histogram curves which also validate the training stage. Fig. 4 shows the trend of the learning process. At the beginning of the learning process the error is high and the model correct himself by modifying the weights until a stable steady situation has been achieved. It is also clear that the all stages including training, validation and testing are approximately in the same order.

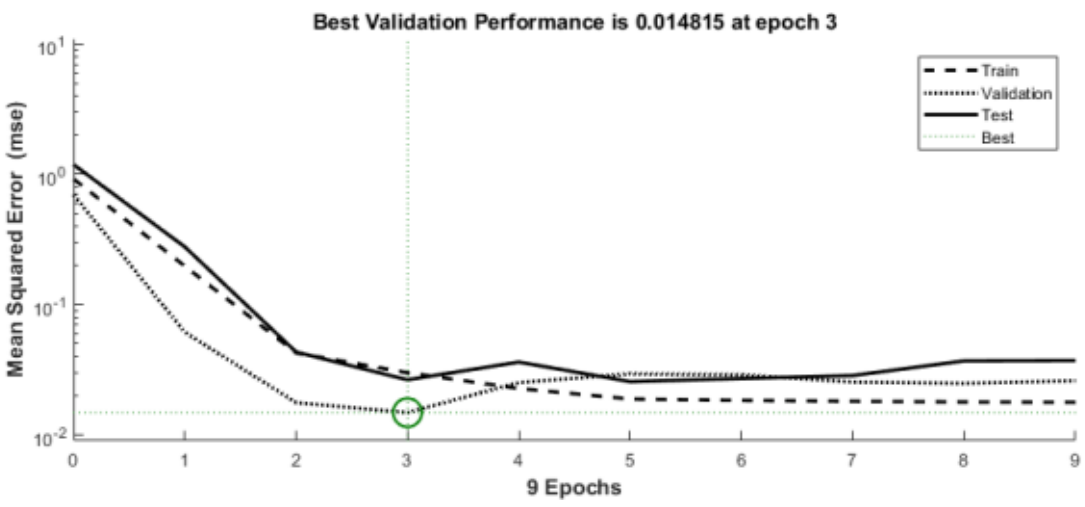

**Fig. 4.** the modeling trend versus the learning Sycle for $R_a$

To model the grinding forces first the number of neurons in the hidden layer has been set to 10 and the errors have been listed in the Fig. 5-left. The results show that the ANN also can precisely predict the grinding forces with an acceptable error. The histogram graph also shows that most of the errors in all stages are around the middle of the histogram curves which validate the training stage (Fig. 5-right).

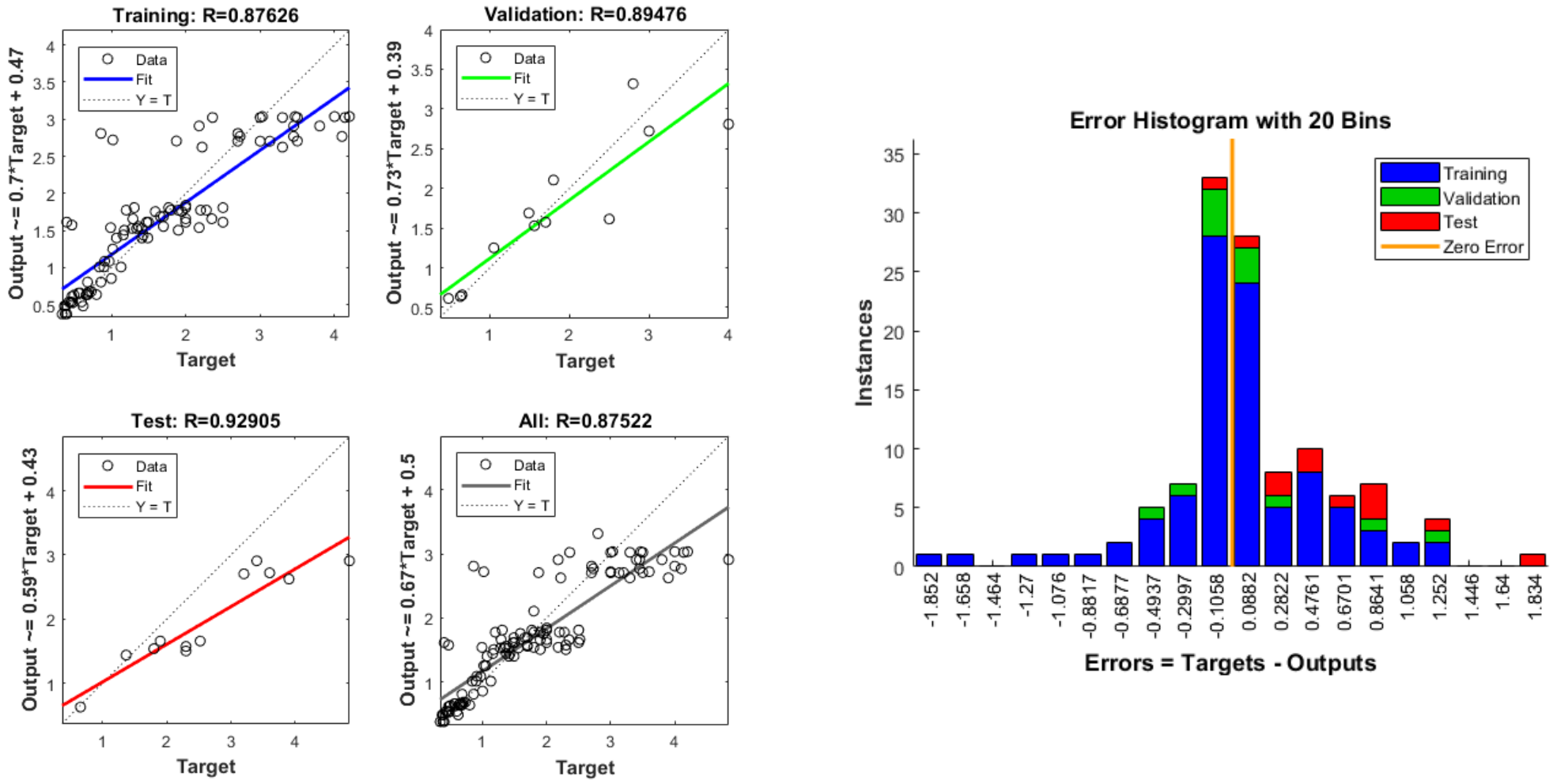

**Fig. 5.** left) The regression for the trained and test data for the grinding forces with 10 neurons in hidden layer, right) the Error Histogram for the grinding forces

## Conclusion

A neural network with different structures was designed and trained to predict the grinding forces and surface roughness of micro grinding of titanium. The cutting forces and surface quality was measured via integrated sensors into the machine tool. It was observed that the accuracy of the training phase as well as testing highly depends on an optimized number of the neurons in the hidden layer. The results showed that the ANNs are capable to model the grinding forces and surface roughness of titanium material with an acceptable accuracy. The results of this study can be used to monitor the process online. As a future work it is suggested to use acoustic emission as an additional sensor to online monitor the process, enabling the prediction of the surface roughness with the help of ANNs.

# New Tool to Determine the Safety-Parameters Based on Safety Standard

Ossmane Krini[1], Josef Börcsök[2], Agnieszka Moranda[1], and Tanja Weiser[1]

[1] Cooperative State University Lörrach (DHBW)
krini@dhbw-loerrach.de
[2] University of Kassel

**Abstract.** FRCaS is a newly developed Software-Tool to calculate the failure rates from diverse components. This tool offers the user the opportunity to determine the failure rate for diverse components on the basis of various Standards of Siemens SN 29500, Military Handbook (MIL-HDBK 217F) and CENELEC. The user is able to obtain results with this tool without previous knowledge of the details of the applied standards. A further development and objective is the integration of the Program Package OrCAD into the developed Software-Tool, in order to determine the failure rates of individual components, which are developed in OrCAD electronic circuits.

**Keywords***:* Reliability, Failure rate, Safety-Standard, Safety-Parameters, Failure probability.

## 1 Introduction

The current research work in the department of computer architecture and system programming at the University of Kassel, developed a calculation tool to determine the expectation values of components. This tool provides the ability to determine the failure rate of components for reference and operating condition on basis of the Siemens-Norm SN 29500. The failure rate plays an important role in reliability and is normally highly dependent on the working conditions. The values that are given in the standard reference conditions are determined default rates resulting from long experimental observations of a large number of identical components, which are used by the Siemens AG as a uniform basis. The failure rate is the risk of one component failing and is normally highly dependent on various factors. The course of the failure rates over the time can be presented by the bathtub curve. By factors such as e.g. the environment, a conversion can be calculated by referring to operating condition.

With the calculation tool, it is possible to represent the inputs and outputs through a user interface. The user takes the opportunity to choose, for example, switches and buttons, as well as alarm and signal lights with the necessary properties. Within an output range, the reference and operating condition can be calculated. Thus, the user has the opportunity to compare, save, and open the result in a data table. A chart in environment and voltage dependence is used to visualize the failure rate in operating condition. The calculation tool is developed on basis of JAVA application.

## 2 Mathematical background

### 2.1 Weibull distribution

The Weibull distribution is one of the most commonly used distributions in reliability engineering because of the many shapes it attains for various values of $\beta$ (shape parameter). It can therefore model a great variety of date and life characteristics. The shape parameter is what gives the Weibull distribution its flexibility. By changing the value of the shape parameter, the Weibull distribution can model a wide variety of data. The Weibull distribution makes it possible to represent time dependent failure probabilities $F(t)$ of components. For this it is necessary to possess the determined function parameters from observed data. In principle these also have a technical important meaning. From these data it is possible to determine whether we are dealing with early, random or aging failures. The required data is failure frequency, number of all components and failure times of the components. Also the Weibull distribution assumes for its application the simplified assumptions that single component failures are independent of each other. The probability of failure according to the Weibull distribution is defined as

$$F(t) = W(t;\beta;t_0;T) = 1 - e^{-\left(\frac{t-t_0}{T-t_0}\right)^{\beta}} \quad (1)$$

where $t$ is the time, $\beta$ is the shape parameter, $t_0$ is the correction parameter and $T$ is the characteristic lifetime or position parameter.
The probability density function of a Weibull distribution is given by

$$f(t) = W(t;\beta;t_0;T) = \frac{\beta}{T-t_0} \cdot \left(\frac{t-t_0}{T-t_0}\right)^{\beta-1} \cdot e^{-\left(\frac{t-t_0}{T-t_0}\right)^{\beta}} \quad (2)$$

If $\beta = 1$, the Weibull distribution is identical to the exponential distribution. If $\beta = 2$, the Weibull distribution is identical to the Rayleigh distribution. If $\beta$ is between 3 and 4 the Weibull distribution approximates the normal distribution.

## 2.2 Exponential distribution

The exponential distribution is useful in many applications in engineering, for example, to describe the lifetime X of a transistor. Therefore, the exponential distribution is the most known distribution. With this distribution it is possible to represent the time dependent probability $F(t)$ of components for which it is necessary to obtain observed data to determine X. The failure probability is defined by the exponential distribution as

$$F(t) = 1 - e^{-\lambda \cdot t} \quad (3)$$

where $\lambda$ is the failure rate. Respectively with failure density

$$f(t) = \begin{cases} \lambda \cdot e^{-\lambda \cdot t} & for \quad t \geq 0 \\ 0 & otherwise \end{cases} \quad (4)$$

## 2.3 Reliability and failure probability

The reliability $R(t)$ is the probability that a unit is functional in one view period $(0,t)$. Fig.1 shows $R(t)$ as function of time.

The probability that the operational time T is within the considered time interval $(0..t)$ is for small t almost equal to one. For larger values of t the probability decreases more and more.

$$R(t) = e^{-\int_0^{\infty} \lambda(t)dt} \quad (5)$$

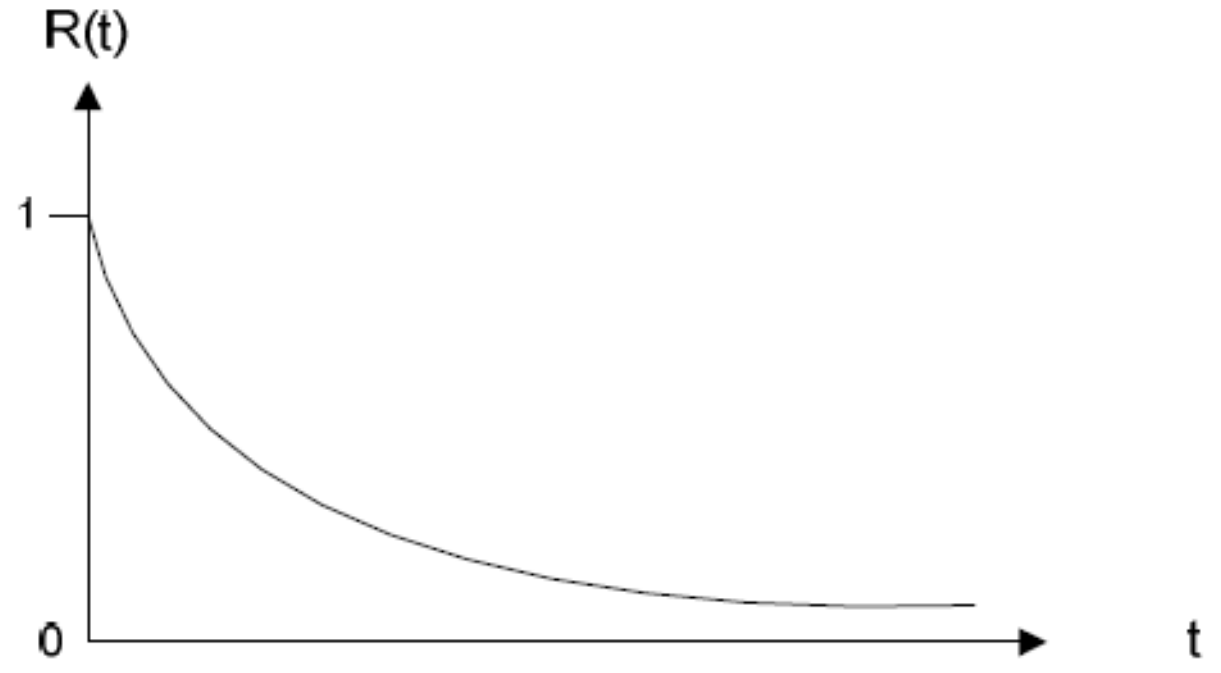

**Fig. 1.** Reliability function $R(t)$ as function of time

If an exponential distribution for the reliability is valid, then the failure rate is constant:

$$\lambda(t) = \lambda \quad (6)$$

Thus the equation can be rewritten as:

$$R(t) = e^{-\lambda \cdot t} \quad (7)$$

An important reliability parameter is the MTTF value (Mean Time To Failure)

$$MTTF = \int_0^{\infty} R(t)dt = \frac{1}{\lambda} \ . \quad (8)$$

Within the interval $(0,t]$ the probability of failure $F(t)$ is calculated applying the reliability function

$R(t)$.

$$F(t) = 1 - R(t) \quad (9)$$

$$F(t) = 1 - e^{-\lambda \cdot t} \quad (10)$$

$$F(t) \approx \lambda \cdot t \quad for \quad \lambda \cdot t << 1 \quad (11)$$

Generally, the time t is applied by T1. The time from point in time zero to time T1 is characterized as proof test interval. At time T1 a periodical test or the maintenance of a safety system is taking place. Tests are carried out to allocate undetected, dangerous failures. After a proof test, the system is regarded as new. The calculated PFD-value depends on the value T1.

# 3 Failure Rate Calculation Tool FRCaS

Many organizations use the standards to perform, for example, the calculation of PFD-values of components. However, the reference conditions do not always apply, for example, components are temperature dependent and the temperature changes with time. When this happens, the reference conditions are damaged.

## 3.1 Objectives of the Tool

The objective is the new development of a Software-Tool, which calculates the failure rates of components.

This Software Tool offers the user some of the basis of following Standards

- Siemens SN 29500
- Military Handbook MIL-HDBK 217F
- CENELEC

The failure rate for the various components is determined. The user is able to use this Tool without knowing exact details of the standards, and obtain valid results thereof. Next, the Program package OrCAD should be included with the Software-Tool in order to determine the failure rate of the components of the developed circuit in OrCAD.

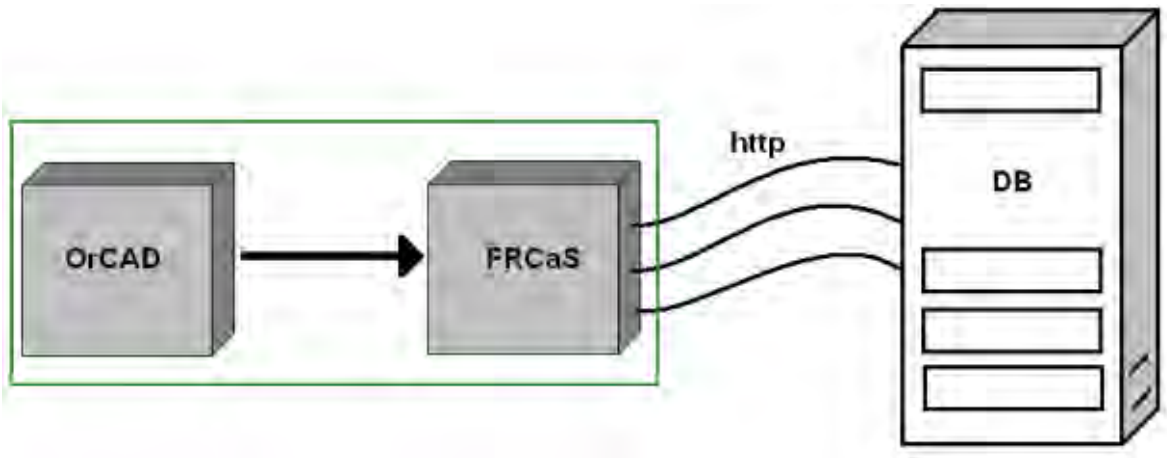

**Fig. 2.** Assembly of the Structure

The FRCaS-Tool should offer the user the opportunity based on this standard, to determine the failure rates of components in operating condition. There are two types of components: local components and components already in the Databank. In order to stimulate a component in the Databank, one must be the authorized user. First, the authorized user can determine the failure rate for the operating condition of chosen components. In addition, the opportunity exists that the FRCaS Tool has the ability to connect with the program package “OrCAD“. The following picture shows the relationship between the OrCAD, the FRCaS-Tool, (Clients) and the Databank. The Databank, which is accessible by www, manages neither the components nor the assignment of user privileges.

## 3.2 Calculation of failure rates on basis of SN 29500

The Siemens-Norm SN 29500 is composed of a total of 15 parts and is used to determine the failure rates of components. The titles of the individual parts of the Siemens-Norm SN29500 are as follows:

**SN29500-2:** Expected values for integrated circuits. This Part is used for reliability calculations such as for example, memory, microprocessors, digital and Family analogs ASICs.

**SN 29500-3:** Expected values for discrete semiconductors. The failure rates apply for leaded and SMT- components. This part is used for reliability calculations such as e.g. Transistors, diodes, and power semiconductors.

**SN 29500-4:** Expected values for passive components. The passive components belong to groups such as capacitors, resistors, inductors and other passive components. The failure rates apply for leaded and SMT- components.

**SN 29500-5:** Expected values for electrical connections, electrical connectors and sockets. This part is used for reliability calculations such as clips, screws, or coaxial.

**SN 29500-6:** withdrawn; content has been integrated in part 5 and 12.

**SN 29500-7:** Expected values for relays. Relays are used in the field of control engineering, data processing, telecommunications and automotive electrical systems. For these applications there are types of relays such as low power relay, general relay, or automotive relays.

**SN 29500-8:** withdrawn

**SN 29500-9:** Expected values for switches and buttons. This part is used for reliability calculations of telecommunication and electronic products with high reliability requirements. There are switches and buttons used e.g. Dipfix or coding switches, buttons and switches for low power applications or higher electrical power handling with mechanical contacts. It is used in the field of control engineering, data processing, communications and process technology. It does not cover installation and high-voltage switch.

**SN 29500-10:** Expected values for signal and pilot lamps. This part is used for reliability calculations in reporting and signal lamps such as incandescent and neon lamps. The failure rates apply for soldered screwed and plugged devices.

**SN 29500-11:** Expected values for contactors. The application focus of this part is the control technology and switching of motors. This part is used for reliability calculations such as equal- and alternating current protectors.

**SN 29500-12:** Expected values for optical components. This part is used for reliability calculations such as optical semiconductors, light emitting diodes, opt couplers, optical sensors, transceivers, transponders, and optical subsystems.

**SN 29500-13:** withdrawn; content has been integrated in part 12.

**SN 29500-14:** withdrawn; content has been integrated in part 12.

**SN 29500-15:** Expected values of electromechanical protection devices. They are applied in areas of control and switchgear technology. For these areas of application there are electromechanical protecting devices such as circuit breaks for motor protection, overload relays, circuit breaks and residual current protection devices.

## 4 Using the Tool of loss in reference and operating conditions on the basis of Siemens Norm 29500-04

The Siemens-Norm part two is analyzed and implemented within the new Tool. Fig. 3 shows the assembly of the Tool.

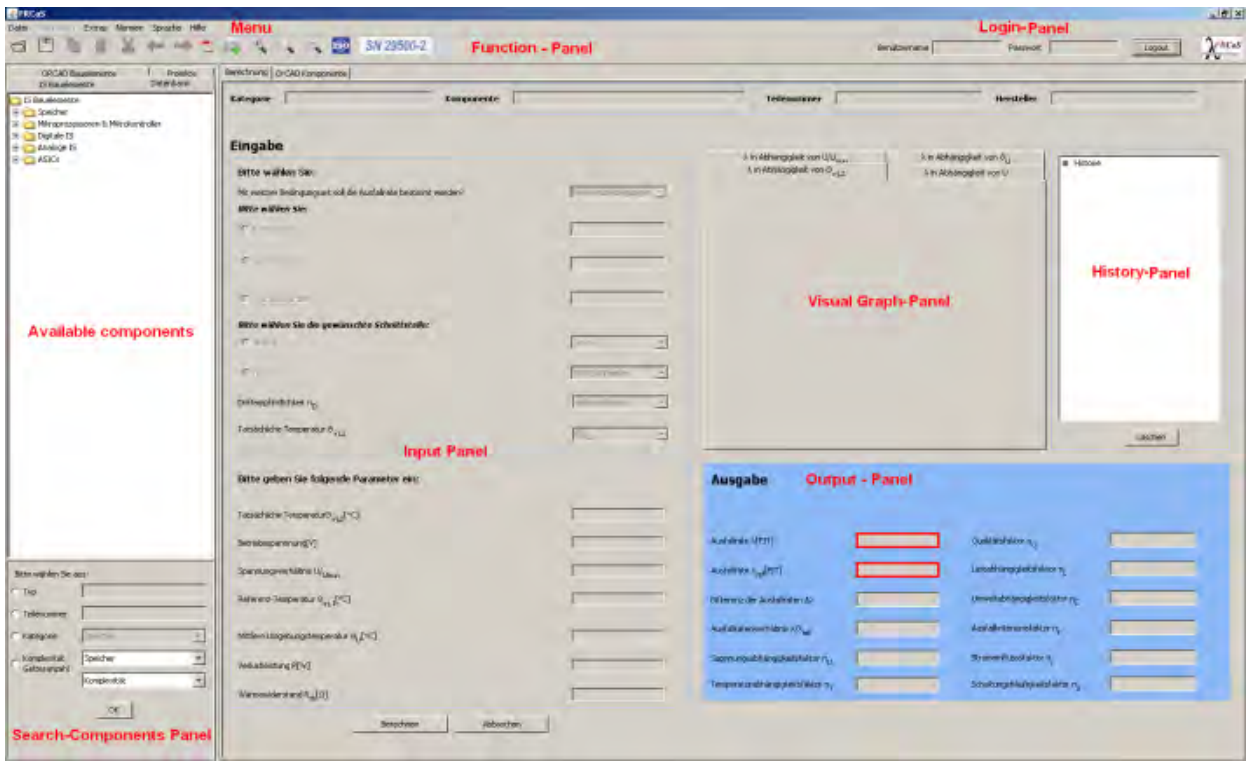

**Fig. 3.** Assembly of the new Tool

## 4.1 Calculation of failure rate with the new tool under the reference conditions

Reference conditions are the values given in a specific environment such as the average environment temperature at 40 $^{\circ}C$ or the operating current. These are the values that prevail without influences.

Failure criterion: Complete failures and changes of major parameters that would lead to a failure in the majority of applications. The failure rates $\lambda_{ref}$ stated in Fig.4 should be understood for operation under the stated reference conditions as expected values for the stated time interval and the entirety of lots. Within the scope of the variations of values, in exceptional lots, the actual value may differ from the expected one a factor of up to two and half.

| **Kondensator/ Capacitor** | | $\lambda_{ref}$ in FIT | $\theta_1$ 1) in °C | $U_{ref} / U_{max}$ |
|---|---|---|---|---|
| ***Metallfolie/ Metal foil*** | | | 40 | 0,5 |
| Polystyrol | KS | 1 | | |
| Polypropylen / *Polypropylene* | KP | 1 | | |
| Polycarbonat / *Polycarbonate* | KC | 2 | | |
| Polyethylenterephtalat / *Polyethylene terephtalate* | KT | 1 | | |
| **Metallisierter Belag / *Metallized film*** | | | | |
| Polyethylenterephtalat / *Polyethylene terephtalate* | MKT | 0,7 | | |
| Polycarbonat / *Polycarbonate* | MKC | 0,7 | | |
| Polypropylen / *Polypropylene* | MKP | 0,7 | | |
| Zelluloseacetat / *Acetyl cellulose* | MKU | 0,7 | | |
| **Metall-Papier(-Kunststoff)** ***Metallized paper (film)*** 2) | MP, MKV 2) | 2 | | |
| **Glimmer / *Mica*** | | 1 | | |
| **Glas / *Glass*** | | 2 | | |
| **Keramik / *Ceramic*** | **NDK / *LDC*** COG, NPO | 1 | | |
| | **MDK / *MDC*** X7R, X5R | 2 | | |
| | **HDK / *HDC*** Z5U, Y5V, Y4T | 5 | | |
| **Al-ELKO / *Aluminium electrolytic*** | | | | 0,8 |
| flüssiger Elektrolyt / *non solid electrolyte* | | 5 | | |
| fester Elektrolyt / *solid electrolyte* | | 3 | | |
| **TA-Elko / *Tantalum electrolytic*** | | | | 0,5 |
| flüssiger Elektrolyt / *non solid electrolyte* | | 10 | | |
| fester Elektrolyt / *solid electrolyte* | | 1 | | |
| **Veränderbare Kondensatoren / *Variable*** | | 10 | | - |

1 FIT = $1 \times 10^{-9}$ 1/h (ein Ausfall pro $10^9$ Bauelementestunden) | 1 FIT equals one failure per $10^9$ component hours

Die genannten Ausfallraten gelten für die Qualitätsklasse LL. Für die Qualitätsklasse GP werden im Abschnitt 4.3 Umrechnungsfaktoren angegeben. | The stated failure rates apply to quality class LL. Conversion factors for quality class GP are given in section 4.3.

1) Kondensatortemperatur | 1) Capacitor temperature

2) Gilt nicht für Leistungskondensatoren nach VDE 0560-T12 | 2) Does not apply to power capacitors according to VDE 0560-T12

3) Der Schaltkreiswiderstand bei Tantal-Kondensatoren hat Einfluß auf die Ausfallrate. Entsprechende Zusammenhänge sind einschlägigen Normen oder Datenbüchern zu entnehmen. Die angegeben Ausfallraten gelten für einen Widerstand von > 3 Ohm/Volt | 3) The circuit resistance of tantalum capacitors influences the failure rate. Corresponding connections are to be taken from the standards in question or from data books. The failure rates given apply to a resistance of ≥ 3 Ohm/Volt.

Durchführungskondensatoren und -filter (DUKO, DUFI) siehe Tabelle 4 | Feed-through capacitors and filters see Table 4

**Fig. 4.** Failure rates for Capacitor

As an example, the failure rate can be determined with the new Tool for operational amplifiers with the Degree of integration. Fig. 5 shows the calculation.

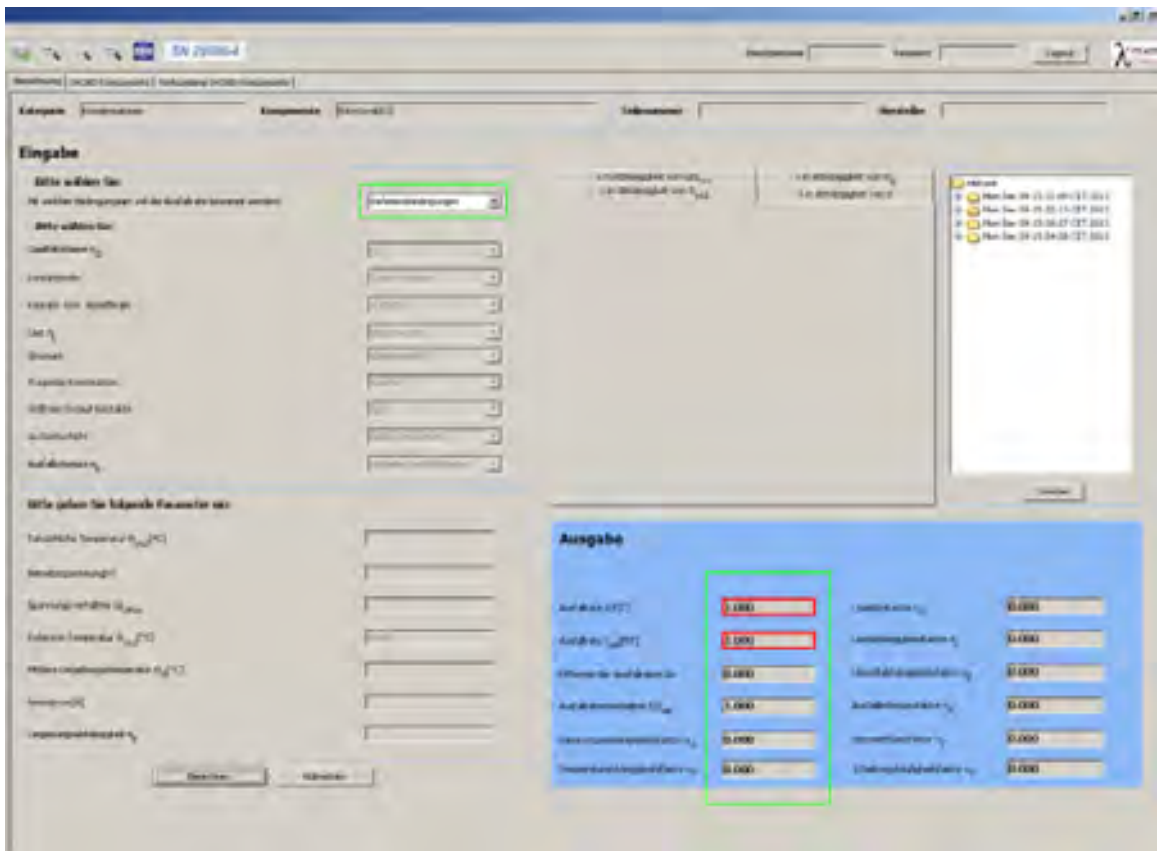

**Fig. 5.** Calculation under the Reference condition

Also, various changes can be made with the Tool. For example, the degree of integration can be changed in the Tool and therefore a fast calculation reference-failure rate can be achieved.

## 4.2 Conversion from reference to operating conditions

An operating condition by a manufacturer for the use of a device specified condition. To calculate for different operating conditions, conversion models are used. They contain constants, which were determined according to the state of the technology. The values given in the Siemens 29500 constants are averages of DIN EN 61709.

If the passive components are not under the electrical stresses and at the average ambient temperature as stated in the reference conditions, the result can be failure rates which differ from the expected value given in for example in Fig. 4. To account for the actual electrical stresses and the average ambient temperature that occur during operation, the expected values under reference conditions need to be converted with the relevant $\pi$ factors. The failure rate under the operating conditions $\lambda$ is calculated for operations as follows:
For capacitors:

$$\lambda = \lambda_{ref} \times \pi_U \times \pi_T \times \pi_Q (12)$$

where $\lambda_{ref}$ is the failure rate under reference conditions, $\pi_U$ is the voltage dependence factor, $\pi_T$ is the temperature dependence factor and $\pi_Q$ is the quality factor.

The voltage dependence factor $\pi_U$ for capacitors is taken into account as the following equation:

$$\pi_U = \exp\left( C_3 \times \left( \left( \frac{U}{U_{\max}} \right)^{C_2} - \left( \frac{U_{ref}}{U_{\max}} \right)^{C_2} \right) \right) \quad (13)$$

| Kondensatoren / *Capacitors* | Constants | | |
|---|---|---|---|
| | $U_{ref}/U_{max}$ | $C_2$ | $C_3$ |
| Papier, Metallpapier (MP), Metallisierter Kunststoff (MKP,MKT, MKU), MKV / *Paper, metallized paper, metallized polypropylene film, metallized polyethylene terephtalate film, metallized acetyl cellulose film, metallized paper film* | 0,5 | 1,07 | 3,45 |
| Polycarbonat (KC, MKC) / *Polycarbonate film metal foil, metallized polycarbonate film* | 0,5 | 1,5 | 4,56 |
| Polyethylenterephtalat (KT), Folien-Polypropylen (KP), Polystyrol (KS) / *Polyethylene terephtalate film metal foil, Polypropylene film metal foil, polystyrene film metal foil* | 0,5 | 1,29 | 4 |
| Glas / *Glass* | 0,5 | 1,11 | 4,33 |
| Glimmer / *Mica* | 0,5 | 1,12 | 2,98 |
| Keramik / *Ceramic* | 0,5 | 1 | 4 |
| Al-Elko, flüssiger Elektrolyt / *Aluminium electrolytic, non-solid electrolyte* | 0,8 | 1 | 1,36 |
| Al-Elko, fester Elektrolyt / *Aluminium electrolytic, solid electrolyte* | 0,8 | 1,9 | 3 |
| Ta-Elko, flüssiger Elektrolyt / *Tantalum electrolytic, non-solid electrolyte* | 0,5 | 1 | 1,05 |
| Ta-Elko, fester Elektrolyt / *Tantalum electrolytic, solid electrolyte* | 0,5 | 1,04 | 9,80 |

**Fig. 6.** Constants $\pi_U$ for capacitors

Also the Temperature dependence is taken into account for capacitors according to equation. The following formula applies up to the maximum permissible junction temperature only.

$$\pi_T = \frac{A \times \exp(Ea_1 \times z) + (1-A) \times \exp(Ea_2 \times z)}{A \times \exp(Ea_1 \times z_{ref}) + (1-A) \times \exp(Ea_2 \times z_{ref})} \quad (14)$$

with

$$z = 11605 \times \left( \frac{1}{T_{U,ref}} - \frac{1}{T_2} \right) \quad (15)$$

$$z_{ref} = 11605 \times \left( \frac{1}{T_{U,ref}} - \frac{1}{T_1} \right) \quad (16)$$

| Kondensatoren / *Capacitors* | Constants | | | | |
|---|---|---|---|---|---|
| | $A$ | $Ea_1$ in eV | $Ea_2$ in eV | $\theta_{U ref}$ in °C | $\theta_1$ in °C |
| Papier, Metallpapier (MP), Metallisierter Kunststoff (MKP, MKT, MKU), Polyethylenterephtalat (KT), Folien-Polypropylen (KP),Polystyrol (KS), MKV / *Paper, metallized paper, metallized polypropylene film, metallized polyethylene terephtalate film, metallized acetyl cellulose film, polyethylene terephtalate film metal foil, polypropylene film metal foil, polystyrene film metal foil, metallized paper film* | 0,999 | 0,5 | 1,59 | 40 | 40 |
| Polycarbonat (KC, MKC) / *Polycarbonate film metal foil, metallized polycarbonat film* | 0,998 | 0,57 | 1,63 | 40 | 40 |
| Glas, Glimmer / *Glass, mica* | 0,86 | 0,27 | 0,84 | 40 | 40 |
| Keramik / *Ceramic* | 1 | 0,35 | - | 40 | 40 |
| Al-Elko, flüssiger Elektrolyt / *Aluminium electrolytic, non-solid electrolyte* | 0,87 | 0,5 | 0,95 | 40 | 40 |
| Al-Elko, fester Elektrolyt / *Aluminium electrolytic, solid electrolyte* | 0,4 | 0,14 | 0 | 40 | 40 |
| Ta-Elko, flüssiger Elektrolyt / *Tantalum electrolytic, non-solid electrolyte* | 0,35 | 0,54 | 0 | 40 | 40 |
| Ta-Elko, fester Elektrolyt / *Tantalum electrolytic, solid electrolyte* | 0,961 | 0,27 | 1,1 | 40 | 40 |
| Veränderliche / *Variable* | 1 | 0,15 | - | 40 | 40 |

**Fig. 7.** Constants $\pi_T$ for capacitors

## 4.3 Database and Norm selection

The Tool works with a database where all standards and their data are saved. For all saved components the user can calculate the failure rate, in reference- or operation condition. In order to keep the database up to date conditions and maintain components can be added or removed, or values can be changed.
First step, selection of a component.

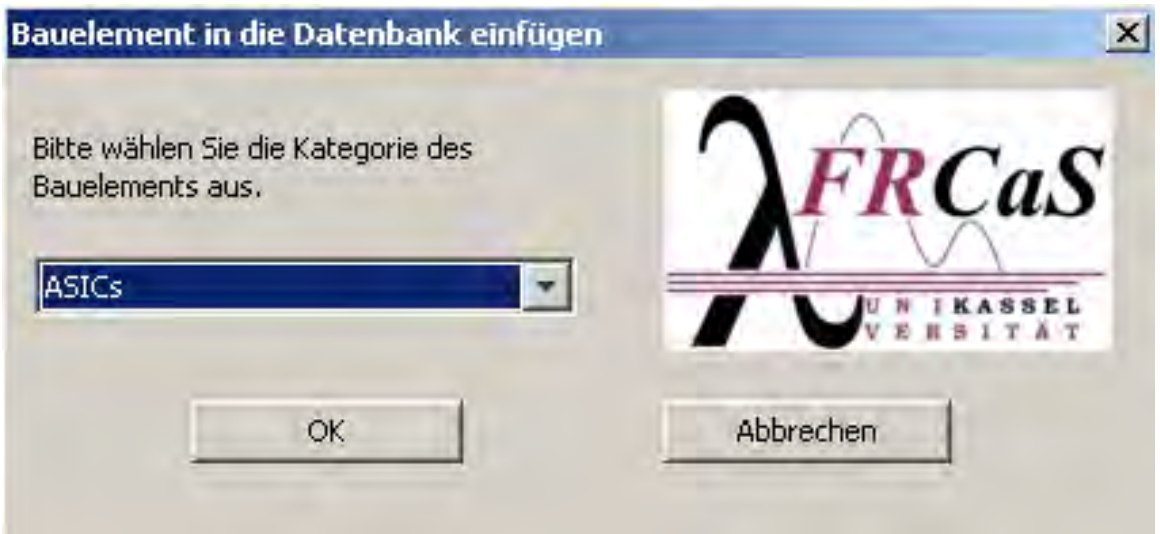

**Fig. 8.** Component selection

Second step, you can change values or put new components in the database.

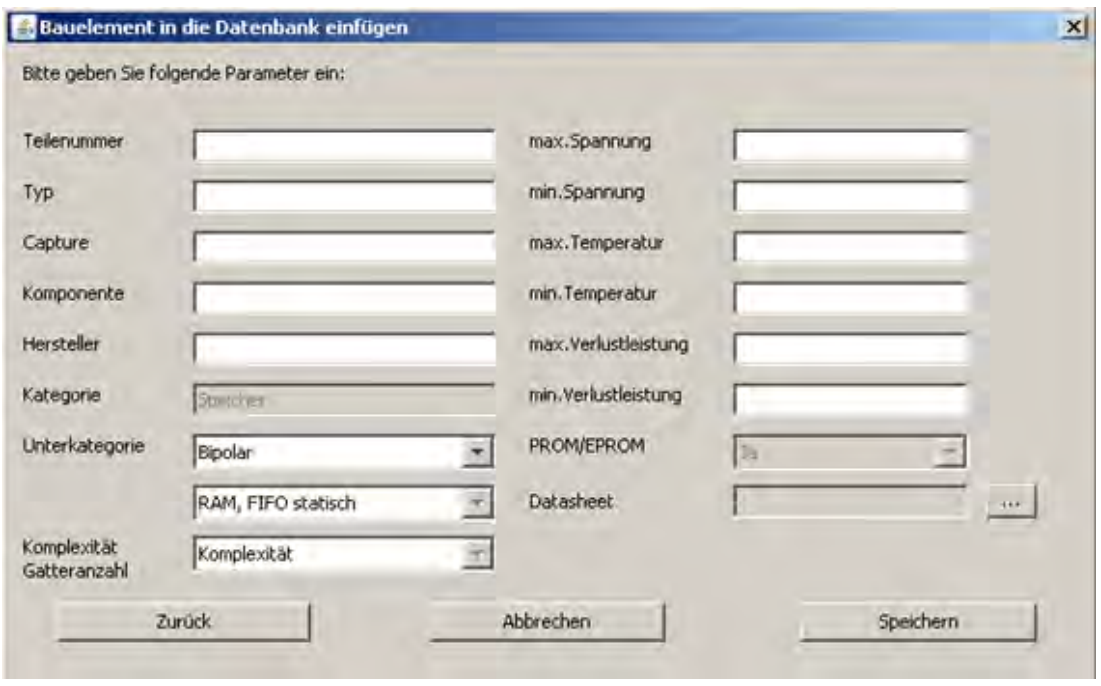

**Fig. 9.** Insert component in database

As already mentioned, this Software Tool offers the user some of the basis of Standards. In this Tool the user can switch the standard in two steps.
First step, the user chooses a standard.

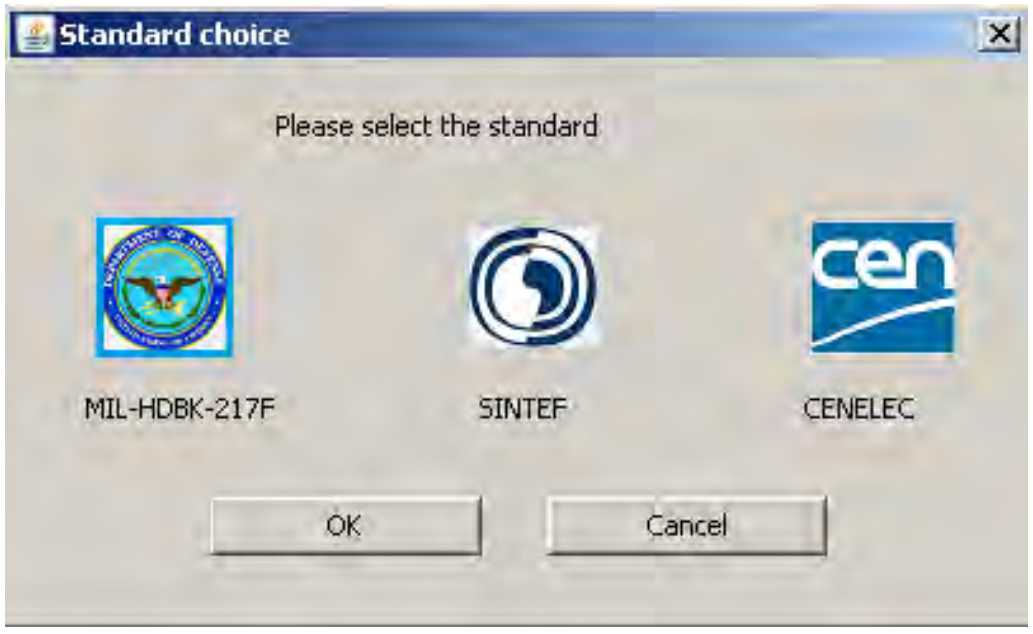

**Fig. 10.** Norm selection

Second step, the Selection of a category of the chosen standard.

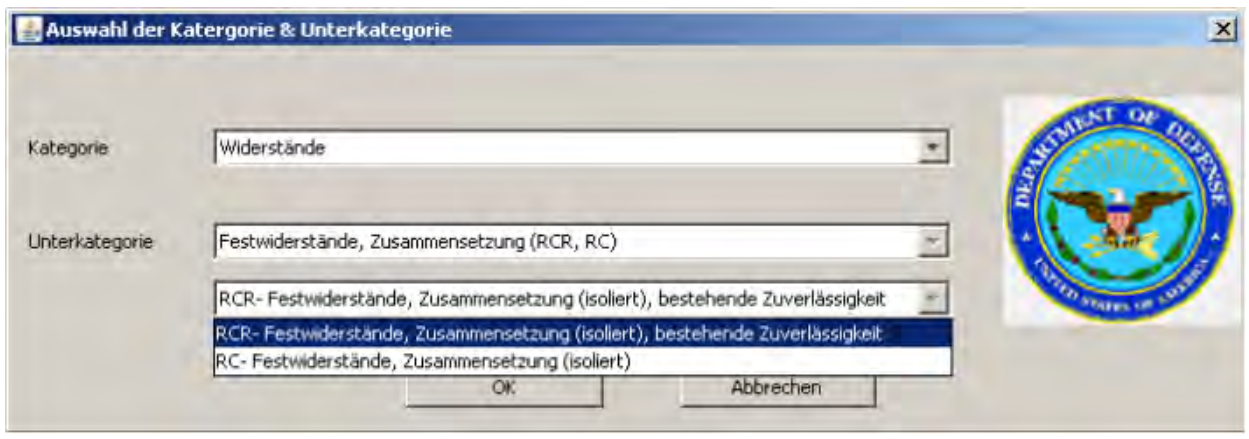

**Fig. 11.** Selection Component from the Military Norm

The tool provides the user with the ability to determine the failure rate of components for operating conditions on the basis of the Military Handbook. Now, the user can compare between the failure rates of the Siemens Standard and the Military Handbook.

## 4.4 Calculation using the new tool of various elements

The Goal in this section is that a circuit from ORCaD can be uploaded to the new Tool (Fig.12).

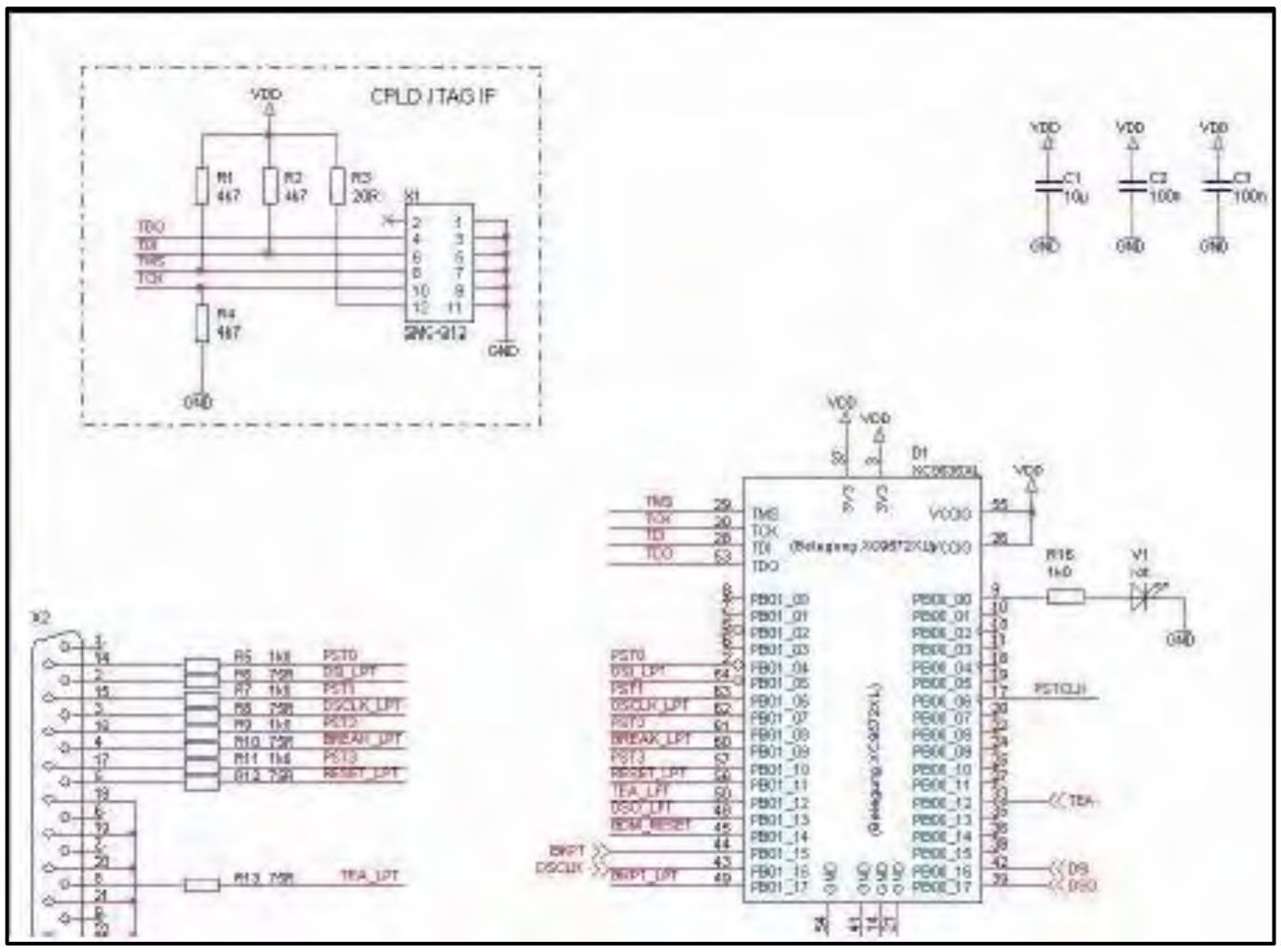

**Fig. 12.** Circuit design in ORCaD

The new FRCaS-Tool is designed in a way that the authorized user has the possibility to determine the failure rate of a component, which is present in the database for operating conditions based on this standard. As a prerequisite, an internet connection must be available. If no internet connection is available, the user may only work with the components that are locally present. The database manages the components out of the assignment of user privileges.

Fig. 13 shows the calculation with the help of an ORCaD – Circuit. To perform such a calculation, one must access the data bank. The database contains the reference failure rates. Using the standard Siemens-Norm, the desired operating failure rate can be determined.

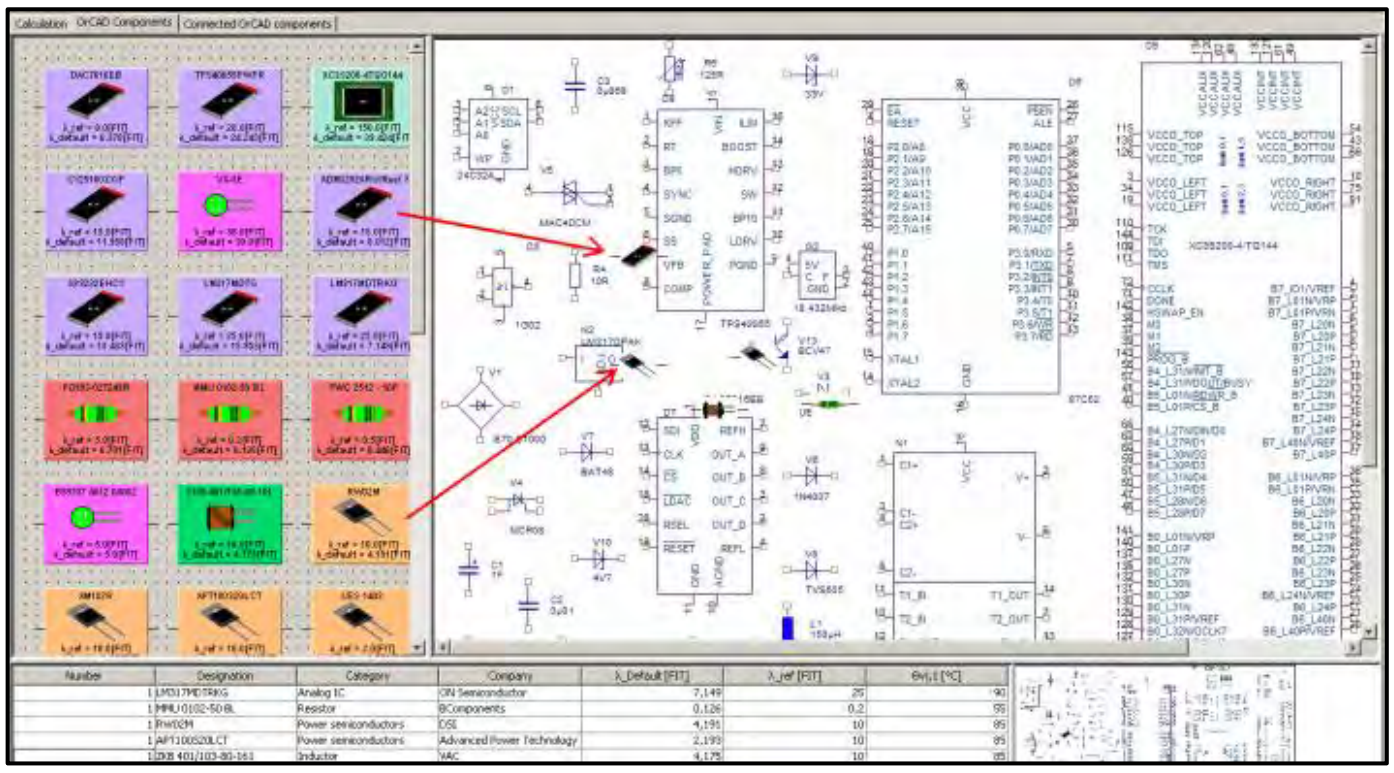

**Fig. 13.** Circuit design in ORCaD

Due to the operating condition, several temperature dependent functions can be determined. For example, the failure rate as a function of $\theta$ can be depicted. Fig. 14 shows the volume function.

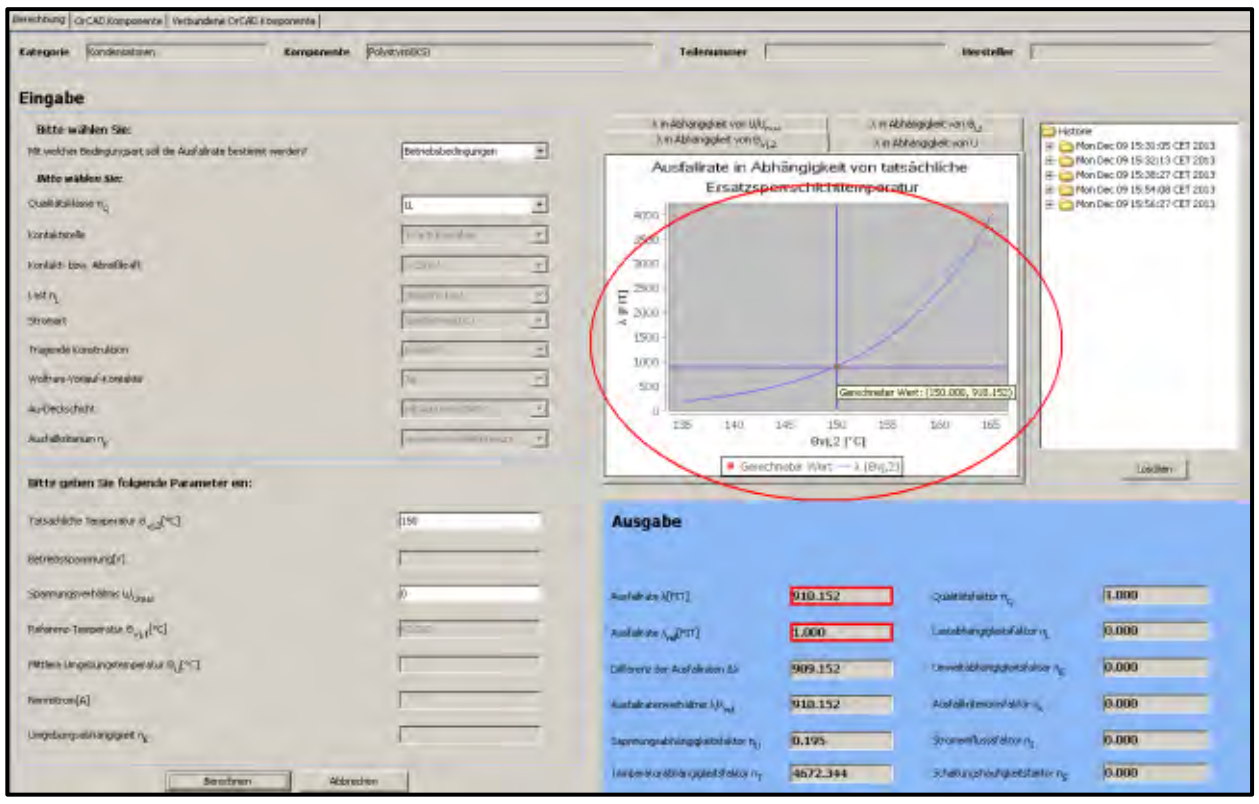

**Fig.14.** Failure rate as a function of temperature

With the help of ORCaD, it is now possible to build different architectures. It is possible to build both safety and availability systems in the new Tool. As an example a one out of tow (1oo2)-architecture is used. The 1oo2 architecture, see Fig. 15, possesses two channels in parallel, where each channel can execute the safety function by itself.

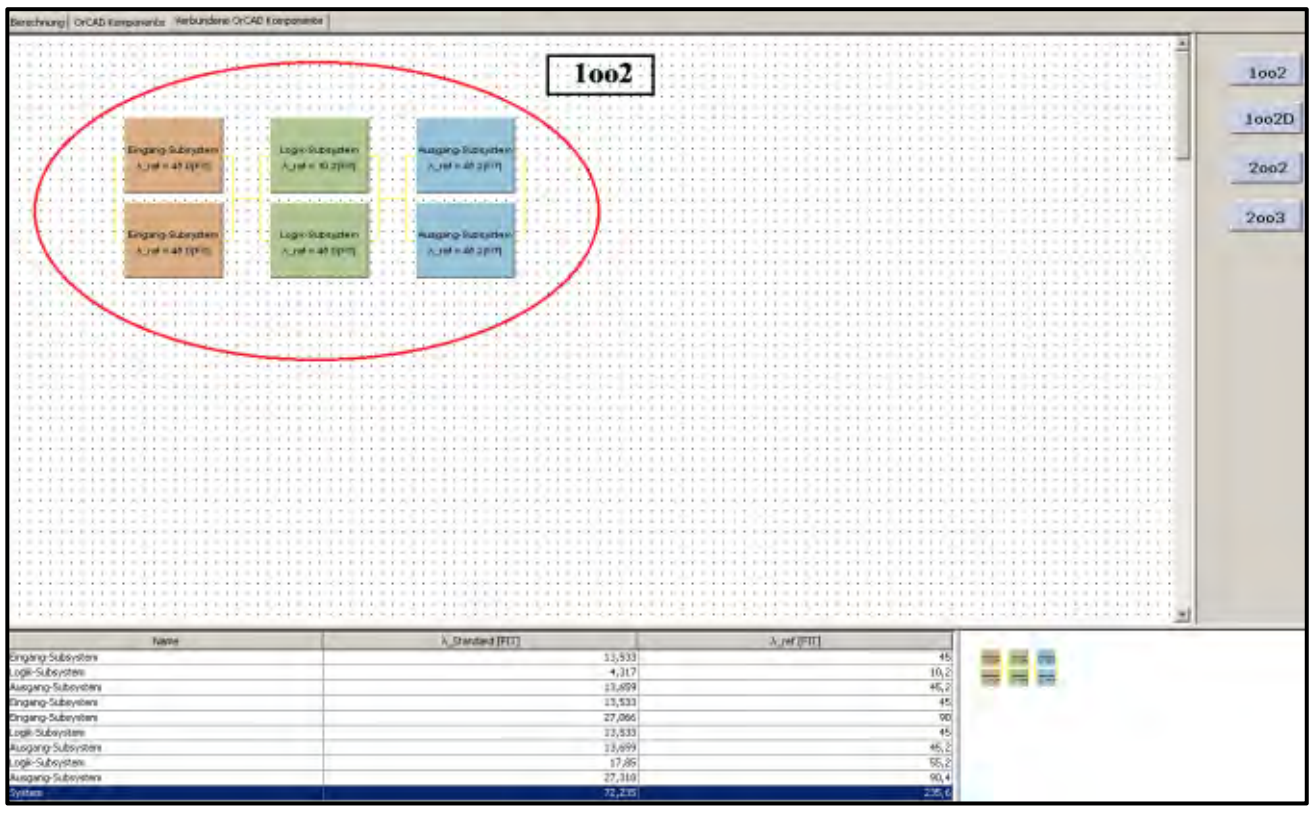

**Fig.15.** Calculation of 1oo2 – Safety Systems under operating conditions

Determining the probability failure on Demand (PFD) based on IEC Standard- 61508:

$$PFD_{avg_1oo2} = 2\left[(1-\beta_D)\lambda_{DD} + (1-\beta)\lambda_{DU}\right]^2 t_{CE} t_{GE}$$

$$+\beta\lambda_{DD} MTTR + \beta\cdot\lambda_{DU}\left(\frac{T_1}{2} + MTTR\right)$$

It is also possible to summarize several differential systems. Therefore the safety-tool can be used for the calculation of different components for reference and company terms on the one hand. On the other hand different calculations can be realized with the help of OrCAD-circuits. Various scenarios can be generated in the Tool. Fig. 16 shows the versatility of the new safety tools. It is still possible to save all the calculations and modeling, and also print out a PDF version.

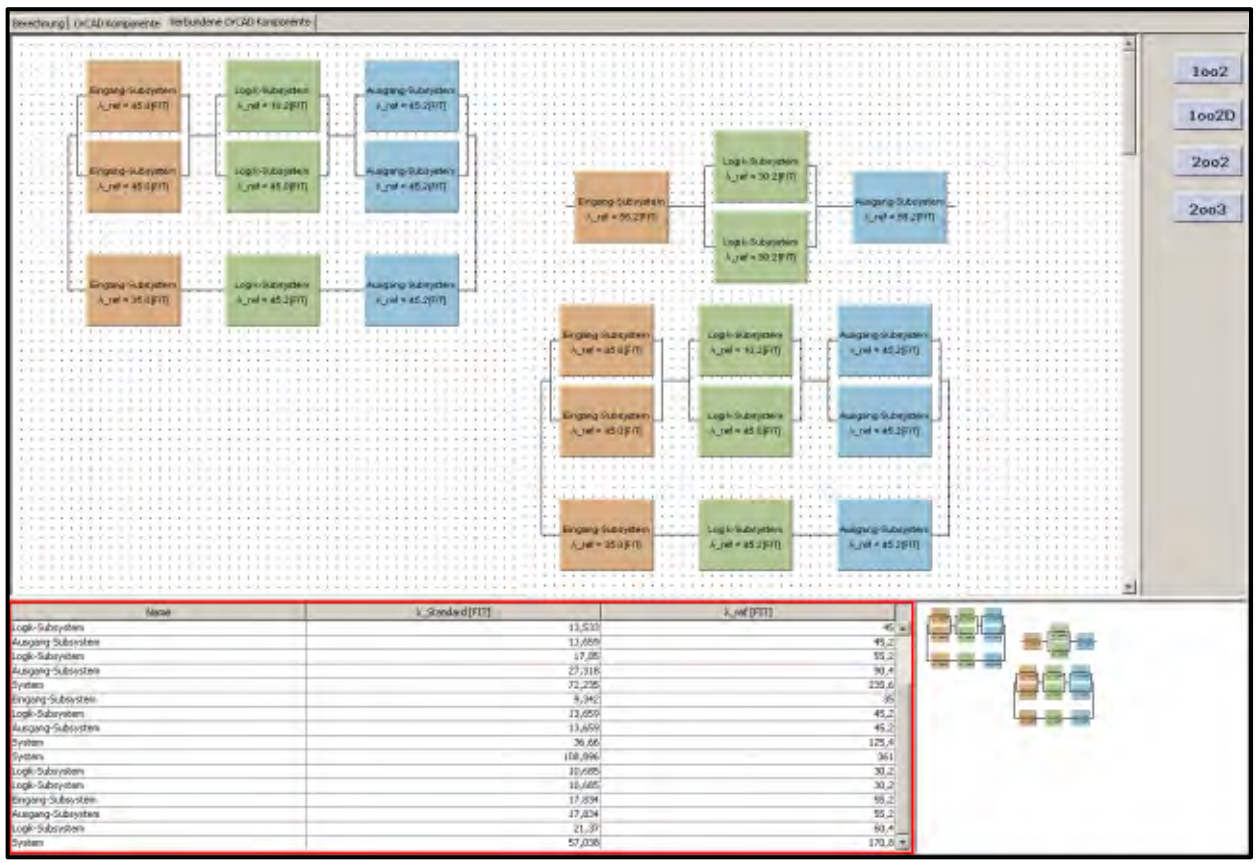

**Fig. 16.** Complex systems

The new Tool will then determine the overall failure rate for both the reference and operating condition.

# 5 Conclusion

Various and complex architectures can be built with the new developed safety tool. Based on the Siemens-Norm, the key safety parameters for both reference and operating conditions can be determined. Further work will be to stipulate that additional standards be implemented in the new Tool to carry out a comparison of the overall failure rates and the PDF value. The next approach will be that different analysis methods are used to achieve the safety parameters.

Thus, the user will have a very simple and convenient hand tool for complex circuit from OrCAD program to analyze and calculate the safety parameters.

# Machine Learning for Contactless Low-Cost Vital Signs Monitoring Systems

Frederico G. C. Lima, A. Albukhari, U. Mescheder

Furtwangen University of Applied Sciences
**lima@hs-furtwangen.de**

**Abstract.** Following a growth of the elderly population in developed countries, a growth in research towards contactless measurement systems for this public has been observed. The development goes often in the direction of intelligent systems to support nursing staffs in assisted living residences. This can also be foreseen for those living alone at home. In this work, two contactless sensing systems are presented, one of them already with an optimized algorithm based on machine learning. Moreover, the optimization of a specific parameter and of an inclusion of a boundary condition for the algorithm are explained.

**Keywords:** Machine learning; Ballistocardiography; Contactless sensors; Cardiac measurements; Low-cost sensors.

## 1 Introduction

An increase of low-cost contactless/cuffless monitoring and alarming systems are required to keep up with the demographic changes in the world. It is estimated that the percentage of the German population above 60 years old will reach 34.6% in 2030 [1]. These systems are important to allow the independence and a normal life of especially elderly people living alone, but they could also be used in retirement homes to alarm nursing or medical assistances.

Different sensor designs have been previously presented to allow vital sign measurements using ballistocardiography (BCG) practically in any position on a bed [2], couch [3] or standing [4] anywhere in the house. In general, unsupervised machine learning (ML) algorithmics (also known as pattern recognition) are used for vital sign monitoring. The reason is that the signals are normally periodical within a known frequency range. Moreover, body movements due to respiration and heartbeat, make vital signals even easier to be sensed. In the case of cardiac measurements, they are realized through BCG. This allows the algorithm to cluster the diverse peaks generated by the human body according to their similarities (normally known as dissimilarities) and determine the cardiac periods. Such measurements can be realized with distinct body positions, in different parts of the house or furniture and diverse sensing principles. However, adjustments related to the clustering parameters and boundary conditions can be optimized to increase the recognition rate of the systems.

In this work, two sensors briefly shown as well as a shown introduction of the ML algorithm. Furthermore, the improvement due to an optimization of parameter and of an inclusion of a boundary condition in the algorithm is present.

## 2 Sensors and Machine Learning

Most of the sensors must be designed considering several different points such as the position of the sensor, sensing object, interface and nature of the signal. To exemplify these in the environments mentioned in the introduction, the furniture selection, measurements required, and interface must be considered in the design to allow proper measurements.

According to the literature, the strongest cardiac component from a human body can be measured along the foot-head axis [5]. Therefore, a low-cost strain-gauge sensor (for scales up to 5 kg) was placed at the top end of the mattress (see Fig. 1). This position choice had the goal of maximizing the cardiac component and reducing the respiration component magnitude in BCG measurements. This allows a higher cardiac recognition rate by the ML algorithm. A 24-bit HX711 analog-to-digital converter from AVIA Semiconductor was used. Electrocardiogram (ECG) measurements using an AD8232 Heart Monitor were synchronized with BCG using an Arduino Uno.

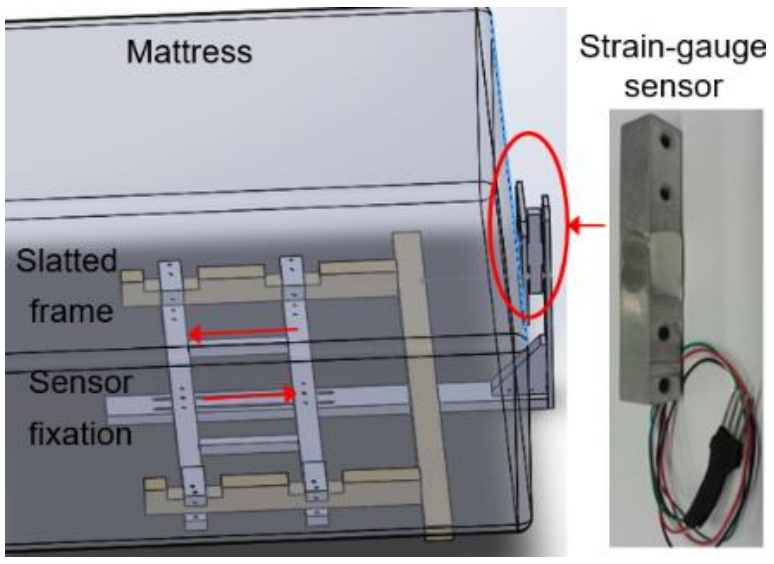

**Fig. 1.** Schematic of the measurement setup using a strain-gauge at the top end of the mattress [2]

For a person seating down, strain-gauges were replaced by piezo disks [3]. This type of setup has the advantage of reacting only to movement changes, and no offset signal due the person's weight is measured. Different signals are generated due to distinct positions. Other than his vital signs, a determination of the subject's position could also be indicated by specific patterns for each position. The dashed circles in Fig 2 were used to mark the different respiratory cycles.

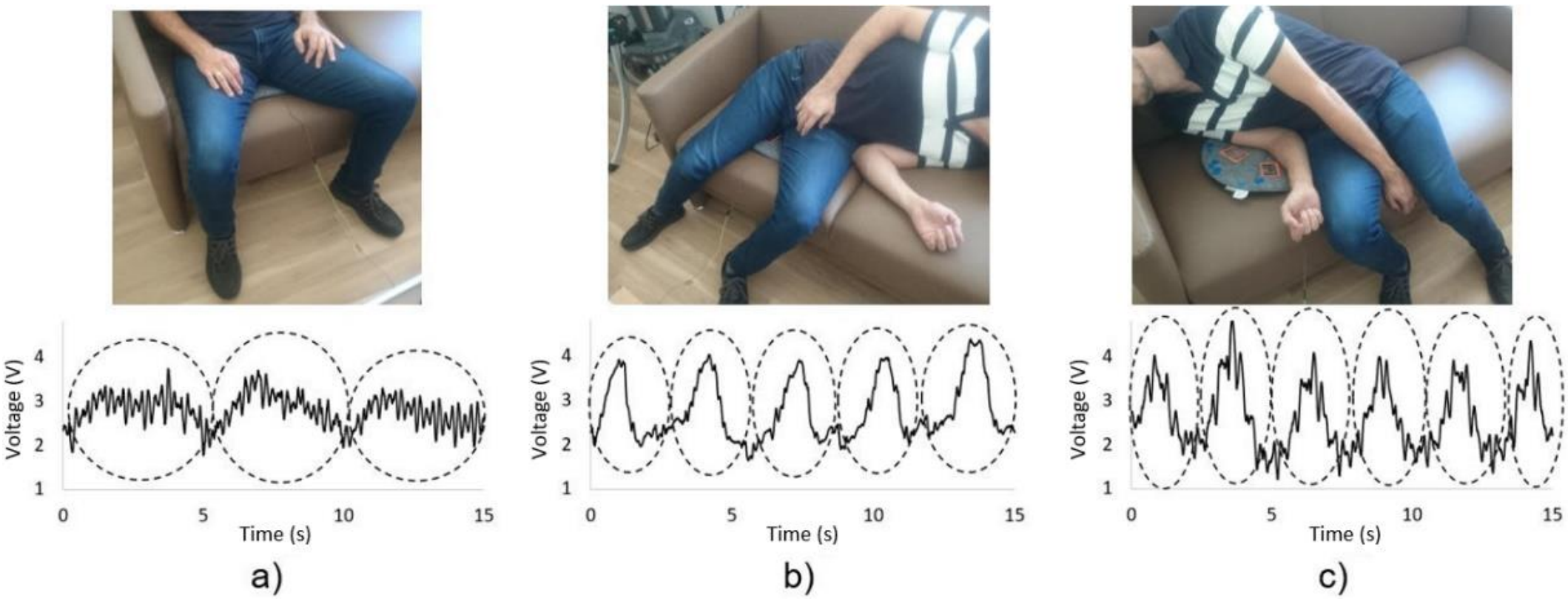

**Fig. 2.** Measurement of a subject a) seating on the coach with the sensors placed right below his trunk, b) seating on the sensor but leaning away from it and c) laying on the coach with the side body over the sensing area [3]

After measuring BCG, different filters were used to separate cardiac and respiratory components [2]. The solution adopted consists of a peak detection function and the allocation of the measurement points in the next 0.66 s in a vector [6]. This corresponds to vector with lengths of 56 for our settings. Then, an arccosine function [7] was chosen to calculate the angle between vectors $x$ and $y$ according to their dissimilarity $d$ for a period of 10.66 s. These angles are compared using equation (1) and its boundaries [2]. For the boundary conditions, $\alpha$ = 3 and $t_{min}$ = 0.33 s were used. After calculating $d$ for all the vectors is this certain window (i.e. 10.66 s), these vectors are clustered, the largest cluster with the lowest $d$ is chosen as the beginning of the cardiac cycle.

$$d(x,y) = \begin{cases} cos^{-1}\left(\frac{x^T y}{\|x\|\|y\|}\right), & if \ \frac{1}{\alpha} < \frac{\|x\|}{\|y\|} < \alpha \ \ and \ \ |t_{x_1} - t_{y_1}| > t_{min} \\ \pi \quad , & otherwise \end{cases} \quad (1)$$

where the vectors first elements are represented by $x_1$ and $y_1$.

## 3 Results and Discussion

The results presented in this section are only from measurements realized on a bed with the strain-gauge setup. An assumption made in our previous work [2], does not consider that the algorithm can select different BCG waves for the following window. In this case, an error is introduced during the transition to the next window. For this reason, effective heartbeat calculations were realized for 10 s, while the detected peaks in the overlapped region of 0.66 s were not considered for calculation (see circle in Fig. 3). Thus, the next calculation is done between first and second selected BCG waves of the second 10 s window shown in Fig. 3. This will be shown in detail in our upcoming study [8].

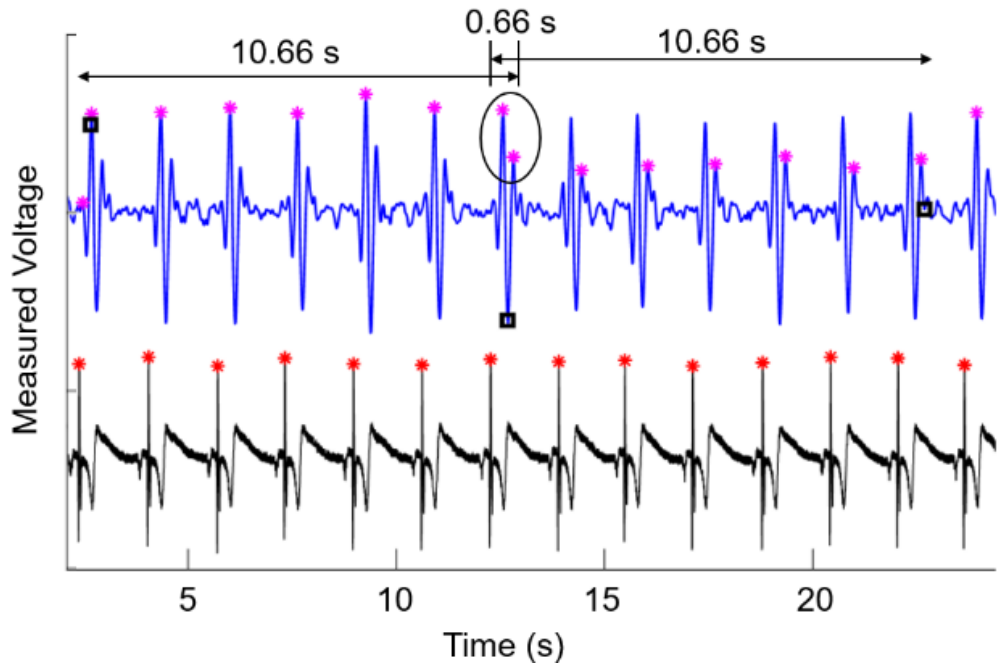

**Fig. 3.** Synchronized ECG and filtered cardiac component from a BCG measurement, *squares* indicate the beginning of a new window and the *pink stars* represent the selected cardiac cycle start

This modification allowed the inclusion of a further boundary condition. All detected peaks received an index number (see Fig. 4, left graph) after peak detection. However, $d$ is directly assumed as $\pi$ if the index difference is smaller than 3. Let us take peak 9 as an example, $d$ is not calculated for peaks 7, 8, 10 and 11. A further improvement was then possible with this boundary condition, a parameter which determines the cluster sizes due to dissimilarity was initially set as 1 rad [2]. This value has been suggested as $\pi/4$ rad for BCG measurements using piezoelectric pressure sensors [7]. However, the optimal value for strain gauges placed at the top end of the bed for a set of almost 10 subjects was determined as 1.2 [8]. The reason for this value might be related to the sensor system, which requires higher dissimilarities to allow the suppression of false positive detections. These adjustments have allowed an average improvement of heartbeat recognition above 8% for different patients laying on a bed in supine, prone, right and left side positions (see Table 1). Even though the same algorithm could be directly used with to analyze the measurements shown in Fig. 2, settings improvements would be required to reach acceptable recognition rates.

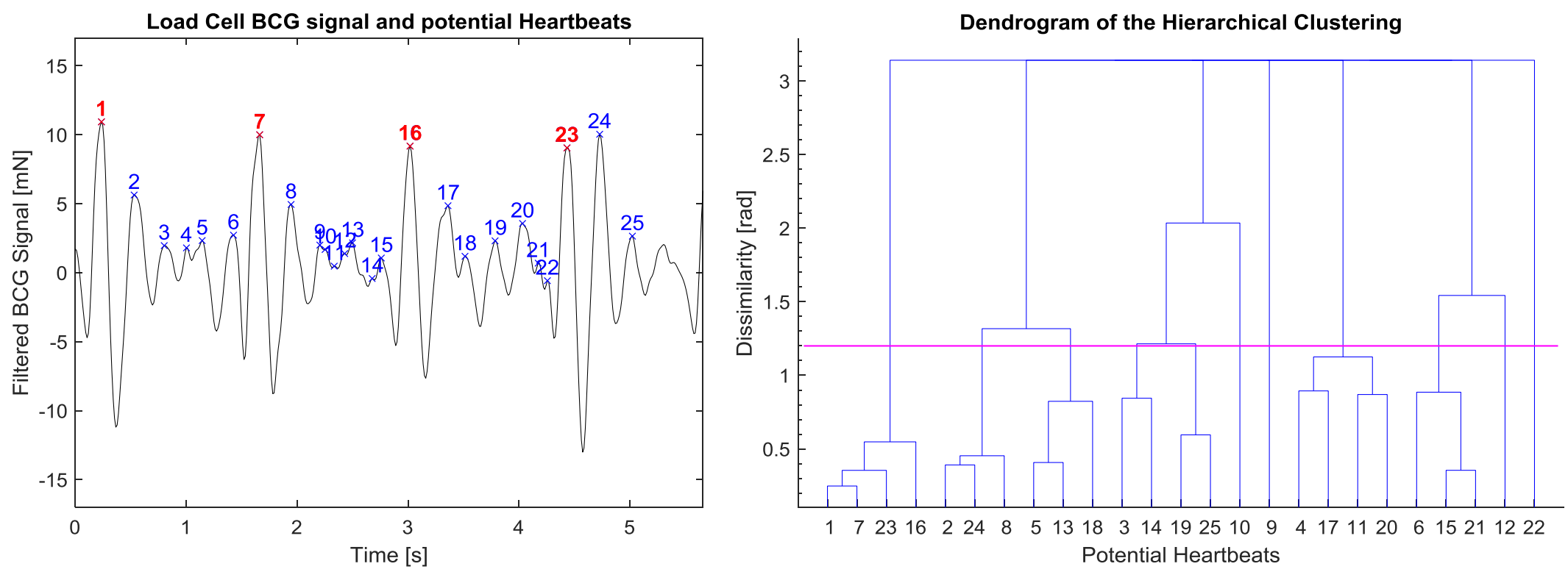

**Fig. 4.** Cardiac component extracted from a BCG measurement after peak recognition, used as starting point for the feature vectors (left side) and dendrogram with clustered vectors (right side)

**Table 1.** Heartbeat recognition in % using either none [2] or index equals 3 [8] for the clustering process

| Subjects | Index | Supine | Prone | Left Side | Right Side |
|---|---|---|---|---|---|
| 1 | None | 96.1 | 75.0 | 89.3 | 90.0 |
| | 3 | 98.2 | 88.6 | 96.6 | 98.0 |
| 2 | None | 89.2 | 91.7 | 78.0 | 77.3 |
| | 3 | 94.3 | 96.4 | 91.2 | 90.0 |

## 4 Conclusion and Outlook

The requirement of contactless monitoring systems especially for elderly people is real, and several measurement principles are already available. Sensor setups still can be improved and optimized to reach better results and more affordable prices. Moreover, it was shown that the actual machine learning algorithms must be adjusted for a certain system and are not universal. Therefore, it is expected that these settings optimizations to improve signal analysis and, consequently, vital sign monitoring be done automatically by the system using artificial neural networking in the future.

## Acknowledgments

This work was partly financed by the Federal Ministry of Education and Research of Germany under the IKT2020 – Innovations in Care for People with Dementia. The authors are grateful for the support from the Alexander Bejan and the staff from the Future Care Lab at the Hochschule Furtwangen where several measurements were carried out.

# Integration of Weighted Terminological Concepts and Vague Knowledge in Ontologies for Decision Making

Nadine Mueller[0000-0002-1658-5002] and Klemens Schnattinger[0000-0003-4565-7930]

Baden-Wuerttemberg Cooperative State University
**{schnattinger,muellnad }@dhbw-loerrach.de**

**Abstract.** Description logics (DLs) are a well-known family of logics for managing structured knowledge. They are the basis for widely used ontology languages. Experience with the use of DLs in applications has shown that their capabilities are not sufficient for every domain. In particular, the decision-making process requires the assessment of two different, sometimes even contradictory influences on decision factors. On the one hand, there are items that belong to certain classes or fulfill certain roles within logically complex constructs, but these memberships are to some extent vague. On the other hand, individual preferences can change depending on the person who drives the decision-making process. Therefore, the challenge when building a framework of decision making, is to take these influencing variables adequately into account by depicting and incorporating both aspects. The paper shows how these requirements can best been modelled by combining fuzzy description logic and weighted description logic. Whereas the first meets the requirement to represent vagueness and ambiguity in ontologies, the second is able to express individual preferences. In addition, the paper shows how to engineer an appropriate and suitable architecture for this purpose.

**Keywords:** Ontology Learning, Weighted Description Logic, Fuzzy Logic, Decision Making

## 1 Introduction

In many cases of decision making, expert knowledge is required. Human experts can identify structural patterns of decision situations in order to model decision processes [1]. From a cognitive-psychological point of view, decision making requires heuristics that ignores some of the information to make decisions more quickly, more economically or more accurately. Being able to work with vague information is critical when dealing with systems that are described by complex ontologies and consist of many instances [2]. Decision making and argumentation interact between processes that use logical thinking or heuristic reasoning. Therefore, it can be argued, that intuitive processes allow access to some form of logical reasoning. But it is also possible, that logic and rationality can be conceived as the domain of explicit higher-level forms of processing.

To formalize this knowledge, description logics offer a powerful tool to structure knowledge and support reasoning. When making decisions, it is often necessary not only to fulfill a set of equivalent requirements but also to take individual preferences into account. This requires an extension of common knowledge bases, the so-called decision bases, which are initially based on multi-attribute utility theory (MAUT) [3]. Since then, various approaches have emerged. Among others, the application of logic for decision and utility theoretical problems are given in [4-6]. In situations where ambiguity occurs, an acknowledged approach is to augment the framework by fuzzy logic, see [7,8]. However, in cases where individual preferences encounter vague knowledge and assertions, neither decision bases nor fuzzy description logic can satisfy these paradigms by its own. To close the gap, this paper offers a framework to model ambiguity and individual preferences at the same time. It brings together fuzzy description logic with weighted description logic. For the ease of understanding, we will first introduce the architecture which has been used in this specific context. Afterwards, to get a fine grasp of the combined framework of weighted description logic and fuzzy description logic, we will familiarize the reader with both separately. Initially, we establish the basics of weighted description logic. Subsequently, we present the fuzzy description logics and focus how it supports the modelling of ambiguous and vague knowledge. At the same time, the demarcation to probabilistic settings is highlighted. After combining these two approaches, our fuzzy decision base framework is introduced. Finally, we show how this framework can support the decision-making process within the respective architecture.

## 2 Preliminaries

The following sections present our architecture for opinion and consensus mining OMA, classical description logic and two extensions, the weighted description logic and fuzzy description logic.

### 2.1 Opinion & Consensus Mining Architecture OMA

The original Opinion Mining Architecture (OMA) is part of a project of the same name. OMA was used for the first time for sentiment analysis from tweets for the financial sector [9]. To achieve an automated calculation of sentiment scores from texts, traditional approaches of natural language processing (such as POS tagging, parsing) and machine learning from texts (such as n-gram, syntactic/semantic features) were used for the preprocessing of the texts [10]. In addition, an extension of description logic [11], the so-called weighted description logic [6], was used to automatically calculate the sentiment scores. The idea of separating the text processing task (filtering out relevant phrases) and the decision support task (evaluating extracted phrases) comes from the text understanding system SYNDIKATE [12] and its qualitative calculus [4].

In order to explain the extension of OMA to include consensus mining and decision making, we first clarify the essential components of the OMA. In **Fig. 1** we see (from top to bottom):

- the $TBox$, which accommodates models via compliances, rules, judgements, etc.
- the $ABox$, which contains unweighted statements on the model of the $TBox$
- $U_iBox$ (on the right side), that contain different preference models of experts

From a technical point of view, the models of the TBox are entirely expressed in description logic by means of terminological concepts, roles and is-a-relations. The elements of the ABox are terminological assertions that enter into an instance-of-relationship with concepts of the TBox. At this point it should be noted that these assertions will be created by the text-processing task from newspapers, social media, political programs, etc. (see the cloud on the left side in **Fig. 1**). The preference model of an expert $e_i$ is shown in the $U_i$Box. A preference model consists of a priori preference relation over attributes of concepts (see black circles in **Fig. 1**). Each model represents the individual utility function of an expert $e_i$. With these a priori preference relations of an expert a first a posteriori preference order for each expert's choice can be derived (see the individual preference orders in **Fig. 1**). Note that the preference model of each expert can be extracted a priori by the text processing task or be entered directly by each expert. Next, the individual preference relations of each expert are used to build consensus or in case of only one expert to directly retrieve the best possible choice respective decision. The former is done by means of incomplete fuzzy preference relations for group decision making [13], which repeatedly adapts the preference relations of all experts until a satisfying consistent consensus is achieved. The theoretical basis of this approach comes from [14] and its IOWA operator. For more details see [15].

**Fig. 1**. The Opinion & Consensus Mining Architecture OMA

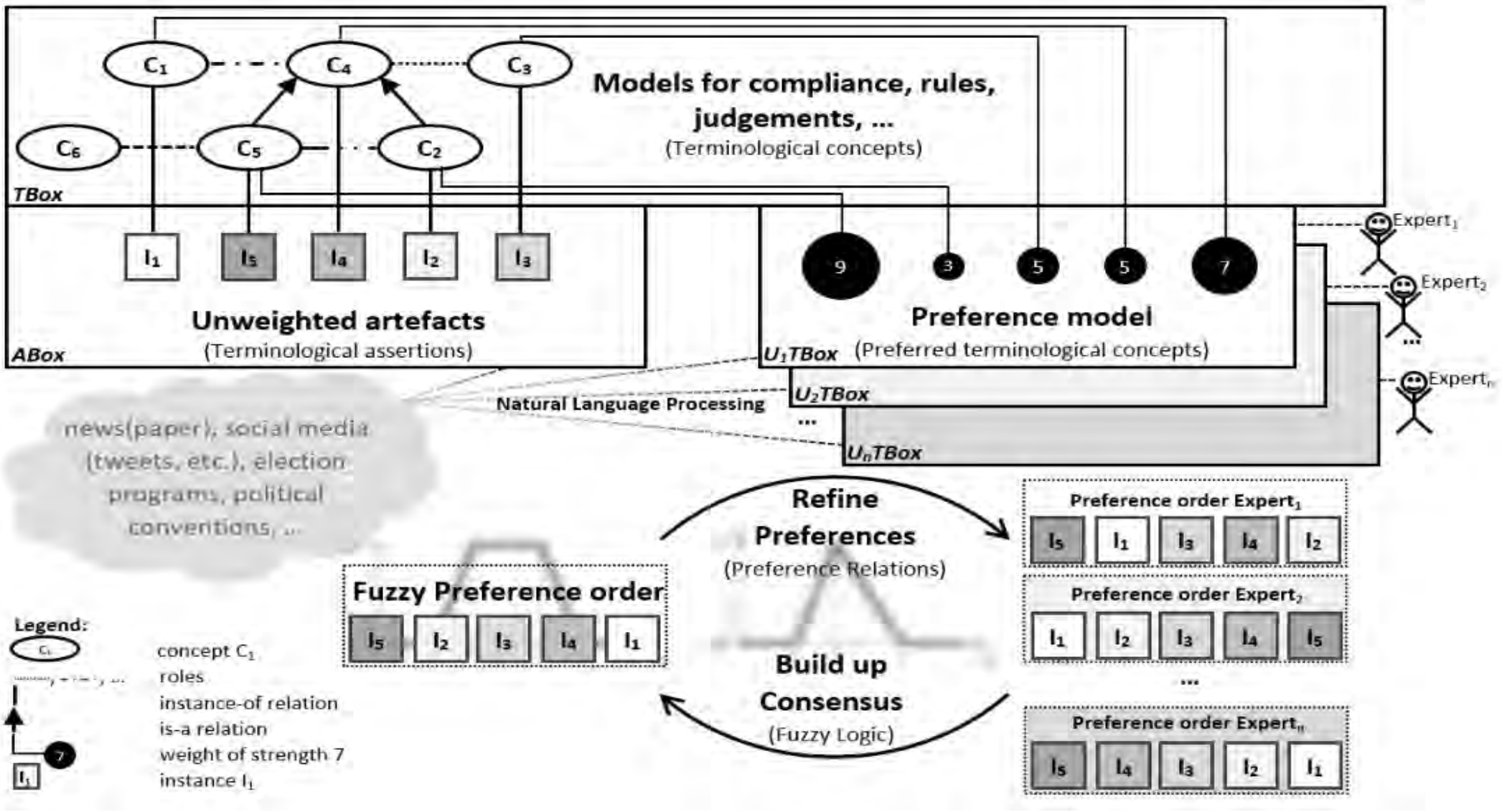

## 2.2 Description Logic

Description Logics (DLs) [11] are a family of logic-based knowledge representation formalisms. They can be used to represent and reason on the knowledge of an application domain. The basis of description logics is a common family of languages, known as description languages, which have a set of constructors to ontologies consisting of create concept (class) and role (property) descriptions.

A description language consists of an alphabet with unique concept names ($N_C$), role names ($N_R$) and individual (object) names ($N_I$). In addition, so-called constructors are used to create concept and role descriptions. Depending on which constructors are allowed, there are many different description languages. Some of them form the basis for the ontology language of the semantic web [16].

**The $\mathcal{SROIQ}$ Description Logic.** An expressive description language is called $\mathcal{SROIQ}$-$\mathcal{DL}$ [17]. $\mathcal{SROIQ}$-$\mathcal{DL}$ is compatible with OWL2, the current standard of the semantic web [18] and thus the most reasonable description language within the architecture introduced above. A formal definition of the notions $\mathcal{SROIQ}$-roles and $\mathcal{SROIQ}$-concepts, as well as the underlying model-theoretic semantics (the interpretation is written as $I$, the domain as $\Delta^I$ and the interpretation function as $\cdot^I$) can be found in [17]. Below are some examples of the syntax and semantics of $\mathcal{SROIQ}$-$\mathcal{DL}$:

**Table 1.**: Example syntax and semantics of $\mathcal{SROIQ}$-DL

| Constructor | Syntax | Semantics |
|---|---|---|
| Top | $\top$ | $\Delta^{\mathcal{I}}$ |
| bottom | $\bot$ | $\emptyset$ |
| general negation | $\neg C$ | $\Delta^{\mathcal{I}} \backslash C^{\mathcal{I}}$ |
| conjunction / disjunction | $C \sqcap D \,/\, C \sqcup D$ | $C^{\mathcal{I}} \cap D^{\mathcal{I}} / \, C^{\mathcal{I}} \cup D^{\mathcal{I}}$ |
| exists restriction | $\exists R.C$ | $\{x \in \Delta^{\mathcal{I}} \mid \exists y.\langle x,y\rangle \in R^{\mathcal{I}} \wedge y \in C^{\mathcal{I}}\}$ |
| value restriction | $\forall R.C$ | $\{x \in \Delta^{\mathcal{I}} \mid \forall y.\langle x,y\rangle \in R^{\mathcal{I}} \rightarrow y \in C^{\mathcal{I}}\}$ |
| at-most restriction | $\leq nR$ | $\{x \in \Delta^{\mathcal{I}} \mid \#\{y \in \Delta^{\mathcal{I}} \mid R^{\mathcal{I}}(x,y)\} \leq n\}$ |
| at-least restriction | $\geq nR$ | $\{x \in \Delta^{\mathcal{I}} \mid \#\{y \in \Delta^{\mathcal{I}} \mid R^{\mathcal{I}}(x,y)\} \geq n\}$ |
| concept definition / concept specialisation | $D \equiv C \,/\, D \sqsubseteq C$ | $D^{\mathcal{I}} = C^{\mathcal{I}} / D^{\mathcal{I}} \subseteq C^{\mathcal{I}}$ |

In DLs, we distinguish between terminological knowledge (so-called $\mathcal{T}$Box) and assertional knowledge (so-called $\mathcal{A}$Box). A $\mathcal{T}$Box is a set of concept inclusions $C \sqsubseteq D$ and concept definitions $C \equiv D$. An $\mathcal{A}$Box is a set of concept assertions $a{:}\,C$ as well as role assertions $(a,b){:}\,R$.

A so-called concrete domain $\mathcal{D}$ is defined as a pair $(\Delta^{\mathcal{D}}, pred(\mathcal{D}))$. $\Delta^{\mathcal{D}}$ is the domain of $\mathcal{D}$ and $pred(\mathcal{D})$ is the set of predicate names of $\mathcal{D}$. The following assumptions have been applied: $\Delta^{\mathcal{I}} \cap \Delta^{\mathcal{D}} = \emptyset$ and for each $P \in pred(\mathcal{D})$ with arity n there is $P^{\mathcal{D}} \subseteq (\Delta^{\mathcal{D}})^n$. According to [11], functional roles are denoted with lower case letters, for example with $r$. In description logics with concrete precise domains, $N_R$ consists of functional roles and ordinary roles. A role $r$ is functional if for every $(x,y) \in r$ and $(w,z) \in r$ it is necessary that $x = w \Rightarrow y = z$. Functional roles are explained as partial functions from $\Delta^{\mathcal{I}}$ to $\Delta^{\mathcal{I}} \times \Delta^{\mathcal{D}}$. Within $\mathcal{SROIQ}$ all statements gathered about roles are captured in an $\mathcal{R}$Box, which for the sake of convenience and for compatibility to the definitions in [11] is not applied to our examples.

Next, we will build a knowledge base (originally introduced in [9]) of a domain that will be used in the further course of the work. Its purpose is pure illustrative, so that reasoning and entailment is obvious. We will note explicit and implicit knowledge (*"- ik -"*):

$\mathcal{T} = \{Device \sqsubseteq \top, Equip \sqsubseteq \top, Device \sqcap Equip \sqsubseteq \bot, PoorEquip \sqsubseteq Equip, WellEquip \sqsubseteq Equip,$
$PoorEquip \sqcap WellEquip \sqsubseteq \bot, Device \equiv \exists hasWeight.>_{0g} \sqcap \exists hasPrice.>_{0€} \sqcap \forall equipped.Equip,$
$Tablet \equiv Device \sqcap \exists hasPrice.>_{200€}, InexpensiveTablet \equiv Tablet \sqcap \exists hasPrice.\leq_{500€},$
$ExpensiveTablet \equiv Tablet \sqcap \exists hasPrice.\geq_{900€}, InexpensiveTablet \equiv \neg ExpensiveTablet,$
$ExpensiveTablet \equiv \neg InexpensiveTablet, LightWeightTablet \equiv Tablet \sqcap \exists hasWeight.\leq_{900\,g},$
$Convertable \sqsubseteq UpperclassTablet, UpperclassTablet \equiv Tablet \sqcap \forall equipped.WellEquip,$
$LowerclassTablet \equiv Tablet \sqcap \forall equipped.PoorEquip,$
$UpperclassTablet \sqcap LowerclassTablet \sqsubseteq \bot,$ - *ik* -}

$\mathcal{A} = \{tab_1{:}\,Tablet, (tab_1, 999€){:}\,hasPrice, (tab_1, 710g){:}\,hasWeight,$
$equipment_1{:}\,WellEquip, (tab_1, equipment_1){:}\,\forall equipped.WellEquip,$
$tab_1{:}\,ExpensiveTablet, tab_1{:}\,LightWeightTablet,$ - *ik* -
$tab_1{:}\,UpperclassTablet,$ - *ik* -
$tab_2{:}\,Tablet, (tab_2, 399€){:}\,hasPrice, (tab_2, 1250g){:}\,hasWeight, equipment_2{:}\,PoorEquip,$
$(tab_2, equipment_2){:}$
$\forall equipped.PoorEquip, tab_3{:}\,Tablet, (tab_3, 600€){:}\,hasPrice, tab_3{:}\,Convertable,$
$tab_2{:}\,InexpensiveTablet, tab_2{:}\,LowerclassTablet,$ - *ik* -
$tab_3{:}\,UpperclassTablet, equipment_3{:}\,WellEquip,$ - *ik* -
$(tab_3, equipment_3){:}\,\forall equipped.WellEquip$ - *ik* -}

A considerable amount of knowledge is implicitly revealed in the terminological knowledge base. In this case, a lot of knowledge is available about the domain but no support to make a comprehensible decision. For this reason, the capabilities of the knowledge base are extended by the possibility to depict individual preferences.

## 2.3 Weighted Description Logic

Weighted description logic (WDL) can be regarded as a generic framework, the so-called decision base [6]. We use an a priori preference relation over attributes (called the ontological classes). Thereby, an *a posteriori* preference relation over choices (called *ontological individuals*) can be derived. Formally, a utility function $U$ over $\mathcal{X}$ (the set of attributes) is defined ($U: \mathcal{X} \to \mathbb{R}$). Additionally, a utility function $u$ defined over choices, which uses logical entailment, extends the utility function U to the subset of choices [19]. Modelling attributes takes place in two steps:

1. Each attribute is modelled by a concept
2. For every value of an attribute a new (sub)concept is introduced

For instance, if *equipped* is an attribute to be modelled, it is simply represented by the concept $Equipment$ (i.e. $Equipment \in \mathcal{X}$). An equipment can be regarded as a value, as if it was a concept of its own. If "*well equipped"* is a value of the attribute *equipped*, the attribute set $\mathcal{X}$ is simply extended by adding the concept $WellEquip$, as a sub-concept of $Equipment$. It should be noted, that an axiom is introduced to guarantee the disjointedness (e.g. $PoorEquip \sqsubseteq \neg WellEquip$) and that this procedure results in a binary term vector for $\mathcal{X}$, because an individual *c* (as a choice) is either a member of a specific attribute of the concept set $\mathcal{X}$ or not.

Given a total preference relation (i.e. $\succcurlyeq_{\mathcal{X}}$) over an ordered set of not necessarily atomic attributes $\mathcal{X}$, and a function $U: \mathcal{X} \to \mathbb{R}$ that represents $\succcurlyeq$ (i.e., $U(X_1) \geq U(X_2)$ **iff** $X_1 \succcurlyeq_{\mathcal{X}} X_2$ for $X_1, X_2 \in \mathcal{X}$). The function $U$ assigns an *a priori* weight to each concept $X \in \mathcal{X}$. Therefore, one can say, that "$U$ makes the description logic weighted". The utility of a concept $X \in \mathcal{X}$ is denoted by $U(X)$. The following applies: The greater the utility of an attribute the more the attribute is preferable.

As mentioned above, a *choice* is an individual $c \in N_I$. $\mathcal{C}$ denotes the finite set of choices. To determine a preference relation (*a posteriori*) over $\mathcal{C}$ (i.e. $\succcurlyeq_{\mathcal{C}}$), which respects $\succcurlyeq_{\mathcal{X}}$, a utility function $u(c) \in \mathbb{R}$ is introduced. $u(c)$ indicates *the utility of a choice c* relative to the attribute set $\mathcal{X}$. Also, a utility function $U$ over attributes as an aggregator is introduced. For simplicity, the symbol $\succcurlyeq$ is used for both choices and attributes whenever it is evident from the context.

Within a consistent knowledge base $\mathcal{K} \coloneqq \langle \mathcal{T}, \mathcal{A} \rangle$, consisting of a $\mathcal{T}$Box $\mathcal{T}$ and an $\mathcal{A}$Box $\mathcal{A}$, the $\sigma$-utility is a particular $u$ and is defined as $u_\sigma(c) \coloneqq \sum\{U(X) \mid X \in \mathcal{X} \text{ and } \mathcal{K} \vDash \text{c}: X\}$ and is called the *sigma utility of a choice* $c \in \mathcal{C}$ . $u_\sigma$ triggers a preference relation over $\mathcal{C}$ i.e., $u_\sigma(c_1) \geq u_\sigma(c_2)$ **iff** $c_1 \succcurlyeq c_2$. Each choice corresponds to a set of attributes, which is logically *entailed* e.g. $\mathcal{K} \vDash \text{c}: X$. Due to the criterion additivity, each selection $c$ corresponds to a result.

Putting things (DL, $U$ and $u$) together, a generic $\mathcal{U}$Box (so-called *Utility Box*) is defined as a pair $\mathcal{U} \coloneqq (u_\sigma, U)$, where $U$ is a utility function over $\mathcal{X}$ and $u_\sigma$ is the utility function over $\mathcal{C}$. Also, a decision base is defined as a triple $D = (\mathcal{K}, \mathcal{C}, \mathcal{U})$ where $\mathcal{C} \subseteq N_I$ is the set of choices and $\mathcal{U} = (u, U)$ is an $\mathcal{U}$Box. Note: $\mathcal{K}$ provides assertional information about the choices and terminological information about the agent ability to reason over choices.

Now, we expand our tablet example by using different utility boxes ($\mathcal{U}_i$) resp. utility functions ($u_{i,\sigma}$) of two experts:

- For expert 1 applies $\mathcal{U}_1 = \{(InexpensiveTablet, 50), (UpperClassTablet, 40), (LightWeightTablet, 40)\}$, $u_{1,\sigma}(tab_1) = 40 + 40 = 80$, $u_{1,\sigma}(tab_2) = 50$ and $u_{1,\sigma}(tab_3) = 40$. It follows that $tab_1 \succ tab_2 \succ tab_3$.
- For the expert 2, however, applies $\mathcal{U}_2 = \{(InexpensiveTablet, 60), (UpperClassTablet, 20), (LightWeightTablet, 10)\}$, $u_{2,\sigma}(tab_1) = 20 + 10 = 30$, $u_{2,\sigma}(tab_2) = 60$ and $u_{2,\sigma}(tab_3) = 20$. It follows, that $tab_2 \succ tab_1 \succ tab_3$.

Within this decision base an expert with the utility box $\mathcal{U}_1$ would classify $tab_1$ as first choice whereas an expert with a different utility box in this example $\mathcal{U}_2$ would prefer $tab_2$. Two different problems appear looking at $tab_3$. One is that for this tablet a weight (in the sense of mass not weighting of a concept according to WDL sense) is not known. Therefore, the reasoning fails doing an instance check for this tablet on the concept *LightWeightTablet*. For this reason, when calculating the utility value, it is treated as it would not be an instance of *LightWeightTablet*. But the membership of this concept is unknown, and one cannot reason that the instance does not belong to the concept *LightWeightTablet*. The other problem is that at a price of 600€, the tablet is neither inexpensive nor expensive (*InexpensiveTablet* resp. *ExpensiveTablet)*. Although the price is well-known, the utility function treats this tablet the same way as expensive ones, which is not quite reasonable in this scenario. To eliminate this problem the knowledge base is extended by fuzzy description logic and then combined with the decision base, which is introduced in subsequent chapters.

## 2.4 Fuzzy Description Logic

To deal with the ambiguity of the underlying domain, it is necessary to clarify, where this uncertainty comes from. Either the uncertainty is due to a probabilistic cause or due to vagueness. If the first situation occurs, then a statement is either "true" or "false" to some possibility (in the sense of likelihood), whereas in the second situation, a statement is to some degree (in the sense of reaching a graded level) either "true" or "false". For more information on these two approaches, see [20].

In the context of the above choice of tablets, the underlying ambiguity arises from vagueness. To model vague knowledge description logic is enriched with fuzzy logic, which enables reflecting the degree of membership to a certain concept. According to [21] a fuzzy set is defined by its characteristics the so-called membership function.

Let $X$ be a non-empty set of individuals, then a class $A$ in $X$ is characterized by its membership function $f_A: X \to [0,1]$ and assigns to each $x \in X$ a real number within the interval [0,1]. This value represents how large the degree of its membership to $A$ is. The membership function defined for fuzzy sets fulfills some essential properties which appear to be natural, see also [22]:

- $\forall x \in X: A = \emptyset$ **iff** $f_A(x) = 0$
- $\forall x \in X: A' = X \setminus A: f_{A'}(x) = 1 - f_A(x)$
- $\forall x \in X: A \sqsubseteq B: f_A(x) \leq f_B(x)$
- $\forall x \in X: A \cup B: f_{A \cup B}(x) = \max(f_A(x), f_B(x))$
- $\forall x \in X: A \cap B: f_{A \cap B}(x) = \min(f_A(x), f_B(x))$

To augment the possibilities of fuzzy sets, algebraic operations can also be defined. There are plenty of definitions e.g. Łukasiewicz logic, Gödel logic. For a detailed comprehension of these algebraic operations and their relation to DLs, see [23]. In this work, the standard fuzzy logic (SFL) is used, but all others can be applied as well. Some definitions can be found in table 2.

This toolset of fuzzy set definitions and algebraic operators can now be applied to description logics to reflect ambiguity and vagueness in knowledge bases. Hence, an individual that is an instance of a concept only to a certain degree for example can be modelled suitably.

**Table 2.** Definitions of algebraic operations

| **Algebraic operator** | **SFL** |
|---|---|
| $a \otimes b$ / $a \oplus b$ | $\min(a, b)$ / $\max(a, b)$ |
| $a \Rightarrow b$ / $\ominus a$ | $\max(1 - a, b)$ / $1 - a$ |

To formally quote this fuzziness of description logic axioms we use the syntax of [7]. The conceptional syntax of fuzzy description logics is the same as for description logics defined above (see chapter 2.2). The semantic however reflects the fuzzy logic. Therefore, a *fuzzy interpretation* is a pair $\mathcal{I} = (\Delta^{\mathcal{I}}, \cdot^{\mathcal{I}})$ consisting of a non-empty set called the domain and a fuzzy interpretation function. This function maps individuals as usual and concepts into membership functions $\Delta^{\mathcal{I}} \rightarrow [0,1]$. Accordingly, the roles are mapped into $\Delta^{\mathcal{I}} \times \Delta^{\mathcal{I}} \rightarrow [0,1]$. Consequently $C^{\mathcal{I}}$ is the membership function of the fuzzy set $C$. Hence, a concept is interpreted as fuzzy set.

**Example**
A specific *tablet* is an instance of the concept *Convertable* only to a certain degree depending on its features. We therefore extend the description logic and allow capturing this degree as a fuzzy value and write $\langle tab_3 \colon Convertable, 0.8\rangle$. This means that $tab_3$ is at least an instance of the concept $Convertable$ with the degree of 0.8. Analogously, $Convertable^{\mathcal{I}}(tab_3)$ gives back the minimal degree of $tab_3$ being a convertable tablet under the interpretation $\mathcal{I}$.

The properties of fuzzy sets and algebraic operators are now applied to interpretations of $\mathcal{SROIQ}$ and, according to [24], lead to the following example rules for all $d \in \Delta^{\mathcal{I}}$ (non-exhaustive list):

**Table 3.** Fuzzy semantics

| **Syntax** | **Semantics** |
|---|---|
| $C \sqcap D$ | $(C \sqcap D)^{\mathcal{I}}(d) = \min\{C^{\mathcal{I}}(d), D^{\mathcal{I}}(d)\}$ |
| $\neg C$ | $(\neg C)^{\mathcal{I}}(d) = 1 - C^{\mathcal{I}}(d)$ |
| $C \sqsubseteq D$ | $(C \sqsubseteq D)^{\mathcal{I}} = \inf_{d \in \Delta^{\mathcal{I}}} C^{\mathcal{I}}(d) \Rightarrow D^{\mathcal{I}}(d)$ |
| $\exists R.C$ | $(\exists R.C)^{\mathcal{I}}(a) = \sup_{b \in \Delta^{\mathcal{I}}}\{\min(R^{\mathcal{I}}(a,b), C^{\mathcal{I}}(b))\}$ |

**Example**
Let $\mathcal{K}$ be the knowledge base above, but now the concept $LightWeightTablet$ is not strictly defined according to classic DL, but intuitively with the help of fuzzy DL. Within the $\mathcal{T}$Box, the row $\exists hasWeight. \leq_{900g} \sqsubseteq LightWeightTablet$ will be replaced by the following two constructs:

$$\langle(\exists hasWeight._{\geq 900g} \sqcap \exists hasWeight._{\leq 1100g}) \sqsubseteq LightWeightTablet, 0.6\rangle$$
$$\langle \exists hasWeight._{\leq 900g} \sqsubseteq LightWeightTablet, 1\rangle$$

indicating that every tablet which has a weight less than 1100g should still be considered as a light tablet to a certain degree (here 0.6). For $tab_3$ the exact weight is not known but relating information vary between 900g an 1100g with strong tendencies to the upper threshold. Therefore, the $\mathcal{A}$Box is adjusted accordingly: $\langle tab_3 \colon \exists hasWeight._{\geq 900g}, 0.5\rangle$ and $\langle tab_3 \colon \exists hasWeight._{\leq 1100g}, 0.9\rangle$. The $\mathcal{T}$Box reveals $tab_3 \colon \exists hasWeight._{\geq 900g} \sqcap tab_3 \colon \exists hasWeight._{\leq 1100g}$

$$= \min\{tab_3 \colon \exists hasWeight._{\geq 900g}, tab_3 \colon \exists hasWeight._{\leq 1100g}\} = \min\{0.5, 0.9\} = 0.5$$

and $tab_3$ is therefore a $LightWeightTablet$ with the minimal degree of $\max\{1 - 0.5, 0.6\} = 0.6$.

# 3 Weighted Fuzzy Description logic

For the weighted fuzzy description logic, the background knowledge base $\mathcal{K} = (\mathcal{T}, \mathcal{A})$ will be allowed to capture also vague knowledge and assertions, which is formally noted as $\mathcal{K} \approx (\mathcal{T}, \mathcal{A})$. This knowledge base is then extended by the set of choices $\mathcal{C}$ and the utility box $\mathcal{U}$ steering the decision making. The advantage of this framework is, that these weights can be independently articulated and do not need to be compared against each other like in [25].

**Definition**
A triple $\mathcal{D} \approx (\mathcal{K}, \mathcal{C}, \mathcal{U})$, where $\mathcal{K}$ is a fuzzy knowledge base, $\mathcal{C}$ a set of choices and $\mathcal{U}$ a utility box is called a *fuzzy decision base*.

Note: Entities of the $\mathcal{U}$Boxes are concepts relevant to the decision-making process, including their specific individual weights. After the reasoning for each of the existing choices, the instance check completes and reveals whether a choice belongs to a specific concept or not. Assume a choice $c: U$ belongs to a concept, then this might also be vague within the fuzzy description logics. Hence it leads to constructs like $\langle c: U, n\rangle$ with $n \in [0,1]$.

**Definition**
Let $\langle c: U, n\rangle$ be a fuzzy assertion and $(U, w)$ a weighted attribute, then a *fuzzy utility value of c respective to U* is $u_{f\sim\sigma}(c: U) \stackrel{\text{def}}{=} w \cdot n$

If the assertion is not fuzzy, then $n$ is simply set to 1. If the instance $c$ belongs to the complement of $U$ with a membership degree of 1, then the fuzzy utility value for $c$ on this attribute is 0 (as $n$ is then 0).

**Example**
In case for $tab_3$ the calculated respective reasoned membership degree to a light weight tablet is 0.6, formally written $\langle tab_3: LightWeightTable, 0.6\rangle$ and expert $U_1$ defines a weight of 40 for this attribute, formally written as $(LightWeightTablet, 40)$, then $u_{f\sim\sigma}(tab_3) = 24$. The individual weight is a bit more than half of the initially defined one, as the degree of membership for this tablet is only 0.6. As the utility function is additive the utility measure for a choice is the sum across all relevant attributes.

**Definition**
Let $\mathcal{D} \approx (\mathcal{K}, \mathcal{C}, \mathcal{U})$ be a fuzzy decision base with a utility box $\mathcal{U}$ of the cardinality $k = |\mathcal{U}|$ then the *$\mathcal{U}$Box fuzzy utility value of c* is $u_{f\sim\mathcal{U},\sigma} \stackrel{\text{def}}{=} \sum_{i=1}^{k} u_{f\sim\sigma}(c: U_i)$.

By calculating the $\mathcal{U}$Box fuzzy utility values of each choice *c*, a total ordering of the set $\mathcal{C}$ is naturally given. The ideal solution is therefore the choice with the highest fuzzy utility value relative to the $\mathcal{U}$Box.

**Definition**
Let $\mathcal{D} \approx (\mathcal{K}, \mathcal{C}, \mathcal{U})$ be a fuzzy decision base with a utility box $\mathcal{U}$ of the cardinality $k = |\mathcal{U}|$ then *the ideal fuzzy choice* is $c_{f\sim s} \stackrel{\text{def}}{=} \arg\underset{c\epsilon\mathcal{C}}{max}(\sum_{i=1}^{k} u_{f\sim\sigma}(c: U_i))$.

**Example**
For an expert with the utility box $U_1$ the summarized utility value for $tab_3$ is $\sum_{i=1}^{k} u_{f\sim\sigma}(c: U_i) = 24 + 40 = 64$. The ranking of choices for this expert, changes to $tab_1 \succ tab_3 \succ tab_2$, which means that $tab_3$ is preferred to $tab_2$. In this scenario, the problem remains that the price of the tablet is neither expensive nor inexpensive, but unknown. Therefore, it is indispensable to design a consistent respective complete fuzzy decision base by ensuring that each attribute listed in the $\mathcal{U}$Box is correct and decidable in the knowledge base.

**Definition**

A fuzzy decision base $\mathcal{D} \approx (\mathcal{K}, \mathcal{C}, \mathcal{U})$ is called *complete*, if for every relevant attribute out of the $\mathcal{U}$ Box $U \in \mathcal{U}$ a fuzzy value for every $c \in \mathcal{C}$ is deducible: $\forall\ U \in \mathcal{U}, \forall c \in \mathcal{C}\ \exists n \in [0,1]: \langle c: U, n\rangle$. Thus, the fundamentals are defined to make a reasonable decision in the above scenario.

# 4 Results and Outlook

To complete the fuzzy decision base of the example above, $tab_3$ requires a fuzzy value for the attribute "$InexpensiveTablet$" and $tab_2$ for "$LightWeightTablet$". Therefore the $\mathcal{T}$Box is extended to reveal fuzzy values also for weights above 1100g and prices in between 500€ and 900€. The following expressions are added to the $\mathcal{T}$Box:

$$\langle \neg\exists hasWeight._{\leq 1100g} \sqsubseteq LightWeightTablet, 0\rangle$$
$$\langle (\exists hasPrice._{>500€} \sqcap \exists hasPrice._{<900€}) \sqsubseteq InexpensiveTablet, 0.5\rangle$$
$$\langle (\exists hasPrice._{>500€} \sqcap \exists hasPrice._{<900€}) \sqsubseteq ExpensiveTablet, 0.5\rangle$$

The first line indicates that tablets with a weight above 1100g do not belong to the concept $LightWeightTablet$ at all. The fuzzy values of the second and third line represent the membership degrees of those tablets which have a price within this interval and is set manually to 0.5 as arithmetic mean between the two categorizations inexpensive and expensive. Thus, $tab_3$ is a member of the concept *Inexpensive* with a degree of 0.5 and a member of *Expensive* with the same degree.

With standard fuzzy logic the fuzzy value of a concept's complement is: $(\neg C)^{\mathcal{I}}(d) = 1 - C^{\mathcal{I}}(d)$, which entails the following implicit knowledge:

$$\langle InexpensiveTablet \sqcap \exists hasPrice._{\geq 900€}, 0\rangle$$

For example, if $tab_1$ has the price of 999€, then the following applies:

$\langle tab_1: ExpensiveTablet, 1\rangle$ and $\langle tab_1: \neg ExpensiveTablet, 0\rangle$

By means of this complete decision base the above decision can be derived properly. The utility values for $tab_3$ within $\mathcal{U}_1$ and $\mathcal{U}_2$ are:

$$u_{f\sim 1,\sigma}(tab_3) = 0.5 \cdot 50 + 40 + 0.6 \cdot 40 = 89$$
$$u_{f\sim 2,\sigma}(tab_3) = 0.5 \cdot 60 + 20 + 0.6 \cdot 10 = 56$$

For an expert with the utility box $U_1$ $tab_3$ is his or her first choice, while the other expert still chooses $tab_2$. Through the augmented decision base by fuzzy logic a model is defined, which represents reality much better. Because there was uncertainty around $tab_3$ a first calculation in a conventional decision base revealed a distorted result. By incorporating the vague knowledge existing in this domain, the expert would have chosen $tab_3$ instead of $tab_1$.The strength of this framework is that vague assertions together with individual preferences are deliberated properly.

Also, it becomes obvious how weighting influences the decision. As the first utility box has almost balanced weights, the second one has a strong tendency towards inexpensive tablets. Using this $\mathcal{U}$Box the first choice is still $tab_2$. But the second choice is now $tab_3$ and not $tab_1$. Both tablets were initially not belonging to the concept $InexpensiveTablet$, but with the help of fuzzy logic, the strong preference for inexpensive tablets causes that $tab_3$ passes $tab_1$.

Summarized, complete fuzzy decision bases offer a strong possibility to model real world situations which need to respect ambiguity and individual preferences and at the same time support a comprehensible decision-making process. Further researches need to reveal supporting algorithms to detect and locate incompleteness to support the creation of complete fuzzy decision bases. Overall, the creation of these underlying ontologies is time-consuming and a non-trivial, manual process. To facilitate this, new approaches with deep learning algorithms have risen [26].

The use of such techniques is another milestone on the way to a fully automated decision making process.

# Proactive Error Prevention in Manufacturing Based on an Adaptable Machine Learning Environment

Holger Ziekow[1], Ulf Schreier[1], Alaa Saleh[1], Christof Rudolph[2], Katharina Ketterer[2], Daniel Grozinger[2], Alexander Gerling[1]

[1] Furtwangen University
{holger.ziekow, ulf.schreier, alaa.saleh, gera}@hs-furtwangen.de
[2] Sick AG
{Christof.Rudolph, Katharina.Ketterer, Daniel.Grozinger}@sick.de

**Abstract.** This paper gives an introduction into a concept for proactive error prevention in manufacturing. The challenges for research include the accuracy of predictions, the automated adaptation of prediction models to heterogeneous and varied processes, and the management of prediction models in changing processes. The solution is a combination of machine learning technology, big data approaches and knowledge modelling.

**Keywords:** Proactive Error Preventing, Machine Learning, Manufacturing.

## 1 Introduction

Technological developments such as the Internet of Things and Industry 4.0 hold the potential for innovation in production processes. The costs are going to be lowered and quality of the production processes will be increased [1] [2]. These developments coincide with significant advances in the processing of mass data and machine learning methods [3]. Currently many tools with different merits and weaknesses are available for the processing of mass data [4], which can be combined with concepts such as the lambda architecture [5]. We are at a point where the current technologies are ready to capture high-resolution production data and quickly evaluate those with complex methods. The challenge for production companies is to exploit the technological potential for specific innovations in production processes and to scale the use of technology in practice.

## 2 Adaptable Prediction Models

Production companies especially in high-wage countries face the challenge of producing high-quality products at competitive prices. With the project PREFERML we want to develop prediction methods which can predict and proactively avoid production errors in current and future processes. These methods, promise to greatly production costs and waste can be greatly reduced. For practical use, especially in heterogeneous and highly dynamic production processes, it is important to increase the degree of automation in the creation and use of prediction methods. Currently, the creation and maintenance of prediction models is highly demanding in terms of manual effort and time of human experts. Production companies carry out many production processes which are changing over time. To be able to create and use numerous models for different processes, it is necessary to increase the degree of automation in modelling and maintenance. The aim of this project is to achieve this and thereby to enabling the widespread use of proactive error prevention in production. The interdisciplinary combination of expertise in data analysis, software architecture of business applications and domain knowledge from production is indispensable for the investigation of practicable prognosis models in the production process.

## 3 Related Work

Using production data and big data technologies for prediction in production is the subject of current research [6] [7]. The challenges for research include (1) the accuracy of predictions, (2) the automated adaptation of prediction models to heterogeneous and varied processes, and (3) the management of prediction models in changing processes.

The prediction accuracy is significantly influenced by the appropriate selection and pre-processing of the data which use feature extraction and selection, as well as the parameterisation and selection of the prediction models [8] [9]. This often happens with the help of domain knowledge and as part of a manual process. However, there are approaches in research to automate these processes and the individual adaptation of models [9] [10] [11]. Recent work has shown the use of semantic models for the modelling of background knowledge in industrial applications [11] [12]. This work [13] already considers their application for the automated selection of model building features for model building. Here it is necessary to tailor and expand such approaches to the domain of error prediction in production processes.

Regarding the administration of analytical models and their scalable use, a research gap has been identified and initial approaches developed [14] [15]. A key challenge is to identify and address changes in model quality and, if necessary, causal changes in the deployment environment. This requires continuous monitoring of performance parameters of the models and of relevant conditions of the deployment environment. Technologies of complex event processing are suitable as a basis to detect certain situations in the available sensor data streams and to initiate necessary reactions [16] [17]. Previous works in the field of complex event processing for the evaluation of data streams in the production already exist [18] [19]. However, the definition of the relevant situations is not set by Complex Event Processing but is to be described for the respective task via event patterns. For monitoring of predictive models, it is necessary to adjust the use of complex event processing technology and to automate the definition of relevant event patterns.

## 4 Solution Approach

The proposed solution is a combination of machine learning technology, big data approaches and knowledge modelling. The combination of these technologies and their specific future development should solve the main challenges at the creation and scalable use of the methods for proactive error prevention. Especially the following problems should be solved: (1) high quality prediction for production errors, (2) automated creation of many predictive models, (3) automated maintenance of the models in the production. **Fig. 1** show the complete concept of the approach to solve the mentioned challenges. The basic components are explained below.

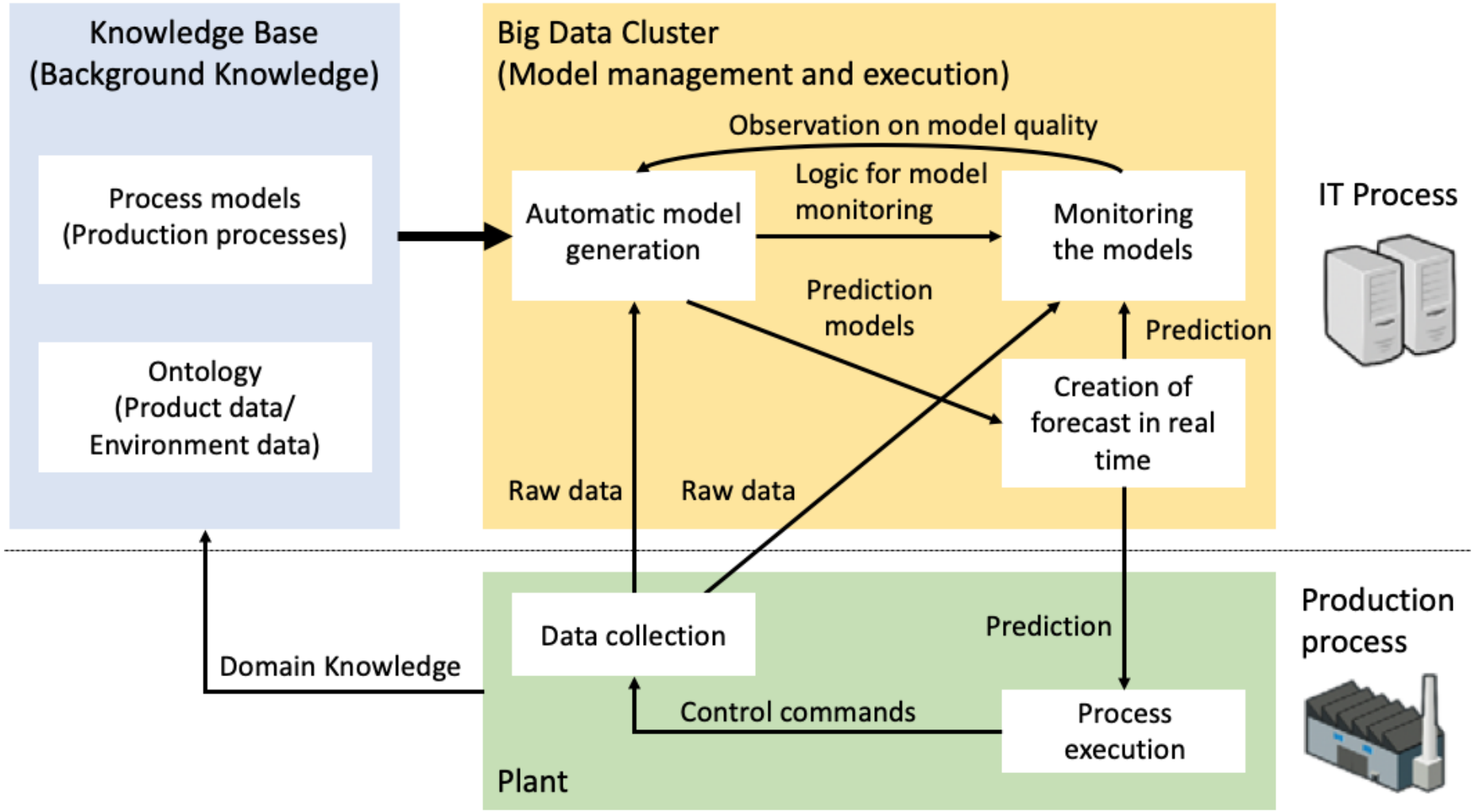

**Fig. 1.** Conception overview over the approach

A fundamental component of the solution is the modelling and use of background knowledge about the production processes and collected data. A domain expert with concrete process knowledge can often name relevant relationships. This background knowledge will get used to support the automatic creation and maintenance of prediction models. For example, when selecting data for a prediction model it is critical to select only relevant and non-redundant sources related to the predicted value. This selection can be done by heuristics but is faced with challenges, in particular for high dimensional data. Our approach is to create a modelling method that can be used to capture the required domain knowledge and make it available for automated data processing. Existing technologies such as ontologies are candidates for modelling language for production data and may be adapted to suit the requirements.

Another key element in the concept is the automatic generation of prediction models. Therefore, statistical models and machine learning models are going to be used. With these models, error predictions can be captured both as classification and as regression problems. The occurrence of certain error classes in current and future processes should be detected using classification and be avoided in advance. Regression will be used to predict the properties of products. If the expected properties differ from the objective, the parameters of the production process can be changed during operation and during production breaks. The corresponding product property can thus be corrected proactively.

With the help of big data technologies for parallel and highly scalable data processing, the solution spectrum for the model building will be investigated to find a case specific optimised solution. The parallelisation possibilities of established and emerging big data technologies should be used and if necessary adapted to achieve an efficient solution of the highly complex calculations.

The creation of models is not related to real-time requirements and can be implemented through batch-processing technologies. In comparison, the execution of prediction models must be continuous and - for some cases - in real time. The implementation of the prediction with technologies of stream processing and complex event processing should be investigated and suitable solutions should be identified. Technologies of stream- and complex event processing should be analysed to identify a suitable solution

Due to changes in the production process or in the data collection, so called concept drift may occur, which may result in a loss of accuracy for the prediction models. Therefore, methods should be explored, which detect crucial changes in an early stage. The necessary logic is to be generated in the context of the automated modelling process. The objective is to automatically adapt the monitoring logic to the specific models and relevant inputs.

# 5 Conclusion

In this paper, we shed light over the need for effective predictive maintenance for manufacturing companies and production factories. Also, we gave a short overview of our scheme to use background knowledge and ontologies to prevent costly waste and increase the production output. The desired result is a combination of background knowledge, big data and machine learning technologies. This outcome will be used to create predictive models and automatically adapt them to newly emerging variants in production lines. Among the main objectives of our intended system is to handle arising concept drifts, and continuously adapt itself to the requirements of the production environment.

# Acknowledgement

This project was funded by the German Federal Ministry of Education and Research, funding line “Forschung an Fachhochschulen mit Unternehmen (FHProfUnt)“, contract number 13FH249PX6. The responsibility for the content of this publication lies with the authors. Also, we want to thank the company SICK AG for the cooperation and partial funding.